\documentclass[final,3p,times,twocolumn]{elsarticle}

\usepackage{graphicx}
\usepackage{amsmath}
\usepackage{amssymb}
\usepackage{booktabs}
\usepackage{multirow}
\usepackage{soul}
\usepackage{caption}
\usepackage{subcaption}
\usepackage{enumitem}
\usepackage{xspace}
\usepackage{algpseudocode}
\usepackage{algorithm}
\usepackage[dvipsnames]{xcolor}
\usepackage{xcolor}

\usepackage[T1]{fontenc}
\usepackage{pgfplots}
\usepackage{adjustbox}

\usepackage{tikz}
\usetikzlibrary{positioning}
\usepackage{caption}
\usepackage{subcaption}
\usepackage{graphicx}
\usepackage{xcolor}
\colorlet{Blue}{red!0!green!0!blue!255}
\pgfplotsset{
    compat=newest,
    xlabel near ticks,
    ylabel near ticks
}

\algrenewcommand\algorithmicrequire{\textbf{Input:}}
\algrenewcommand\algorithmicensure{\textbf{Output:}}
\algnewcommand\algorithmicforeach{\textbf{for each}}
\algdef{S}[FOR]{ForEach}[1]{\algorithmicforeach\ #1\ \algorithmicdo}

\usepackage[pagebackref,breaklinks,colorlinks]{hyperref}
\usepackage[capitalize]{cleveref}
\crefname{section}{Sec.}{Secs.}
\Crefname{section}{Section}{Sections}
\Crefname{table}{Table}{Tables}
\crefname{table}{Tab.}{Tabs.}

\newcommand{\cQ}{\mathcal{Q}}

\newcommand{\bA}{\mathbf{A}}

\newcommand{\cP}{\mathcal{P}}

\newcommand{\cF}{\mathcal{F}}

\newcommand{\bF}{\mathbf{F}}

\newcommand{\bM}{\mathbf{M}}

\newcommand{\bp}{\mathbf{p}}

\newcommand{\bq}{\mathbf{q}}

\newcommand{\bS}{\mathbf{S}}

\newcommand{\cN}{\mathcal{N}}

\newcommand\Set[2]{\{\,#1\mid#2\,\}}

\makeatletter
\newcommand{\onedotA}{\ifx\@let@token.\else.\null\fi\xspace}
\DeclareRobustCommand\onedot{\futurelet\@let@token\onedotA}
\newcommand{\eg}{\emph{e.g}\onedot} 
\newcommand{\ie}{\emph{i.e}\onedot} 
 
\newcommand{\etc}{\emph{etc}\onedot}

\newcommand{\etal}{\emph{et al}\onedot}
\makeatother

\usepackage{amssymb}

\journal{ISPRS Journal of Photogrammetry and Remote Sensing}

\begin{document}

\begin{frontmatter}


\title{Semantic Segmentation on 3D Point Clouds with High Density Variations}

\author[inst1]{Ryan Faulkner}

\affiliation[inst1]{organization={Australian Institute for Machine Learning - University of Adelaide},
            addressline={Cnr North Terrace \& Frome Road}, 
            city={Adelaide},
            postcode={5000}, 
            state={SA},
            country={Australia}
            }
\affiliation[inst2]{organization={Maptek},
            addressline={31 Flemington St}, 
            city={Glenside SA},
            postcode={5065}, 
            state={SA},
            country={Australia}
            }
\author[inst2]{Luke Haub}
\author[inst2]{Simon Ratcliffe}
\author[inst1]{Ian Reid}
\author[inst1]{Tat-Jun Chin}

\begin{abstract}
  LiDAR scanning for surveying applications acquire measurements over wide areas and long distances, which produces large-scale 3D point clouds with significant local density variations. While existing 3D semantic segmentation models conduct downsampling and upsampling to build robustness against varying point densities, they are less effective under the large local density variations characteristic of point clouds from surveying applications. To alleviate this weakness, we propose a novel architecture called HDVNet that contains a nested set of encoder-decoder pathways, each handling a specific point density range. Limiting the interconnections between the feature maps enables HDVNet to gauge the reliability of each feature based on the density of a point, \eg, downweighting high density features not existing in low density objects. By effectively handling input density variations, HDVNet outperforms state-of-the-art models in segmentation accuracy on real point clouds with inconsistent density, using just over half the weights.
\end{abstract}


\begin{keyword}
Semantic segmentation \sep 3D point clouds \sep Density variation \sep Large scale point clouds \sep Multi-resolution
\end{keyword}
\end{frontmatter}


\section{Introduction}\label{sec:intro}

\begin{figure*}[htp]\centering
\begin{subfigure}[b]{0.85\textwidth}\centering
\includegraphics[width=0.85\textwidth]{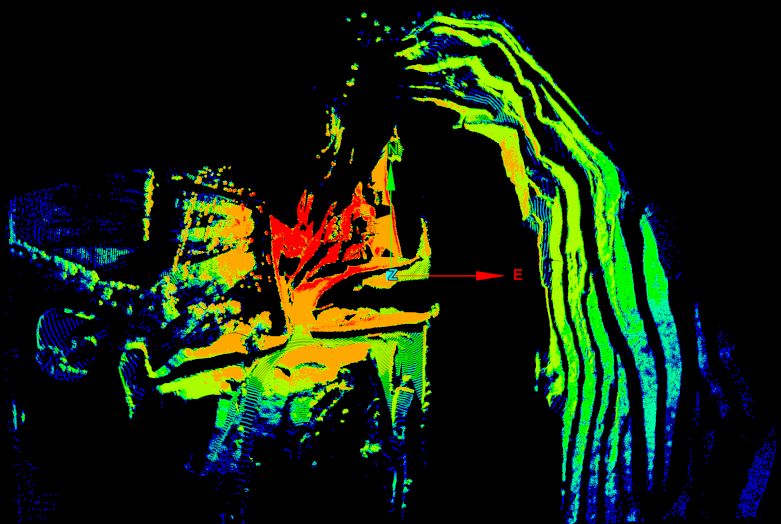}
\caption{BEV of an open pit mine acquired for surveying. Few obstructions result in a large scan area.}
\label{fig:single_mine}
\end{subfigure}
\begin{subfigure}[b]{0.85\textwidth}\centering
\includegraphics[width=0.85\textwidth]{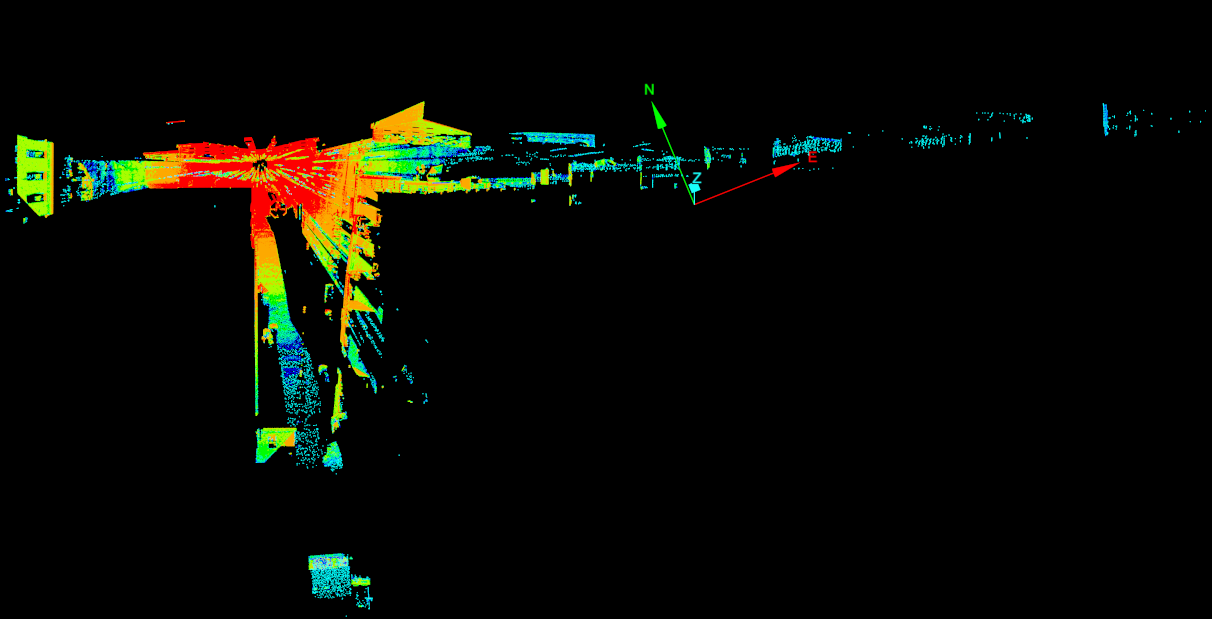}
\caption{BEV of an urban area acquired with terrestrial LiDAR as part of Semantic3D. Building obstructions limit coverage.}
\end{subfigure}
\begin{subfigure}[b]{0.85\textwidth}\centering
\includegraphics[width=0.85\textwidth]{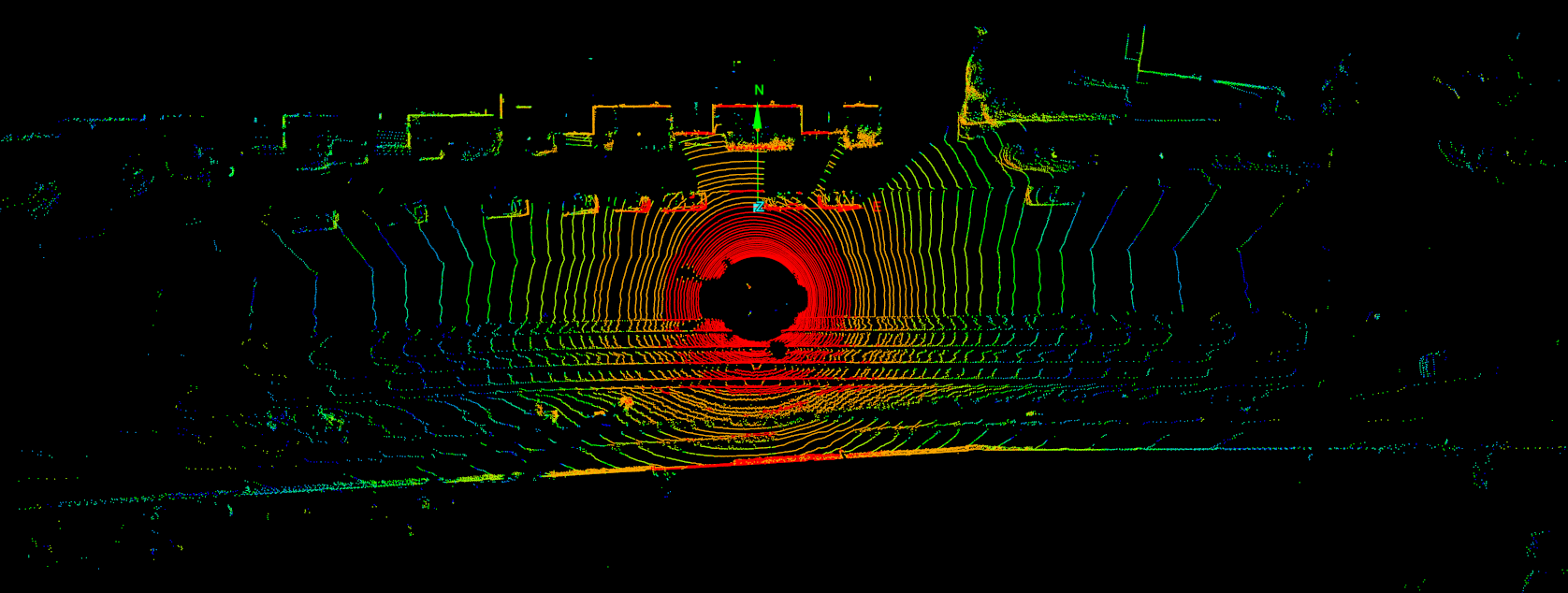}
\caption{BEV of a single street acquired with automotive LiDAR (KITTI).}
\end{subfigure}
\caption{Contrasting birds eye view (BEV) of different LiDAR scan types, high to low density represented by red to blue.}
\label{fig:ScanComparisons}
\end{figure*}

Light Detection and Ranging (LiDAR) devices generate accurate 3D measurements of their surroundings. While the generated point clouds have useful geometric information, practical application often requires semantic labels to be applied to the points. Recent progress of deep models in processing 3D point clouds~\citep{Survey,RecentOverviewSurvey,SurveyThatMentionsDensityVariationAsAChallengeButGivesNoExistingSolutions} has opened up many applications of LiDAR. In this paper, we focus on semantic segmentation of LiDAR scans~\citep{LiDARSegmentationSurvey}, \ie, assign each point a semantic label.

Many advances in point cloud semantic segmentation relate to autonomous driving, where the aim is the perception of the immediate surrounds of the vehicle~\citep{AutonomousDrivingReviewPaper}. Typically, automotive scans \citep{KITTI,NuScenes} do not extend much further than a 100~m; indeed, the hardware limitations of automotive LiDAR devices are such that scans reaching 250~m can be considered long range \citep{CIRRUS}. The low resolution scans (approximately 10\textsuperscript{5} points) have a fast collection rate, making them useful for time-sensitive problems such as obstacle avoidance. On the other hand, terrestrial LiDAR scans of surveying grade are slower, but of higher resolution, benefiting problems which require very high precision but not real-time solutions. 

One of the largest public datasets using a surveying-grade scanner, Semantic3D \citep{Semantic3D}, has high resolution scans of up to 10\textsuperscript{8} points, but only reaches physical dimensions as large as 240~m horizontally, and 30~m vertically. In comparison, terrestrial LiDAR scans such as those acquired in mining sites often have dimensions over a kilometre in the horizontal axes, and over 100~m vertically, covering a significantly larger area.

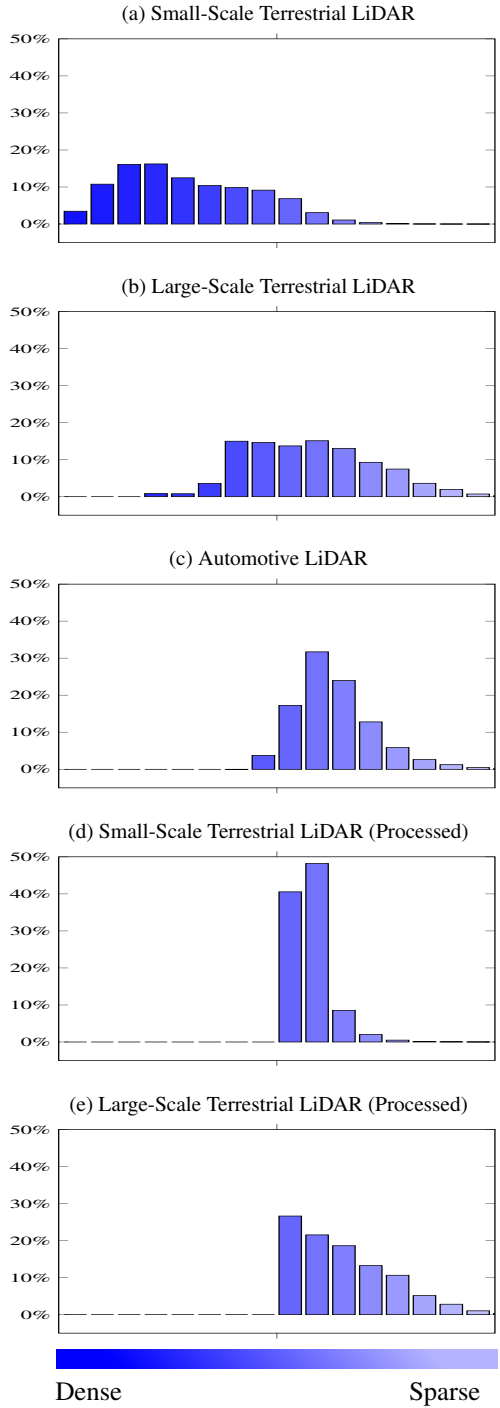
\begin{figure}
    \begin{subfigure}[b]{0.95\columnwidth}
    \caption{Small-Scale Terrestrial LiDAR}
    \begin{minipage}{0.95\columnwidth}
        \begin{adjustbox}{width=0.95\columnwidth,height=3cm}
            \begin{tikzpicture}
                \begin{axis}[
                    ybar,
                    ymax=50,
                    enlarge x limits=1,
                    legend style={at={(0.5,-0.15)},
                      anchor=north,legend columns=-1,},
                    ylabel=\empty,
                    symbolic x coords={Small-Scale Terrestrial},
                    xtick=data,
                    xticklabels =\empty,
                    yticklabel={$\pgfmathprintnumber{\tick}\%$}
                    ]
                \addplot[fill=Blue!100] coordinates {(Small-Scale Terrestrial,0.047905265)};
                \addplot[fill=Blue!93] coordinates {(Small-Scale Terrestrial,3.454456113)};
                \addplot[fill=Blue!90] coordinates {(Small-Scale Terrestrial,10.7618813)};
                \addplot[fill=Blue!87] coordinates {(Small-Scale Terrestrial,16.11515191)};
                \addplot[fill=Blue!84] coordinates {(Small-Scale Terrestrial,16.22632003)};
                \addplot[fill=Blue!80] coordinates {(Small-Scale Terrestrial,12.47479272)};
                \addplot[fill=Blue!76] coordinates {(Small-Scale Terrestrial,10.40677006)};
                \addplot[fill=Blue!71] coordinates {(Small-Scale Terrestrial,9.866319588)};
                \addplot[fill=Blue!65] coordinates {(Small-Scale Terrestrial,9.149059867)};
                \addplot[fill=Blue!60] coordinates {(Small-Scale Terrestrial,6.844395176)};
                \addplot[fill=Blue!55] coordinates {(Small-Scale Terrestrial,3.069345347)};
                \addplot[fill=Blue!50] coordinates {(Small-Scale Terrestrial,1.08769602)};
                \addplot[fill=Blue!45] coordinates {(Small-Scale Terrestrial,0.380245783)};
                \addplot[fill=Blue!40] coordinates {(Small-Scale Terrestrial,0.092420222)};
                \addplot[fill=Blue!35] coordinates {(Small-Scale Terrestrial,0.023240596)};
                \addplot[fill=Blue!30] coordinates {(Small-Scale Terrestrial,0.007073382)};
                \addplot[fill=Blue!25] coordinates {(Small-Scale Terrestrial,0.001760213)};
                \addplot[fill=Blue!20] coordinates {(Small-Scale Terrestrial,0.000459029)};
                \end{axis}
            \end{tikzpicture}
        \end{adjustbox}
    \end{minipage}%
    \end{subfigure}
    \begin{subfigure}[b]{0.95\columnwidth}
    \caption{Large-Scale Terrestrial LiDAR}
    \begin{minipage}{0.95\columnwidth}
        \begin{adjustbox}{width=0.95\columnwidth, height=3cm}
            \begin{tikzpicture}
                \begin{axis}[
                    ybar,
                    ymax=50,
                    enlarge x limits=1,
                    legend style={at={(0.5,-0.15)},
                      anchor=north,legend columns=-1,},
                    ylabel=\empty,
                    symbolic x coords={Large-Scale Terrestrial},
                    xtick=data,
                    xticklabels =\empty,
                    yticklabel={$\pgfmathprintnumber{\tick}\%$}
                    ]
                \addplot[fill=Blue!100] coordinates {(Large-Scale Terrestrial,0)};
                \addplot[fill=Blue!93] coordinates {(Large-Scale Terrestrial,0)};
                \addplot[fill=Blue!90] coordinates {(Large-Scale Terrestrial,0)};
                \addplot[fill=Blue!87] coordinates {(Large-Scale Terrestrial,0)};
                \addplot[fill=Blue!84] coordinates {(Large-Scale Terrestrial,0.857623021)};
                \addplot[fill=Blue!80] coordinates {(Large-Scale Terrestrial,0.786154533)};
                \addplot[fill=Blue!76] coordinates {(Large-Scale Terrestrial,3.575510769)};
                \addplot[fill=Blue!71] coordinates {(Large-Scale Terrestrial,14.97271353)};
                \addplot[fill=Blue!65] coordinates {(Large-Scale Terrestrial,14.63758064)};
                \addplot[fill=Blue!60] coordinates {(Large-Scale Terrestrial,13.71075044)};
                \addplot[fill=Blue!55] coordinates {(Large-Scale Terrestrial,15.10156952)};
                \addplot[fill=Blue!50] coordinates {(Large-Scale Terrestrial,13.05067163)};
                \addplot[fill=Blue!45] coordinates {(Large-Scale Terrestrial,9.277620147)};
                \addplot[fill=Blue!40] coordinates {(Large-Scale Terrestrial,7.456461216)};
                \addplot[fill=Blue!35] coordinates {(Large-Scale Terrestrial,3.611309155)};
                \addplot[fill=Blue!30] coordinates {(Large-Scale Terrestrial,1.972683975)};
                \addplot[fill=Blue!25] coordinates {(Large-Scale Terrestrial,0.724371529)};
                \addplot[fill=Blue!20] coordinates {(Large-Scale Terrestrial,0.264979897)};
                \end{axis}
            \end{tikzpicture}
        \end{adjustbox}
    \end{minipage}%
    \end{subfigure}
    \begin{subfigure}[b]{0.95\columnwidth}
    \caption{Automotive LiDAR}
    \begin{minipage}{0.95\columnwidth}
        \begin{adjustbox}{width=0.95\columnwidth, height=3cm}
            \begin{tikzpicture}
                \begin{axis}[
                    ybar,
                    ymax=50,
                    enlarge x limits=1,
                    legend style={at={(0.5,-0.15)},
                      anchor=north,legend columns=-1,},
                    ylabel=\empty,
                    symbolic x coords={Automotive},
                    xtick=data,
                    xticklabels =\empty,
                    yticklabel={$\pgfmathprintnumber{\tick}\%$}
                    ]
                \addplot[fill=Blue!100] coordinates {(Automotive,0)};
                \addplot[fill=Blue!93] coordinates {(Automotive,0)};
                \addplot[fill=Blue!90] coordinates {(Automotive,0)};
                \addplot[fill=Blue!87] coordinates {(Automotive,0)};
                \addplot[fill=Blue!84] coordinates {(Automotive,0)};
                \addplot[fill=Blue!80] coordinates {(Automotive,0)};
                \addplot[fill=Blue!76] coordinates {(Automotive,0)};
                \addplot[fill=Blue!71] coordinates {(Automotive,0.001638377)};
                \addplot[fill=Blue!65] coordinates {(Automotive,3.760895209)};
                \addplot[fill=Blue!60] coordinates {(Automotive,17.27750836)};
                \addplot[fill=Blue!55] coordinates {(Automotive,31.71898552)};
                \addplot[fill=Blue!50] coordinates {(Automotive,23.97847172)};
                \addplot[fill=Blue!45] coordinates {(Automotive,12.80719575)};
                \addplot[fill=Blue!40] coordinates {(Automotive,5.885051445)};
                \addplot[fill=Blue!35] coordinates {(Automotive,2.642702667)};
                \addplot[fill=Blue!30] coordinates {(Automotive,1.254177862)};
                \addplot[fill=Blue!25] coordinates {(Automotive,0.467756734)};
                \addplot[fill=Blue!20] coordinates {(Automotive,0.205616358)};
                \end{axis}
            \end{tikzpicture}
        \end{adjustbox}
    \end{minipage}%
    \end{subfigure}
    \begin{subfigure}[b]{0.95\columnwidth}
    \caption{Small-Scale Terrestrial LiDAR (Processed)}
    \begin{minipage}{0.95\columnwidth}
        \begin{adjustbox}{width=0.95\columnwidth, height=3cm}
            \begin{tikzpicture}
                \begin{axis}[
                    ybar,
                    ymax=50,
                    enlarge x limits=1,
                    legend style={at={(0.5,-0.15)},
                      anchor=north,legend columns=-1,},
                    ylabel=\empty,
                    symbolic x coords={Small-Scale Terrestrial (Processed)},
                    xtick=data,
                    xticklabels =\empty,
                    yticklabel={$\pgfmathprintnumber{\tick}\%$}
                    ]
                \addplot[fill=Blue!100] coordinates {(Small-Scale Terrestrial (Processed),0)};
                \addplot[fill=Blue!93] coordinates {(Small-Scale Terrestrial (Processed),0)};
                \addplot[fill=Blue!90] coordinates {(Small-Scale Terrestrial (Processed),0)};
                \addplot[fill=Blue!87] coordinates {(Small-Scale Terrestrial (Processed),0)};
                \addplot[fill=Blue!84] coordinates {(Small-Scale Terrestrial (Processed),0)};
                \addplot[fill=Blue!80] coordinates {(Small-Scale Terrestrial (Processed),0)};
                \addplot[fill=Blue!76] coordinates {(Small-Scale Terrestrial (Processed),0)};
                \addplot[fill=Blue!71] coordinates {(Small-Scale Terrestrial (Processed),0)};
                \addplot[fill=Blue!65] coordinates {(Small-Scale Terrestrial (Processed),0)};
                \addplot[fill=Blue!60] coordinates {(Small-Scale Terrestrial (Processed),40.5348284)};
                \addplot[fill=Blue!55] coordinates {(Small-Scale Terrestrial (Processed),48.19339507)};
                \addplot[fill=Blue!50] coordinates {(Small-Scale Terrestrial (Processed),8.545834742)};
                \addplot[fill=Blue!45] coordinates {(Small-Scale Terrestrial (Processed),2.00941646)};
                \addplot[fill=Blue!40] coordinates {(Small-Scale Terrestrial (Processed),0.50156168)};
                \addplot[fill=Blue!35] coordinates {(Small-Scale Terrestrial (Processed),0.146106009)};
                \addplot[fill=Blue!30] coordinates {(Small-Scale Terrestrial (Processed),0.048090223)};
                \addplot[fill=Blue!25] coordinates {(Small-Scale Terrestrial (Processed),0.014769088)};
                \addplot[fill=Blue!20] coordinates {(Small-Scale Terrestrial (Processed),0.005998323)};
                \end{axis}
            \end{tikzpicture}
        \end{adjustbox}
    \end{minipage}%
    \end{subfigure}
    \begin{subfigure}[b]{0.95\columnwidth}
    \caption{Large-Scale Terrestrial LiDAR (Processed)}
    \begin{minipage}{0.95\columnwidth}
        \begin{adjustbox}{width=0.95\columnwidth, height=3cm}
            \begin{tikzpicture}
                \begin{axis}[
                    ybar,
                    ymax=50,
                    enlarge x limits=1,
                    legend style={at={(0.5,-0.15)},
                      anchor=north,legend columns=-1,},
                    ylabel=\empty,
                    symbolic x coords={Large-Scale Terrestrial (Processed)},
                    xtick=data,
                    xticklabels =\empty,
                    yticklabel={$\pgfmathprintnumber{\tick}\%$}
                    ]
                \addplot[fill=Blue!100] coordinates {(Large-Scale Terrestrial (Processed),0)};
                \addplot[fill=Blue!93] coordinates {(Large-Scale Terrestrial (Processed),0)};
                \addplot[fill=Blue!90] coordinates {(Large-Scale Terrestrial (Processed),0)};
                \addplot[fill=Blue!87] coordinates {(Large-Scale Terrestrial (Processed),0)};
                \addplot[fill=Blue!84] coordinates {(Large-Scale Terrestrial (Processed),0)};
                \addplot[fill=Blue!80] coordinates {(Large-Scale Terrestrial (Processed),0)};
                \addplot[fill=Blue!76] coordinates {(Large-Scale Terrestrial (Processed),0)};
                \addplot[fill=Blue!71] coordinates {(Large-Scale Terrestrial (Processed),0)};
                \addplot[fill=Blue!65] coordinates {(Large-Scale Terrestrial (Processed),0)};
                \addplot[fill=Blue!60] coordinates {(Large-Scale Terrestrial (Processed),26.62183861)};
                \addplot[fill=Blue!55] coordinates {(Large-Scale Terrestrial (Processed),21.53386271)};
                \addplot[fill=Blue!50] coordinates {(Large-Scale Terrestrial (Processed),18.60941478)};
                \addplot[fill=Blue!45] coordinates {(Large-Scale Terrestrial (Processed),13.22928708)};
                \addplot[fill=Blue!40] coordinates {(Large-Scale Terrestrial (Processed),10.63243207)};
                \addplot[fill=Blue!35] coordinates {(Large-Scale Terrestrial (Processed),5.149493596)};
                \addplot[fill=Blue!30] coordinates {(Large-Scale Terrestrial (Processed),2.812919929)};
                \addplot[fill=Blue!25] coordinates {(Large-Scale Terrestrial (Processed),1.032907012)};
                \addplot[fill=Blue!20] coordinates {(Large-Scale Terrestrial (Processed),0.377844218)};
                \end{axis}
            \end{tikzpicture}
        \end{adjustbox}
    \end{minipage}%
    \end{subfigure}
    \begin{subfigure}[t]{0.85\columnwidth}
        \hspace*{0.85cm}
        \begin{tikzpicture}
            \node [shading = axis,rectangle, left color=Blue, right color=Blue!30!white,shading angle=135, anchor=north, minimum width=0.875\columnwidth, minimum height=0.25cm] (box) at (current page.north west){};
        \end{tikzpicture}
        \newline
        \hspace*{0.9cm}Dense \hspace{3.6cm} Sparse
    \end{subfigure}
\caption{Proportion of scan for different density groups. As high resolution LiDAR is downsampled during preprocessing, distribution after is also shown. Our preprocessing reduces the terrestrial LiDAR to one point every 6cm, or approximately 1100 points per metre.}

\label{fig:density_histogram}
\end{figure}

LiDAR scans of a physically larger scale tend to suffer from high density variations; see \cref{fig:ScanComparisons,fig:density_histogram}. Fundamentally, fewer nearby occlusions yield more scan points further from the scanner, where density is lower. While not to the same extent as surveying-grade scans, the inherently lower resolution and distance limitations of automotive LiDAR cause it to also have density variation even in urban environments. 




State-of-the-art 3D semantic segmentation methods~\citep{LiDARSegmentationSurvey} struggle on large-scale surveying point clouds, due to the higher density variation. In particular, while the methods which operate directly on point clouds ~\citep{PointNet++,Randlanet,MonteCarlo,ConvPoint,RSCNN} extract local features in a multi-scale manner through down- and up-sampling, details of how to best propagate and utilise features of different scales are left to the neural network to learn. Some do combat density variation, however they only target variation within the scope of individual feature extraction steps and not across the entire network architecture.




\paragraph{Density variation vs class imbalance}

It is vital to contrast density variation and class imbalance, both related factors that influence segmentation accuracy. Classes with fewer point samples tend to be smaller objects with lower point density. While this is an important challenge to tackle, our focus in this paper is the effects of density variation \emph{independent of} the population size of the class. A single class can appear in a point cloud with each instance having vastly different densities. A wall close to the scanner for example, will have a higher density of points than one far away; see \cref{fig:density_histogram} for density distributions of different LiDAR types.

\vspace{-1em}
\paragraph{Contributions}

We highlight the importance of effectively accounting for local density variations in semantic segmentation on 3D point clouds, particularly those acquired from real-world surveying tasks. To this end:
\begin{itemize}[leftmargin=1em,itemsep=0pt,parsep=0pt,topsep=2pt]

\item We propose \textbf{HDVNet} (\textbf{h}igh \textbf{d}ensity \textbf{v}ariation \textbf{net}work), a point cloud segmentation model that contains a nested set of feature extraction pipelines, each handling a specific input local density; see \cref{fig:proposedmodel}. Interactions between the pipelines is tightly controlled to exploit potential correlations between density levels. An aggregation layer applies attention scores to the features accordingly, such that low density objects are not classified based on (potentially non-existent) high resolution features, while higher density points remain able to take advantage of their fine features.

\item We collected a new dataset, named \textbf{HDVMine}, that consists of LiDAR scans from open-cut mines to evaluate our ideas. Our point cloud scans cover geographic areas which are kilometres in scale, making them larger than existing terrestrial LiDAR datasets \citep{Semantic3D}. A single scan is comparable in scale to an automotive LiDAR drive's frames combined. In addition, existing datasets comprise of ``above-ground'' scenes where there is a single and consistent ground plane. In contrast, an open-cut mine can have multiple physical tiers, with complex structures embedded therein.

\end{itemize}
As we will show in Sec.~\ref{sec:results}, HDVNet yields up to $6.7$ percentage points higher accuracy in semantic segmentation on our dataset, compared to a state-of-the-art point-cloud models~\citep{Randlanet} despite HDVNet using almost half as many weights.

\section{Related work}



Point clouds have useful geometric information for each point, but the lack of any inherent structure to the data makes local context difficult to determine. We first survey existing methods for point cloud segmentation, from those that preprocess the point cloud to alternative representations, to those which directly take the raw point cloud as input.

\newpage

\subsection{Grid-based methods}


Many point cloud networks take inspiration from image-processing techniques. Unlike a pixel image however, a point cloud has no inherent grid structure. For the purpose of using convolutions and similar techniques on the point cloud, a common step is first converting from points to a grid-based representation. These representations include two-dimensional pixel images \citep{MultiView,MVPN}, a birds eye view of the scene \citep{Birdseye,PillarSeg,PointPillars}, or a three dimensional voxel grid \citep{ModelNet,Voxnet}.

Large sections of empty space in the scene lead to poor memory scaling in grid representations. Data structures such as octrees \citep{EfficientProcessing,efficientImplementation,OctNet,O-CNN,sphereOct} avoid wasting memory on empty space, but information is still lost where multiple points are combined into a single voxel. These grid structure representations have demonstrated particular success for low-resolution LiDAR scans where there are less fine details to be lost. State of the art methods for such scans range from modified forms of three-dimensional voxel structures \citep{SphericalTransformer,Cylinder3D} to representing the scan in two dimensions such as with a  Range Image \citep{RangeFormer}.

\subsection{Point Based Methods}

Convolutions are performed on grid structures, which makes operating directly on the raw point cloud data difficult. A raw point cloud is simply a set of points, with no consistent ordering. PointNet \citep{PointNet} is a pioneering work in directly processing point clouds, which demonstrated the success of using network layers with Multi-Layer Perceptrons (MLPs). Each MLP is limited to operate only on individual points (with shared weights), and any operations performed on the entire point cloud being order-invariant and low-cost such as max-pooling. More research rapidly followed, extending it directly such as PointNetLK \citep{PointNetLK} and PointNet++ \citep{PointNet++}, or developing new algorithm using MLPs as a base. 

These alternative point processing methods are designed to better utilise the local relationship between points in the scene. RandLA-Net by Hu \etal \citep{Randlanet} does this using K-Nearest Neighbours and MLPs to aggregate features for each point which represent the local neighbourhood. Like other MLP based methods, it is very efficient, scales well to large point clouds, and uses an encoder-decoder structure to get features from multiple scales. 

An alternative approach is to apply convolutions to the raw point cloud as if it had a more grid-like structure. This requires modifying the implementation of a convolution \citep{ConvPoint, RSCNN, PointCNN} to apply to unordered points. One example of this is assigning coordinates to the convolution kernel, and using a MLP to determine how much each kernel weight affects a point based on the point's relative position to the kernel \citep{ConvPoint,RSCNN}. This contrasts to a traditional grid-structure kernel where each weight fully affects the value in one specific pixel or voxel co-ordinate and no others.

\subsection{Coarse, then fine processing}
 Raw point clouds have limited features for each point (\eg x,y,z,r,g,b), lacking any local context. To account for this, some networks generate useful features first. Taeo \etal~\citep{CGANet} have a network identify which points belong to distinct objects, before then classifying each point with semantic labels. Multi-pass approaches to first identify edges \citep{HierarchicalPoint-EdgeInteractionNetwork} or narrow down areas of interest before more fine processing \citep{FrustrumPointNets,CoarseFineKidney} are also common. Others such as Varney \etal~\citep{PyramidPoint} and Li \etal~\citep{GeometryAttentional} first extract fine features before downsampling to a sparser point cloud as usual, but then go back and do so a second time after the point cloud's coarse features have been extracted. These methods all assume that fine features exist when extracting and propagating them, which does not hold when scan's density is inhomogeneous.

An alternative approach is to perform coarse segmentation into ``superpixels'' or ``simple objects'', followed by a graph-based approach \citep{SPG}. Such graph-based networks do not scale well to large and complicated scenes. In addition, coarse segmentation which quickly identifies the ground points \citep{GroundNotGround} or the edges of the road \citep{curbDetection}, relies on assumptions such as ``the lowest points detected are the ground'' which do not hold in contexts such as mining.

\subsection{Dealing with density variation}

Consider a point cloud of $N$ points $\cP=\{ \bp_i \}^{N}_{i=1}$. The density of the point cloud as a whole is the ratio of points $N$ to the volume occupied by the point cloud. For each given point $\bp_i$, we define its local density $\rho_i$ using the density of its immediate neighbourhood of $K$ nearby points $\cN_i$, where $\cN_i = [\cN_{i,1}, \cN_{i,2}, ... \cN_{i,K}]$. Many existing methods inherently assume a homogenous density, such that the local density $\rho_i$ of any point is roughly the same as the average density of the entire point cloud. As shown previously in \cref{fig:density_histogram} this is not always the case, the local density of points can vary greatly.

In both our method, and many existing pointcloud networks, the local neighbourhood of a point $\bp_i$ is determined using the  K-Nearest neighbours, with $K$ a fixed hyperparameter, $K \in \mathbb{Z}$. This creates a receptive field around each point, the $K$ points within making up the local neighbourhood. Each time the point cloud is downsampled, it becomes more sparse, enabling the receptive field to grow in physical size. This enables early network blocks with small receptive fields to extract fine object features, while later blocks extract sparser features using larger receptive fields. These receptive fields encounter issues when density throughout the point cloud is inhomogeneous.

\begin{figure}[ht]\centering
\includegraphics[width=0.45\textwidth]{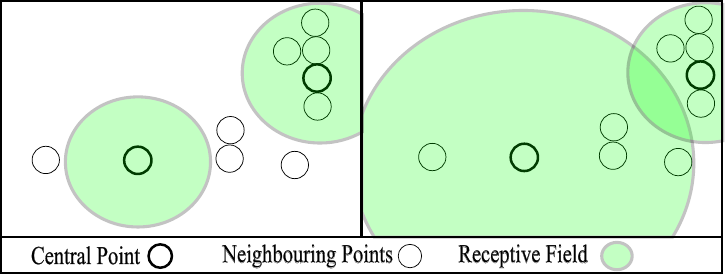}
\caption{Two variations of the receptive field when aggregating local information around a point. On the left, a fixed size field, with one receptive field providing no useful information. On the right, a fixed number of $K$ = 4 neighbours, resulting in two receptive fields of vastly different sizes being used at the same local-feature-aggregation step}
\label{fig:ReceptiveFields}
\end{figure}

Objects which exist far away from the scanner or near-parallel to the laser will appear in the initial scan with a low point density. Early layers cannot extract useful high-density features when the point's local neighbourhood is sparse to begin with. This causes one of two density-variation issues, depending on whether the receptive field uses a fixed number of neighbours $K$, or a fixed radius. If $K$ is fixed, the network layers are required to learn how to extract useful information from a wide variety of receptive field sizes, all using the same shared weights. Alternatively, if the physical size of the receptive field is fixed, then the neighbourhood feature will sometimes be generated from no neighbouring points at all.  \cref{fig:ReceptiveFields} visualises this issue. In HDVNet, we fix the number of neighbours $K$, and then take further steps to counter the issue of inconsistent receptive field size, detailed in \cref{sec:methodology}.

Existing methods do not address density variation across the entire network like our HDVNet does, but they do take steps to limit the effect on individual network layers \citep{Dance-Net,PointNet++, MonteCarlo}. Alternative point cloud representations such as voxels tackle density by either weighting each voxel based on how many points it has \citep{PointDensityAwareVoxels}, or implement a minimum density floor, ignoring sparse sections entirely \citep{DAPNet}.

The reliance on high density features can be addressed by first aggregating high density points together to represent the scene in a more homogeneous, coarse manner \citep{HierarchicalClusteringAKAOctree,TriangleNet}. Such approaches inevitably result in information loss as higher-density sections are downsampled to achieve consistent density, although performance on low density objects does improve.

\begin{figure*}[htb!]\centering

\includegraphics[width=0.99\textwidth]{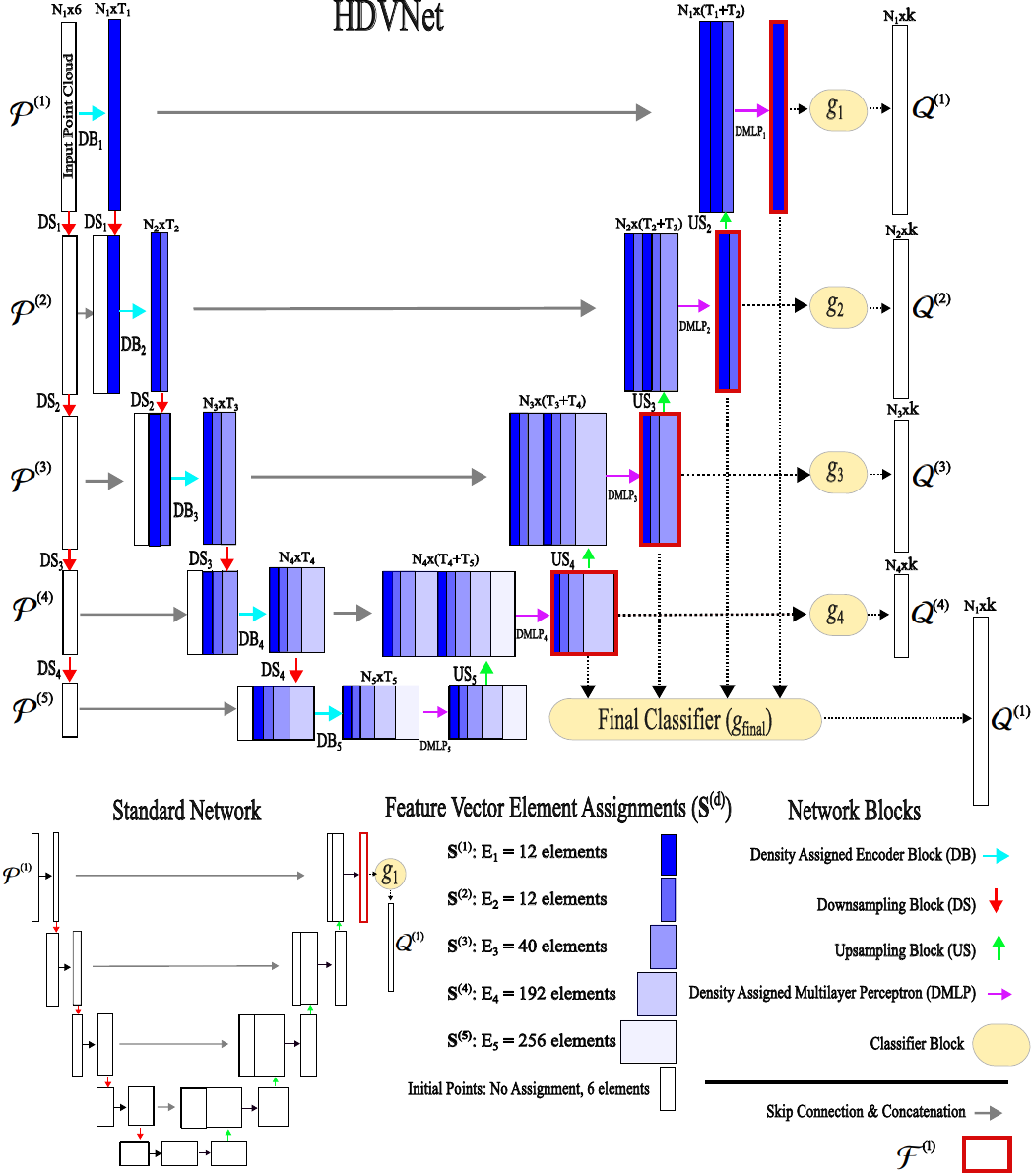}
\caption{Proposed point cloud segmentation model HDVNet. Feature element assignments are visualised at each step of the architecture as shades of blue. The number of elements $E_d$ for each subsection $S^{(d)}$ are those used in our experimental setup. Final features $\cF^{(a)}$ are extracted and passed to four classifiers $g$ during initial training, and a single classifier $g_{final}$ during fine tuning. Details of the classifiers are shown later in \cref{sec:InitialTraining} and \cref{sec:FineTuning}. A standard U-net style network architecture for point cloud segmentation is shown in bottom left for comparison. Similar existing networks \citep{Randlanet,PointCNN} use this style, which does not include our density assignments, re-submission of the original point cloud, or multiple classifiers.}
\label{fig:proposedmodel}
\end{figure*}


\section{High density variation network - HDVNet}
\label{sec:methodology}


HDVNet is an architecture which processes a point cloud of $N$ points $\cP=\{ \bp_i \}^{N}_{i=1}$. The raw point values $\bp_i$ initially passed to the network are $[x_i,y_i,z_i,r_i,g_i,b_i]$, where $x,y,z$ are the point's spatial co-ordinates, and $r,g,b$ are the colour values.

The number of points $N$ varies throughout the network as shown in \cref{fig:proposedmodel}. Each Downsampling Block $DS$ removes points, subsampling the point cloud from one density state $d$ to a sparser density $d+1$, where $d \in \{1,2,3,4,5\}$. The density state of the initial point cloud being $d=1$. Formally, $DS_d$ takes $N_d$ points as input, and returns the smaller subset of $N_{d+1}$ points, such that $\{ \bp_j \}^{N_{d+1}}_{j=1} = DS_d(\{\bp_i \}^{N_{d}}_{i=1})$. We index the pointcloud based on how downsampled it is, with the initial point cloud being $\cP^{(1)}=\{ \bp_i \}^{N_1}_{i=1}$, and the most downsampled being $\cP^{(5)} =\{ \bp_j \}^{N_5}_{j=1}$ such that $\cP^{(d+1)} \subseteq \cP^{(d)}$. The number of points at each state $\cN=\{ N_d \}^{5}_{d=1}$ is set as a hyperparemeter. We index upsampling and downsampling blocks using their input pointcloud's density state $d$, for example the upsampling block $US_5$ upsamples the pointcloud from $N_5$ to $N_4$ points (from $\cP^{(5)}$ to $\cP^{(4)}$).

Each point $\bp_i$ has a corresponding feature vector $\bF_{i}$, containing a total of $T$ elements such that $\bF_{i} \in \mathbb{R}^{T}$. Unique to HDVNet, each feature vector $\bF_{i}$ can be separated into assigned subsections $\bS^{(d)}_{i} \subseteq \bF_i$, where the vector elements of each subsection are ``assigned'' to a corresponding density state $d$. Each of these subfeature vectors contains $E_d$ elements, where $E_d \in \mathbb{Z}^+$. The set of integers $\{E_d\}^5_{d=1}$ is defined as a hyperparameter, and is constant throughout the network. In \cref{fig:proposedmodel}, we visualise each feature vector subsection $\{\bS^{(d)}_i\}^{N_d}_{i=1}$ as different shades of blue.

 Each Density Assigned Encoder Block ($DB$) adds a new subsection $\bS^{(d)}$ of $E_d$ elements to each point's corresponding feature vector. We use the number of assigned subsections $a$ to index each feature vector $\bF^{(a)}_i$, though the network, initialising as $\bF^{(0)}_{i}$ with no assignments and no elements. For example, $\bF^{(3)}_{i}$ has the subsections $\bS^{(1)}_i, \bS^{(2)}_i, \bS^{(3)}_i$ and a total of $E_1+E_2+E_3$ elements. The total number of elements in a given $\bF^{(a)}_i$ is therefore $T_a$, such that $T_a = \sum^{a}_{d=1}E_d$.

We also index $DB_a$ blocks using the number of assigned subsections they involve. Each $DB_a$ takes as input both the existing feature vectors $\{\bF^{(a-1)}_{i}\}^{N_d}_{i=1}$ and the original point values, with a formal definition of 

\begin{equation}
\{ \bF^{(a)}_{i} \}^{N_d}_{i=1} = DB_a(\{ \bF^{(a-1)}_{i} \}^{N_d}_{i=1}, \{ \bp_i \}^{N_d}_{i=1})
\end{equation}

\noindent The first DB block $DB_1$ takes only the original point values as input, with every $\bF^{(0)}_{i}$ being empty. Details on density states $d$ are given in \cref{sec:DensityGroupsAndGroupSets}, while the effect of subsection ``assignments'' $S^{(d)}_i$ are provided in \cref{sec:DensityEncoderBlock} and \cref{sec:InitialTraining}.


As shown in \cref{fig:proposedmodel}, HDVNet maps the the initial point cloud $\cP^{(1)}$ to four final feature vectors $\{ \cF^{(a)} \}^{4}_{a=1}$. During training each is passed to one of four classifiers $\{ g_a(\cF^{(a)})\}^4_{a=1}$. Each classifier maps the feature vectors to four sets of probability distributions $\{\cQ^{(d)}\}^4_{a=1}$. Each $\cQ^{(d)}$ has $N_d$ distributions, and is created using the corresponding classifier $g_a$, such that $a=d$. We define $\cQ^{(d)} = \{ \tilde{\bq}_i \}^{N_d}_{i=1}$ where $\tilde{\bq}_i = [\tilde{q}_{i,1}, \tilde{q}_{i,2}, \dots, \tilde{q}_{i,k}]$, such that $\tilde{q}_{i,k}$ is the estimated confidence that the corresponding point $\bp_i$ is of semantic class $k$. After a fine tuning step this is simplified to a single classifier $\cQ^{(final)}=\{ \tilde{\bq}_i \}^{N_1}_{i=1} = g_{final}(\{ \cF^{(a)} \}^{4}_{a=1})$, used in inference and shown in \cref{fig:proposedmodel} as the ``Final Classifier''.

\subsection{Density Groups and Density States}
\label{sec:DensityGroupsAndGroupSets}

Our solution to high density variation is to make point density a central part of the network architecture. To do so, HDVNet relies on three different measures of a single point $\bp_i$'s local density. The continuous density estimate $\rho_i$, which is quantised evenly into discrete groups $\delta$, and unevenly into the density states $d$. 

Density is not a native measure from LiDAR, so the estimate comes from the the K-nearest neighbours. A sphere around each point is made, with the radius $r$ being the euclidean distance from the point to the most distant of its $K$ neighbours. Dividing $K$ by the volume creates a density estimate $\rho_i$ in points per $m^3$

\begin{equation}    
\rho_i = \frac{K}{\frac{4}{3}\pi r^3}
\end{equation}

\noindent This was chosen as a computationally efficient way of estimating a point's volume, with $\cN_i$ already being calculated in order to generate a point's local features. The point clouds in \cref{fig:ScanComparisons} were coloured using $\rho_i$, so can be referred to for a visualisation of the density estimate.

While $\{ \rho_i \}^{N_d}_{i=1}$ is a useful estimate of every point's density, it is often too continuous in nature for steps in HDVNet which expect a more discrete input. For this purpose, the points are quantised into discrete grouping buckets $\delta \in \mathbb{Z}$, with $\delta=0$ being the first group, $\delta=1$ the second, and so forth. The distributions in \cref{fig:density_histogram} were created using these groups $\delta$.

In our experimental setup the initial grouping $\delta=0$ was set to the very high density threshold of $\rho_i > 2\times10^6$ points per $m^3$. This value was chosen to ensure that very few points fall into the first grouping, with even high-density small-scale urban scans such as those in Semantic3D having very few points over this threshold. The lower threshold $t$ of the density grouping $d=0$ is thus $t_0=2\times10^6$, and subsequent thresholds $t_\delta$ are calculated to ensure that the minimum density (in m$^{-3}$) is consistently a quarter that of the prior grouping. Each threshold is thus calculated as:

\begin{equation}    
t_\delta = \frac{t_{\delta-1}}{4}
\end{equation}

\noindent For better reflection of the point cloud's density throughout the network, these quantised groups $\delta$ are then combined into larger density states $d \in \{0,1,2,3,4,5\}$. Each $d$ is the point cloud's estimated density state at a given point in the network. There are more density groups $\delta$ than there are downsampling blocks $DS$, so each $DS$ reduces the point cloud's maximum density by multiple groups $\delta$ at a time.


We set which contiguous groups $\delta$ make up each density state $d$ by analysing the density distribution across all the points in the training dataset. With this average distribution we estimate how many points $N$ will remain after downsampling to each density group $\delta$. As the target number of points for each density state $\{ N_d \}^{5}_{d=1}$ is a known hyperparameter, we use the density group which results in the closest number of remaining points to the target. 

\begin{figure}[ht]\centering
\includegraphics[width=0.45\textwidth]{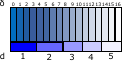}
\caption{Each density state $d$ is comprised of multiple groups $\delta$}
\label{fig:DensityGroups}
\end{figure}

Downsampling to a threshold $t_d$ therefore results in approximately $N_d$ points remaining, $t_d = t_\delta \mid \{\rho_i\}^{\approx N_d}_{i=1} >= t_\delta$. An example of how contiguous density groups $\delta$ are combined into a state $d$ is shown in \cref{fig:DensityGroups}.

A key point to clarify is a point being within a specific density state of overall pointcloud $\bp_i \in \cP^{(d)}$ as opposed to a point's inherent density state. The initial pointcloud $\cP^{(1)}$ for example contains all the input points, regardless of how sparse they are. For when we refer exclusively to the subset of points with an inherent density $\rho_i$ between \textit{both} that density state's lower threshold $t_d$, and the prior state's threshold $t_{d-1}$ we define the subset as 
\begin{equation}    
I^{(d)} := \Set{\bp_i}{t_d < \rho_i <= t_{d-1}}
\end{equation}

\begin{figure}[ht]\centering
\includegraphics[width=0.45\textwidth]{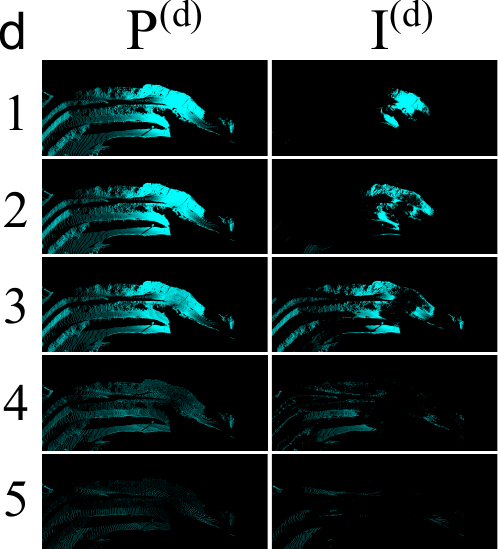}
\caption{A visualisation of $P^{(d)}$ and $I^{(d)}$ using the same hyperparameters as our experiments on HDVMine. Each $DS_d$ subsamples the entire pointcloud, while $I^{(d)}$ contains only points with a matching inherent density. All points are coloured bright blue for visual clarity.}
\label{fig:PdvsId}
\end{figure}

\noindent A visualisation of $P^{(d)}$ and $I^{(d)}$ is shown in \cref{fig:PdvsId}

\begin{figure*}[ht]\centering
\includegraphics[width=0.99\textwidth]{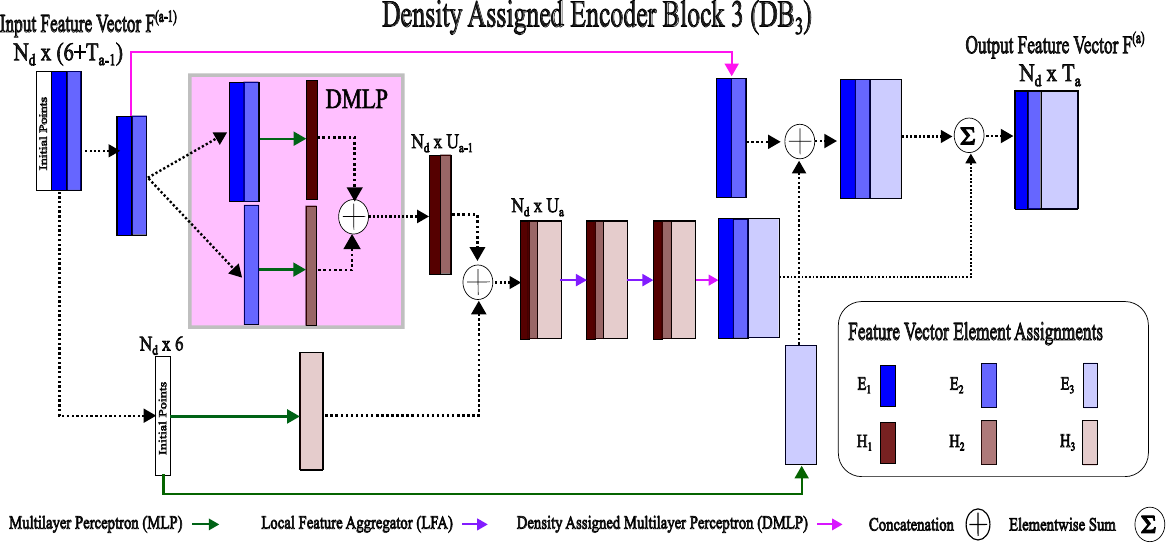}
\caption{Detailed example of the density assigned encoder block $DB_3$, wherein features assigned to three different density states $d$ are all processed. $N_d$ and $(6+T_{l-1})$ are the input dimensions. The feature vector elements assigned to $a=1$ and $a=2$ are propagated from prior layers, and $a=3$ assigned elements are newly added in the block. Two Local Feature Aggregation (LFA) blocks are used in addition to a skip connection. A box is drawn around the series of steps which together form an example DMLP, while a second DMLP is left as a single arrow to reduce visual clutter.}
\label{fig:DensityAssignedDecoderBlock}
\end{figure*}


\subsection{Density assigned encoder block (DB)}
\label{sec:DensityEncoderBlock}


Our method's key architecture modification is the assignment of feature elements to density states, as visualised in  \cref{fig:proposedmodel,fig:DensityAssignedDecoderBlock,fig:DensityConnectedLayer,fig:FinalClassifier} as shades of blue for the $E_d$ elements of each subsection $\bS^{(d)}$. For our feature extraction blocks, we use our novel density assigned encoder block (DB). Many of our multilayer perceptrons (MLP) and fully connected layers (FC) are also replaced with our density aware MLP (DMLP) and density connected layer (DC).

Some of the operations performed in the density assigned encoder block's hidden layers have higher memory requirements. As shown in \cref{fig:DensityAssignedDecoderBlock} we reduce the number of elements in each subsection $\bS^{(d)}_i$ from $E_d$ to $H_d$, with $H_d$ also set as a hyperparameter. We visualise feature vector subsections with $H_d$ elements as shades of maroon instead of blue. As the total number of elements normally is $T_a = \sum^{a}_{d=1}E_d$, we define the total number of elements in these hidden layers as $U_a = \sum^{a}_{d=1}H_d$.

\subsubsection{Continued input of original scene}
One addition is the reintroduction of the original raw point values after every downsampling as shown in the bottom left corner of \cref{fig:DensityAssignedDecoderBlock}. Each DB block requires not only the input point feature vector $\bF^{(a-1)}$, but also the raw point values of $\bp_i$.


\begin{equation}    
\{ \bF^{(a)}_{i} \}^{N_d}_{i=1} = DB_a(\{ \bF^{(a-1)}_{i} \}^{N_d}_{i=1}, \{ \bp_i \}^{N_d}_{i=1})
\label{eq:DBDefinition}
\end{equation}

\noindent This enables sparser feature extraction using original point data instead of relying on unreliable propagated features from higher densities. The features in HDVNet assigned to lower density states, are in this way made robust to the original density of the object.

\subsubsection{Density assigned multi-layer perceptrons}
\label{sec:DMLP}

To prevent reliance on higher density features, and enforce the ``assigned subsections'' $\bS^{(d)}_i$ of a feature vector, element separation is applied within the encoder blocks. A normal MLP or FC would treat all feature elements the same, so we instead use our density assigned MLP (DMLP) or Density Connected Layer (DC). They both operate on feature vector elements without mixing information across density assigned subsections. For reference, we define a standard multi-layer perceptron (MLP) or fully connected layer (FC) as a mapping from $\bF^{in}_i$ features to $\bF^{out}_i$. The difference being that a MLP also includes layernorm (LN) and activation (AVN) layers:

\begin{align}   
\bF^{out}_i & = FC(\bF^{in}_i) \\
\bF^{out}_i & = MLP(\bF^{in}_i) = AVN(LN(FC(\bF^{in}_i)))
\label{eq:MLPDefinition}
\end{align}

\noindent In HDVNet, feature elements are each assigned either to the current point cloud's density state $d$ or a previous one ($d$-1, $d$-2, etc). During feature extraction and processing, such as DMLPs, we allow higher density feature elements to use elements assigned to lower densities as input, but not vice-versa. This rule comes from the fact that an object with high density information will always have low density information once it is downsampled, but the same does not necessarily hold true in reverse. An object which is sparse to begin with will not have any useful high density information to be considered. 

The number of feature vector elements assigned to a specific density is $E_d$. The elements of a feature vector $\bF^{(a_{in})}_i$ assigned to any of a contiguous set of subsections $\bS_{start}$ to $\bS_{end}$ is referred to as $\bF_{i,start:end}$. Multiple MLPs each viewing different subsections of a point's features $\bF^{(a_{in})}_i$ are thus combined into a Density Assigned MLP (DMLP), along with the number of element assignments to include in the output feature vector $a_{out}$:

\begin{align}
      \nonumber DMLP(\bF^{in}_i, a_{out}) = & MLP(\bF_{i,1:a_{in}},E_1) \\ 
      & \nonumber  \oplus MLP(\bF_{i,2:a_{in}},E_2)  
      \\
      & \nonumber \vdots \\ 
      & \oplus MLP(\bF_{i,a_{out}:a_{in}},E_{a_{out}}),
      \label{eq:DMLP}
\end{align}

\begin{figure}[htb!]
    \includegraphics[width=0.45\textwidth]{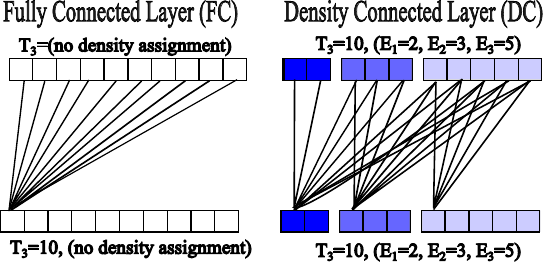}   
    \caption{A visual representation of a density connected layer ($DC_3$). For simplicity, a feature vector with only ten elements $T_3 = 10$ is shown. On the left is a traditional FC with no assignments. On the right is an example where the elements are split among three density assigned subsections, with a $E_1,E_2,E_3$ being $2,3,5$ for both the input and output. Weights are only visualised for the first feature of each subsection for clarity. A DMLP has additional layernorm and activation layers, but follows the same density preservation rules.}\label{fig:DensityConnectedLayer}
\end{figure}

\newpage

\noindent Where $\oplus$ is the concatenation of feature vectors, and $a_{out} <= a_{in}$. This creates a feature extractor which is robust to density variation, yet still extracts fine features. A pink square is drawn around this step in \cref{fig:DensityAssignedDecoderBlock}. As sparse features are generated without using fine ones, they can be trusted to be robust to density variation. Our density connected layers (DC) follow the same method, stacking fully connected layers (FC) to preserve density assignments. \cref{fig:DensityConnectedLayer} is an example of this for $d=3$, and similar to a DMLP, a DC can be defined as:

\begin{align}
      \nonumber DC(\bF^{in}_i, a_{out}) = & FC(\bF_{i,1:a_{in}},E_1) \\
       \nonumber & \oplus FC(\bF_{i,2:a_{in}},E_2) \\
      & \nonumber \vdots \\ 
      & \oplus FC(\bF_{i,a_{out}:a_{in}},E_{a_{out}}),
      \label{eq:DC}
\end{align}

\begin{algorithm}
\caption{Density Assigned MLP}\label{alg:DMLP}
\begin{algorithmic}
\Require{Feature Vector $\bF^{{in}}_i$, output's number of assigned subsections $a_{out}$} 
\Ensure{Feature Vector $\bF^{out}_i$}
\State{$\bF^{out}_i \gets empty \: tensor$}
\ForEach{density state $d$, $d < a_{out}$}
{
    \State{$ConcatenatedF \gets concatenate(\bS^{(d)}_i,...\bS^{(a_{in})}_i)$}
    \State{$\bS^{propagated}_i \gets MLP(ConcatenatedF,E_{d})$} 
    \State{$\bF^{out}_i \gets concatenate(\bF^{out}$, $\bS^{propagated}_i$)}
\EndFor
}
\State{Return $\bF^{out}_i$}
\end{algorithmic}
\end{algorithm}


\noindent Our local testing confirmed that as found in other works \citep{RooftopDensityAnalysisShowsHighResAddsLittleBenefit} low level features are enough for the majority of the analysis with the benefit of features continually decreasing as they become finer. For this reason all new feature vector elements added to a point cloud of density state $d$ are assigned to the new subsection $\bS^{(d)}_i$, maximising the number of features assigned to lower densities. This is visualised in \cref{fig:proposedmodel} and \cref{fig:DensityAssignedDecoderBlock} by the width of subsections remaining constant throughout.

\subsection{LiDAR-grid subsampling}
\label{sec:Lidar-gridsubample}

\begin{algorithm}
\caption{LiDAR retaining subsampling}\label{alg:LiDARSubSample}
\begin{algorithmic}
\Require{Point cloud with $N$ points $\cP$, and either target density grouping $\delta_t$ or target number points $N_t$}
\Ensure{Subsampled Point cloud $\cP_s$}
\If {target is $N_t$}
    \State{$\delta_t \gets$ highest $\delta$ grouping in training dataset}
\EndIf
\ForEach{Density grouping $\delta$, 0:$\delta_t$}
{
    \ForEach {$\bp_i \in \mathcal P $}
    {
        \State{current point's density grouping is $\delta_p$}
        \State {$\Delta \delta \gets \delta_t - \delta_p$}
        \If {$c \mod 2^{\Delta \delta} \neq 0$ or $r \mod 2^{\Delta \delta} \neq 0$}
        \State{Remove $\bp_i$ from $\cP$, add to point set $\cP_\delta$}
        \EndIf
    }
    \EndFor
}
\EndFor
\State{$\cP_s \gets empty set$}
\If{target is $N_t$}
\State{$\delta \gets $ highest d in training data}
\While{$size(\cP_s) \leq N_t$}
\State{concatenate($\cP_s$, $\cP_\delta$)}
\State{$\delta \gets d - 1$}
\EndWhile
\State{$N_{required} \gets N_t - size(\cP_s)$}
\State{$\cP_{\delta(random)} \gets RANDOM(P_\delta,N_{required})$}
\State{concatenate($\cP_s,\cP_{\delta(random)}$}
\State{Return $\cP_s$}
\EndIf
\If{target is $\delta_t$}
\State{$\delta \gets $ highest d in training data}
\While{$\delta > \delta_t$}
\State{concatenate($\cP_s$, $\cP_\delta$)}
\State{$\delta \gets \delta - 1$}
\EndWhile
\State{Return $\cP_s$}
\EndIf
\end{algorithmic}
\end{algorithm}

\begin{figure}[hp]\centering
\includegraphics[width=0.9\columnwidth]{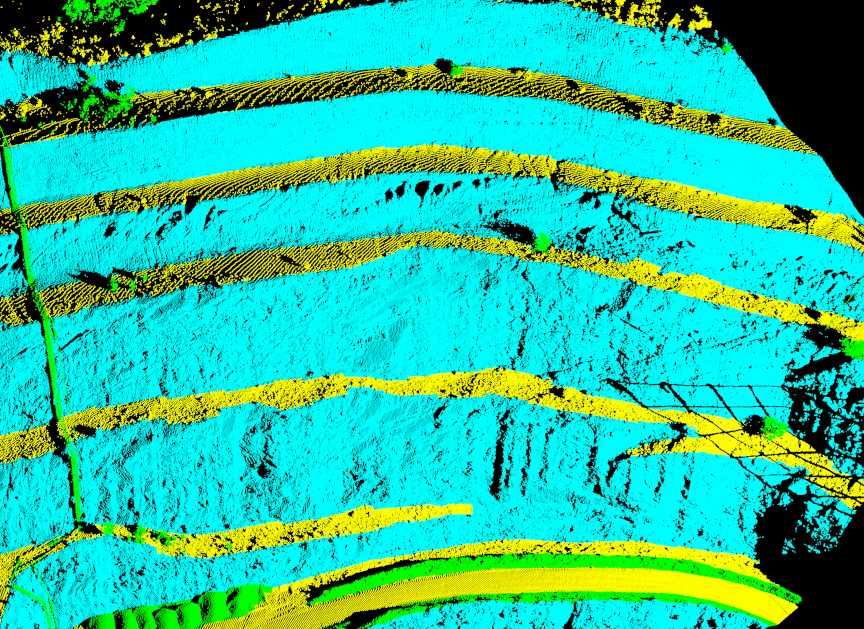}
\includegraphics[width=0.9\columnwidth]{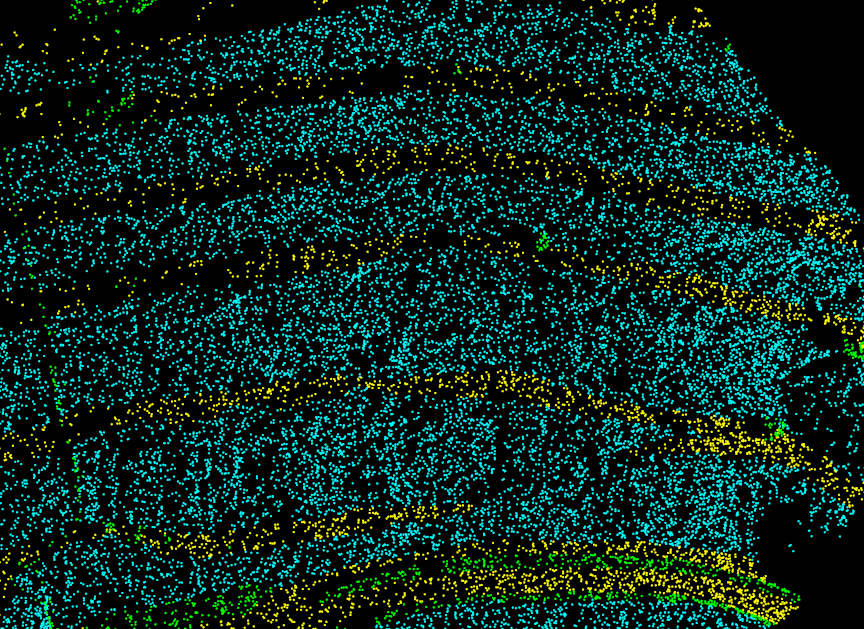}
\includegraphics[width=0.9\columnwidth]{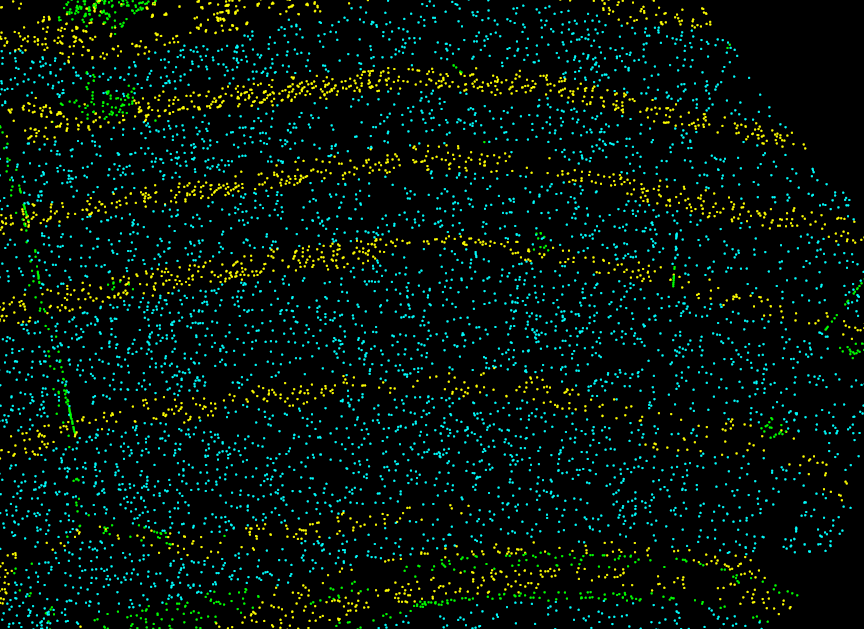}
\includegraphics[width=0.9\columnwidth]{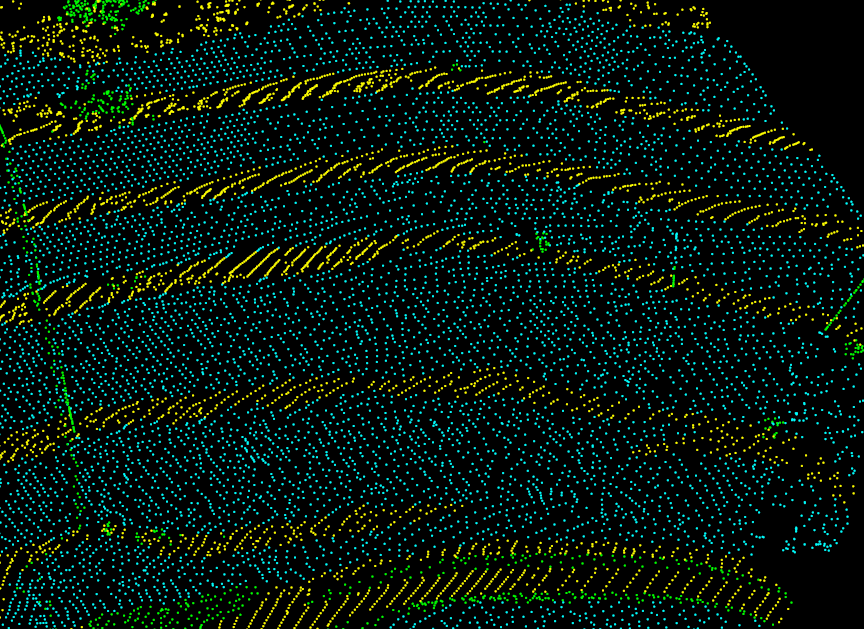}
\caption{Subsampling variants. Left-Right, Top-Down: Original, random, $\delta$-guided random, and LiDAR-grid. Classes Wall/Ground/Other are bright Blue/Yellow/Green for visual clarity}
\label{fig:SamplingComparisons}
\end{figure}

Like existing networks, HDVNet subsamples the point cloud $\cP$ to smaller subsets of points. This allows both for more features to be encoded into each point without running into hardware limitations, as well as obtaining features of lower density resolutions. In \cref{fig:proposedmodel} this is represented by the downsampling step DS.

Downsampling methods in previous models vary from random sampling~\citep{Randlanet} and farthest point sampling~\citep{PointNet++,Dance-Net}, to having the network itself choose which points to keep \citep{TrainableDownsampling}. Such downsampling methods do not retain the inherent scan ordering of terrestrial LiDAR. We use a pseudo-LiDAR downsampling method similar to that of other works \citep{LiDARScanlineAugmentationPaper, LiDARDistillation} to preserve scan lines. With the goal of making the scan more homogeneous in density with each downsampling step, objects with higher density in the scan have more points removed, while the lowest density sections of the scan are left untouched.

Such terrestrial LiDAR scanners output not only the 3D coordinates $(x,y,z)$ of each scan point, but often also the (spherical) row and column coordinates $(r,c)$ of the corresponding scan direction. While they can be estimated when the scanner's co-ordinates are known (usually the point of origin for the scan), it is preferable to use the original scanner's row and column values if they are available for better LiDAR scan metadata. We propose that respecting the scan structure of the LiDAR while downsampling enables high-density sections of a scene to better resemble their low-density counterparts after downsampling; see \cref{fig:SamplingComparisons}. When downsampled sections of the point cloud do not resemble naturally sparse objects in the LiDAR scan, the network is less effective at extracting coarse features.

For each point $\bp_i$, the original metadata $\bM_i = [r_i,c_i]$ is also input to the network if available. The metadata is used exclusively for downsampling, and not directly used in mapping from $\bp_i$ to the class confidence distribution $\tilde{\bq}_i$. Instead, it enables a more accurate downsampling of high-density points, removing the disparities and differences between different density groupings. We define the difference between the downsampling's target grouping $\delta_t$ and the point's inherent density grouping $\delta_i$ as $\Delta \delta = \delta_t - \delta_i$.
\begin{equation}
DS_{lidar}(c_i,r_i, \Delta \delta) =
    \begin{cases}
  \text{if}~c_i \% 2^{\Delta \delta} = 0 \text{ and } r_i \% 2^{\Delta \delta} = 0 & 1,\\
  \text{otherwise} & 0,
    \end{cases}       
\end{equation}

\noindent We keep $\bp_i$ if $DS_{lidar}(c_i,r_i, \Delta \delta) = 1$, and discard it in the downsampling otherwise. Downsampling based on the difference between a point's original density and the target density creates a more homogeneous result, while using rows and columns allows the LiDAR scanline structure to be retained, as shown in the bottom row of \cref{fig:SamplingComparisons}. 

One notable downside of this subsampling approach is that the number of points removed is inconsistent, as the number of points $\cP_\delta$ in each density grouping varies scan by scan. As a simple solution, LiDAR-grid subsampling is used for successive density groups until a new target density would result less points than desired. The points which would be removed when downsampling to the next target density $\delta_t$ are then randomly removed to achieve the desired number of points $N_t$ as shown in \cref{alg:LiDARSubSample}. 

We use random subsampling as if we were to select the points based on their local density it would likely remove a small object or section of the scan. Randomly subsampling is both computationally efficient and spreads the sampling throughout the scan. By randomly selecting the points which would have been removed if the LiDAR-grid subsampling was used once more, we also retain scan line structure as much as possible while avoiding subsampling of low-density areas of the scan.

\begin{figure}[htb!]
    \includegraphics[width=0.45\textwidth]{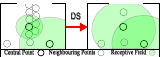} 
    \caption{Visualisation of how two points have had vastly different receptive fields initially (left), After downsampling they are both neighbours of each other (right). The reliability of their dense features is not equal, so feature aggregation and propagation should be done with care.}
    \label{fig:PointExistenceExample}
    
\end{figure}

\subsection{Existential local feature aggregator (ELFA)}
\label{sec:ELFA}

\begin{figure*}[htb!]
    \includegraphics[width=0.99\textwidth]{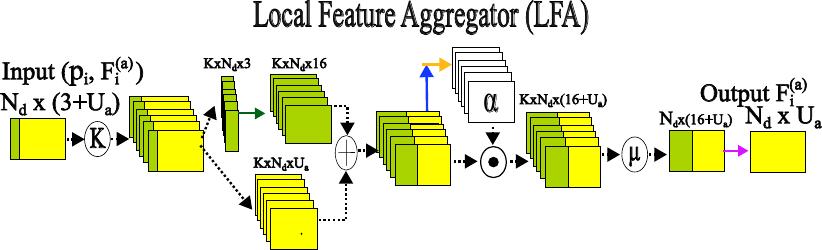} 
    \includegraphics[width=0.99\textwidth]{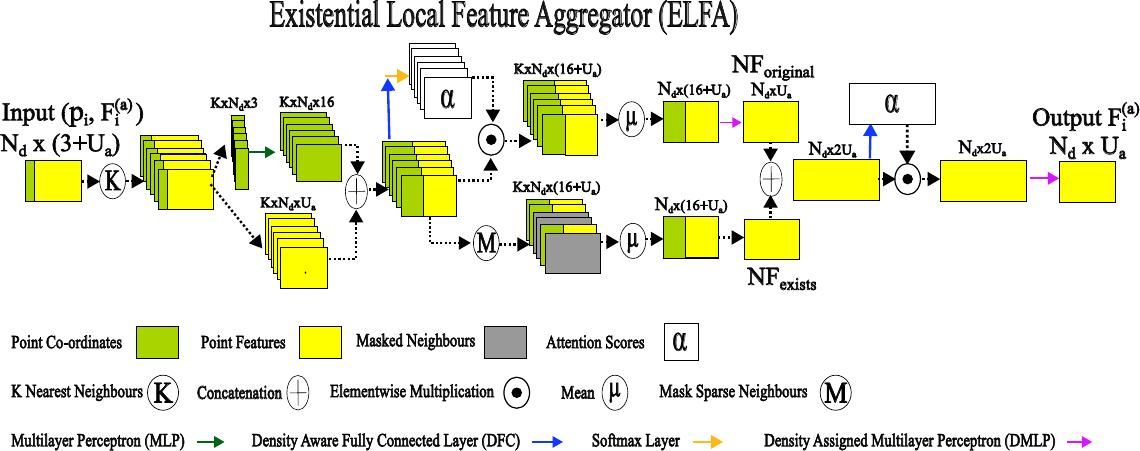}  
    \caption{A standard LFA (above) and our ELFA (below). ``K'' represents the stacking of the point's $K$ neighbouring points. HDVNet uses DMLP and DC's to ensure feature assignments are upheld. As point coordinates are not density-assigned features a normal MLP is used to increase them from 3 to 16 elements. In later DMLPs involving point features and density assignments, co-ordinates are treated as $d=5$ (visible by all densities). ELFA is optionally used instead to simultaneously calculate the feature based both on neighbours which ``exist'' and those which do not. Due to K-neighbours' features multiplying memory use by K, reduced feature vector element total $U_a$ is used for LFA blocks as shown in \cref{fig:DensityAssignedDecoderBlock}}
    \label{fig:ExistentialLocalFeatureAggregator}
    
\end{figure*}

In HDVNet, Local features are extracted and aggregated via a nearest neighbours approach. As shown in \cref{fig:ExistentialLocalFeatureAggregator}, we have two alternative feature aggregation blocks. LFA is similar to that used in existing networks, specifically using the LFA from RandLA-Net \citep{Randlanet} as a base. The point co-ordinates and features for the $K$ neighbours in $\cN_i$ are found, and both are used to generate a local feature vector for each point $\bp_i$. The only modification of note to our LFA implementation is that the MLPs which involve point features are replaced with DMLPs, and fully connected layers (FC) are similarly replaced with density connected layers (DC). The density assignment $\bA_d$ of the point features is thus preserved. ELFA is a more modified, optional variant, which further counters density-variation. 

As the feature elements which are assigned to higher densities are calculated for sparse points, there will be ``unreliable'' or ``junk'' features, such as the dense features of the bottom right point in \cref{fig:PointExistenceExample} with a larger receptive field. While the network can be designed not to use them at all, that removes both the ability of sparse points to utilise fine features of their higher-density neighbours, as well as take unreliable (due to varying receptive field size) but still potentially useful fine features into consideration.

In ELFA, two neighbourhood features are created. All $K$ neighbours are used to generate $\mathbf{NF}_{original}$ as usual, while $\mathbf{NF}_{exists}$ is created from what remains after masking out points which exist at a sparser density than the point cloud's current density state $d$.

\begin{equation}
Mask(\bp_i,d) =
\begin{cases}
  \text{if}~\bp_i \in \{I^{(j)}\}^{d}_{j=1} & 1,\\
  \text{otherwise} & 0,
\end{cases}
\label{eq:Masking}
\end{equation}

\noindent This masking ensures only points with the expected receptive field size contribute to $NF_{exists}$. Both neighbourhood features are concatenated, multiplied by an attention score used to determine which features are most reliable, and then a finally passed to a DMLP. This is visualised in \cref{fig:ExistentialLocalFeatureAggregator}. Through ELFA, the network has a local neighbourhood feature $\mathbf{NF}_{exists}$ which it can learn whether or not to trust. Without it, there is a higher risk of the network taking and using ``junk'' dense features from neighbouring points which have a sparser inherent density.

\begin{algorithm}
\caption{Existential Local Feature Aggregator}\label{alg:ELFA}
\begin{algorithmic}
\Require{Local Neighbourhood Feature vector $NF$, Inherent density state of the $K$ neighbours $I_K$, Inherent density state of the current point $I_p$}
\Ensure{Feature vector $F$}
\State{$mask \gets I_K \leq I_p$}
\State{$NF_{exists} \gets NF * mask$}
\State{$Attention \gets FC(NF)$}
\State{$NF_{original} \gets NF * Attention$}
\State{$NF_{original} \gets DMLP(NF_{original},1)$}
\State{$NF_{exists} \gets \mu(NF_{exists})$}
\State{$F \gets concat(NF_{original},NF_{exists})$}
\State{$Attention \gets FC(F)$}
\State{$F \gets F * Attention$}
\State{$F \gets DMLP(F,size(F)$}
\State{Return $F$}
\end{algorithmic}
\end{algorithm}

\subsection{Initial Training Classifiers}

\begin{figure}[htb!]\centering
\includegraphics[width=1\columnwidth]{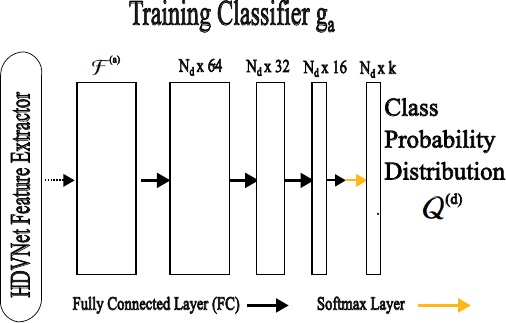}
\label{fig:trainingClassifier}
\caption{The simple classifiers used during training. $g_1$, $g_2$, $g_3$ and $g_4$ all use this architecture, using their corresponding input $\cF^{(d)}$}
\end{figure}

 \begin{figure*}[!htb]\centering
\includegraphics[width=0.99\textwidth]{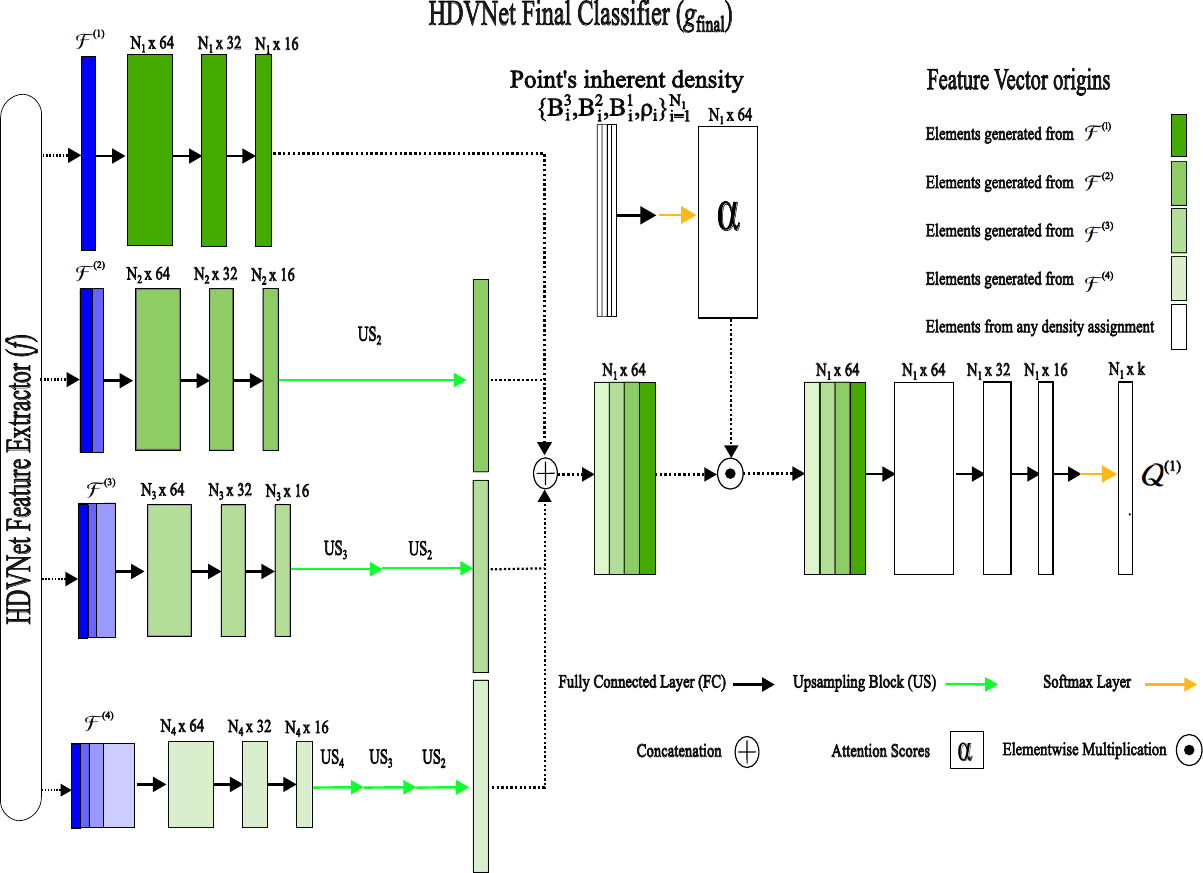}
\caption{The final classifier used during inference $g_{final}$. The $\cF^{(d)}$ from each density state $d$ is passed to a single classifier. Attention scoring is then used to generate a reliable class probability distribution $\cQ^{(1)} = \{ \tilde{\bq_i} \}^{N_1}_{i=1}$. Elements generated from each specific feature vector is visualised in shades of green to make the purpose of the density-based attention scoring clearer.}
\label{fig:FinalClassifier}
\end{figure*}

\label{sec:InitialTraining}

As the network architecture is assigned by density throughout, we are able to utilise multiple classifiers $g_1....g4$ at the end, each predicting a class confidence distribution $\tilde{\bq}_i$ for each point. Each $g_a$ takes the features from a different density state of the decoder as its input, $g_4$ using $\cF^{(4)}$, $g_3$ the features from $\cF^{(3)}$ and so forth. 

The class-weighted cross-entropy loss is calculated for each separate $\{ \tilde{\bq_i} \}^{N_d}_{i=1}$ produced by $g_a$, masked to include only points with inherent densities belonging to that density state or a prior one, $\bp_i \in \{I^{(j)}\}^{d}_{j=1}$. This specialises each classifier for its intended density, preventing $g_1$ from being expected to classify sparse points of $I^{(2)},I^{(3)},I^{(4)}$ or $I^{(5)}$ (each density is visualised in \cref{fig:PdvsId}). We include earlier density states due to the LiDAR-grid subsampling making high density objects resemble sparse ones, making them suitable as extra training data for sparser densities.

$I^{(5)}$ makes up a negligible proportion of any point cloud $P^{(1)}$, so it does not have a corresponding classifier and cross-entropy loss is not calculated for it. Any points which belong to $I^{(5)}$ are treated as $I^{(4)}$ when masking the output $\{ \tilde{\bq_i} \}^{N_d}_{i=1}$ and calculating the loss.

To combine them together, the loss $L$ for each density state $d$ is then multiplied by the square of the density state number itself, so that the network can be trained simultaneously for all densities.
\begin{equation}
      L_{total} = 1^{2}L_1 + 2^{2}L_2 + ......d^2 L_d
      \label{eq:LossFunction}
\end{equation}
\noindent The lower density weights are thus prevented from being too strongly affected by the higher density outputs which also use coarse features in their calculations, and thus affect the coarse features in their backpropagation. 


\subsection{Fine tuning for final prediction}

\label{sec:FineTuning}
While the loss in the section above is used during initial training, there is a final fine-tuning step afterwards. Simply predicting the class label probability $\{ \tilde{\bq_i} \}^{N_d}_{i=1}$ using the output from the classifier $g_a$ corresponding to the point's inherent density $\bp_i \in I^{(d)}$ is sufficient. However a benefit can be gained by locking the weights previously trained and fine-tuning new ones which take all the the extracted features as input into a singular $g_{final}$ shared by all the points.

As shown in \cref{fig:FinalClassifier}, the features at each density are first up-sampled to cover all the original input points, before being attention scored for each point. This attention score $\boldsymbol{\alpha}_i$ is created based on which density states $d$ the point $\bp_i$  ``exists'' in as well as it's specific density estimate $\rho_i$. A boolean value $B_i^{(d)}$ is used, with the value being true using the same ``existence'' definition as in ELFA - whether the point belongs to $I^{(d)}$ or that of a prior density state (\cref{eq:Masking}).

As the network is initially trained for the classifiers $g_1....g_4$, there are no features assigned to $d=5$ to be attention scored. Therefore no boolean is made for $d=5$. At $d=4$ all points other than the negligible amount existing in $I^{(5)}$ would be given a value of 1 according to \cref{eq:Masking} so $B_i^4$ is not calculated or included either.

\begin{equation}
      \boldsymbol{\alpha}_i = MLP(B_i^1, B_i^2, B_i^3, \rho_i)
\end{equation}

\noindent As the point's density is known, and each feature is assigned to a designated density state, the network is able to learn which features to rely on for the final prediction of $k$ classes, and apply the attention score $\boldsymbol{\alpha}_i$ accordingly. The loss for this final step of the training is simply a class-weighted cross entropy loss using the $\{ \tilde{\bq_i} \}^{N_1}_{i=1}$ output by $g_{final}$.

\section{Dataset - HDVMine}

With the assistance of an industry partner, we collected 53 individual terrestrial LiDAR scans across five different mine locations; \cref{fig:single_mine} shows the point cloud from an individual scan. The scope of the individual point clouds range from 183M in one direction to  8.4KM, with an average of 577M. \cref{fig:ScanComparisons} displays one of the scans from above.

\begin{figure}[ht]
    \begin{subfigure}[t]{\textwidth}
        \includegraphics[width=0.24\textwidth]{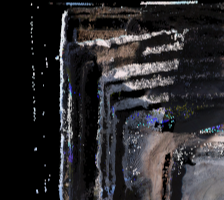}
        \includegraphics[width=0.24\textwidth]{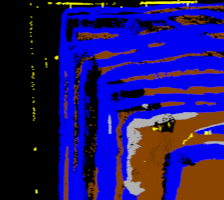}
    \end{subfigure}
    \caption{BEV of the corner of an open-cut mine in HDVMine. (top) Scene stitched from 7 point clouds, textured with RGB photos. (bottom) Ground truth semantic labels \texttt{wall} (blue), \texttt{ground} (brown), loose rock (grey), and manmade objects (yellow)}
    \label{fig:stitched}
\end{figure}

\begin{figure}[ht]\centering
\begin{subfigure}[b]{0.99\columnwidth}\centering
\includegraphics[width=0.99\columnwidth,height=7em]{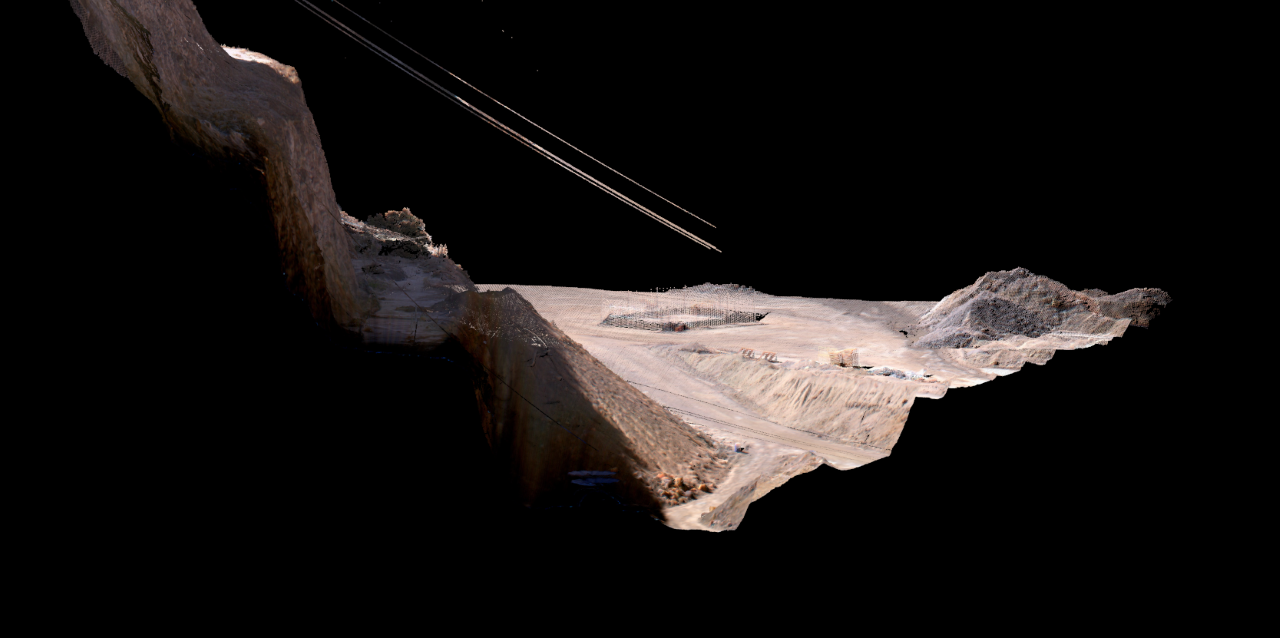}
\caption{\texttt{ground} can be flat road or rocky bench. \texttt{wall} similarly varies in smoothness, and the angle between their orientations is not consistent.}
\label{fig:WallSlant}
\end{subfigure}
\begin{subfigure}[b]{0.99\columnwidth}\centering
\includegraphics[width=0.99\columnwidth,height=7em]{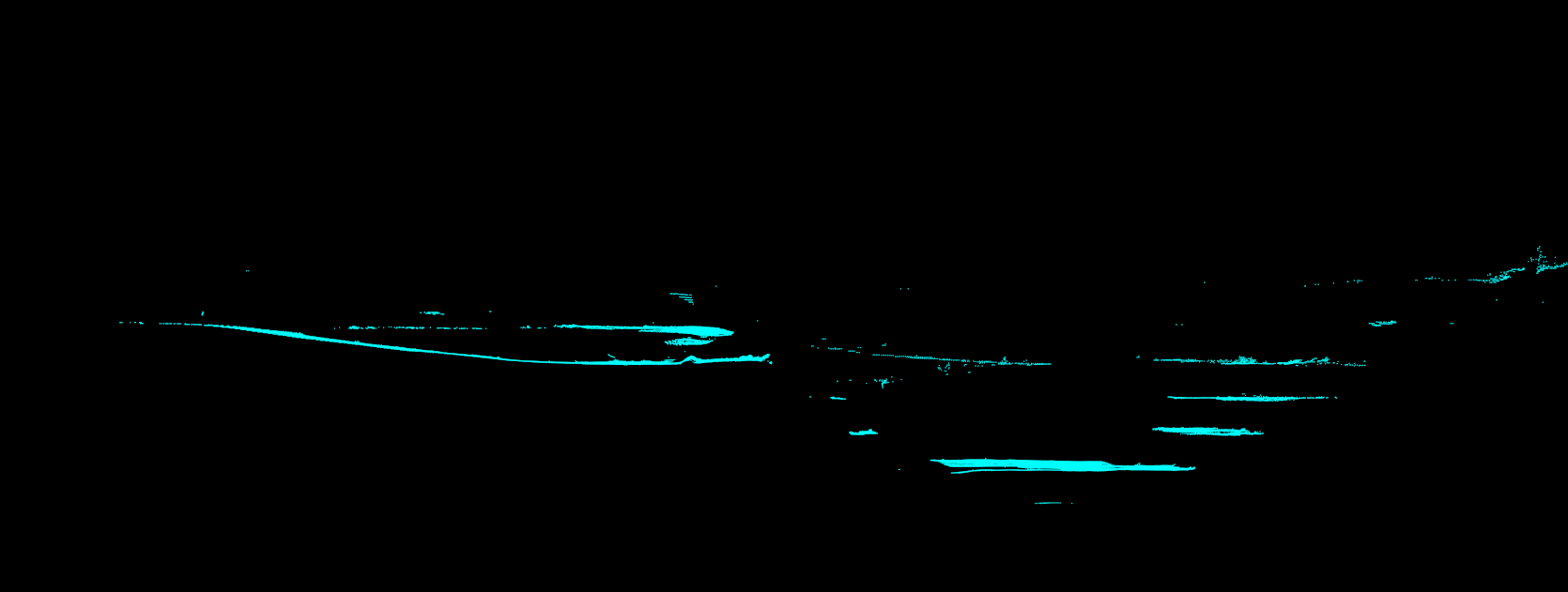}
\caption{Classes defy simple geometric definitions. Multiple ``ground planes'' shown blue in a side-view of a single scan with other points removed.}
\label{fig:GroundPlanes}
\end{subfigure}
\begin{subfigure}[b]{0.99\columnwidth}\centering
\includegraphics[width=0.99\columnwidth,height=7em]{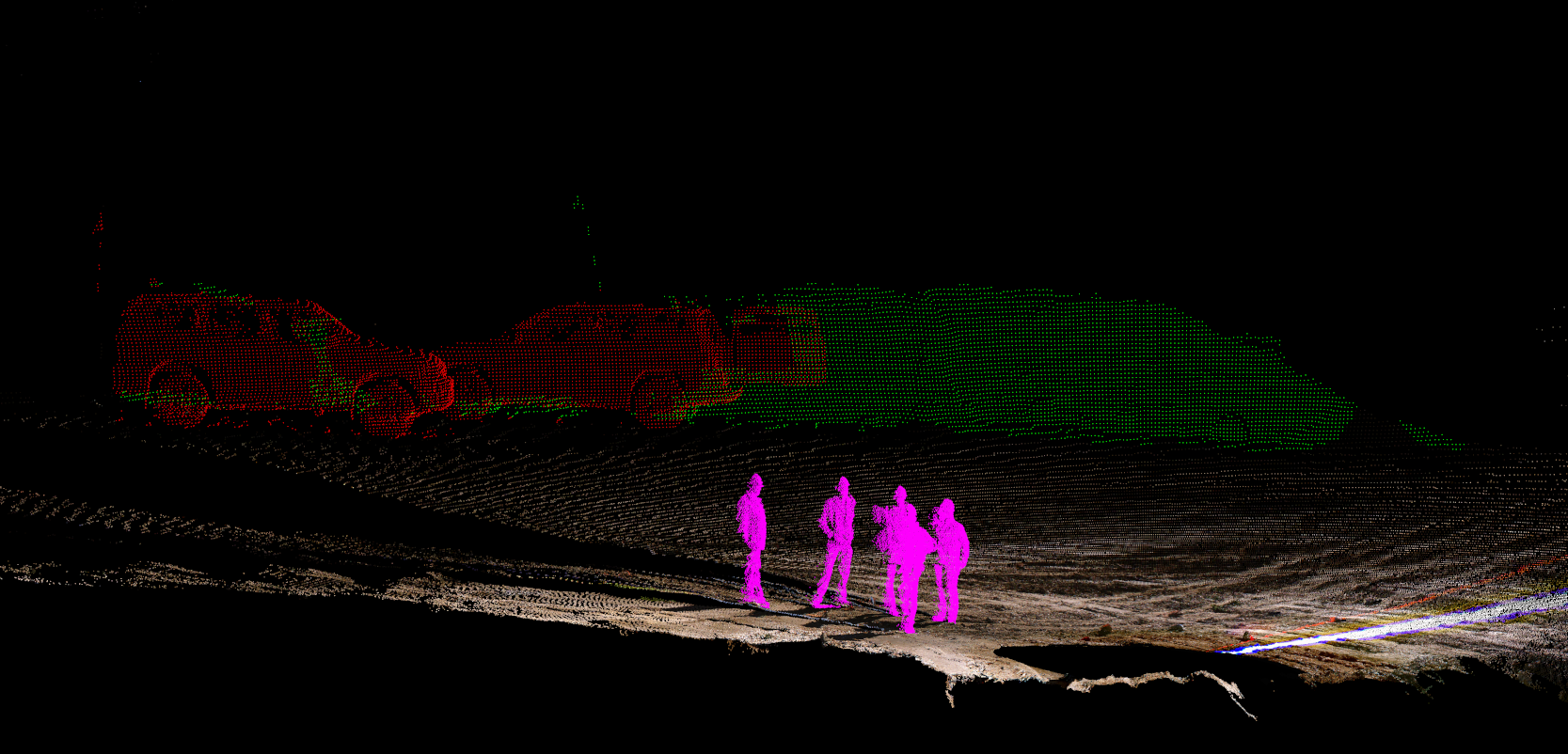}
\caption{People, Vehicles, and a pile of rock, all examples of the \texttt{other} class.}
\label{fig:OtherClass}
\end{subfigure}
\begin{subfigure}[b]{0.99\columnwidth}\centering
\includegraphics[width=0.99\columnwidth,height=7em]{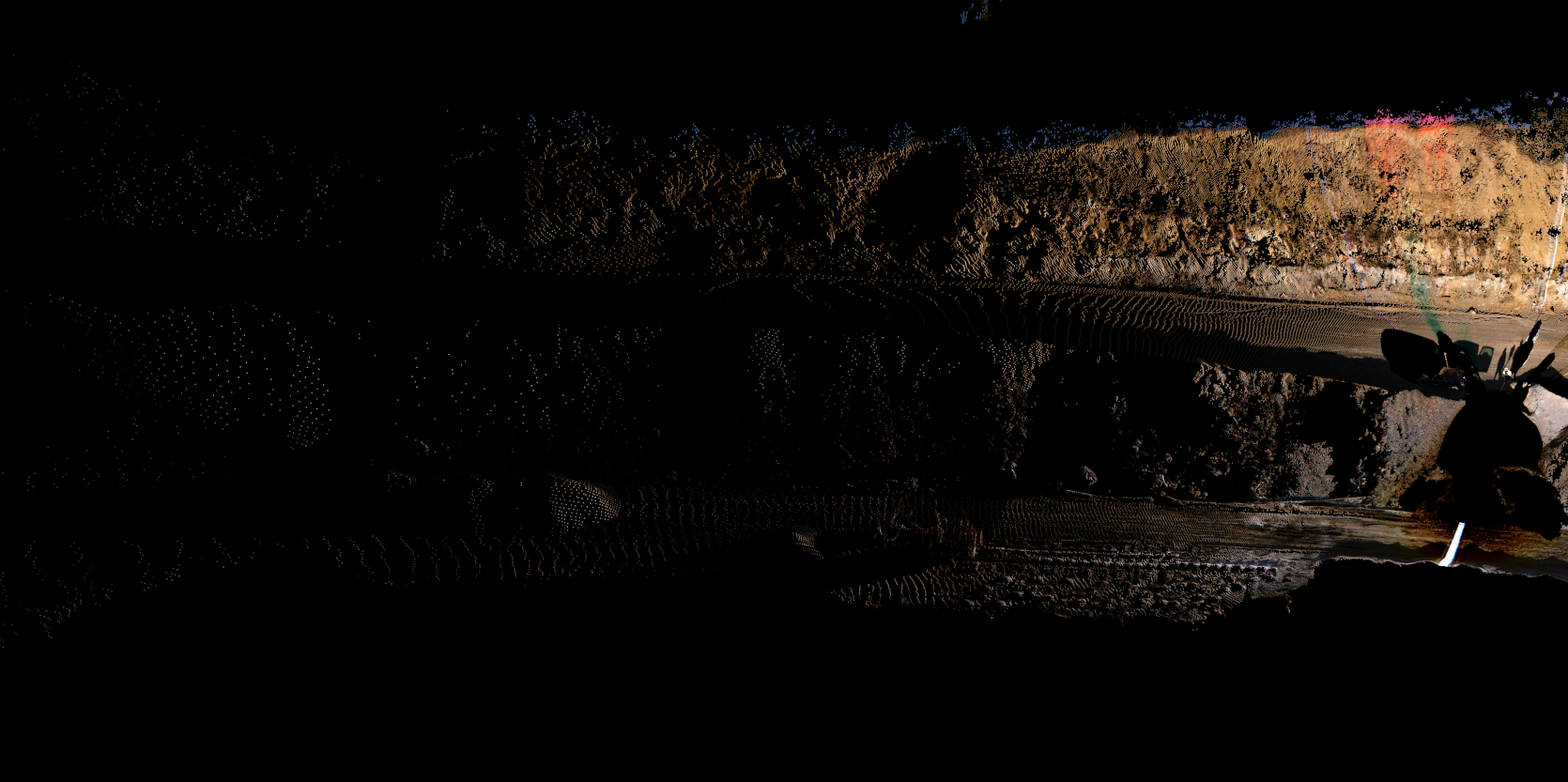}
\caption{High intra-class local density variation, even within the same instance. Wall and Ground shown losing density as distance from scanner grows.}
\label{fig:DensityVariation}
\end{subfigure}
\caption{Challenging features of the HDVMine dataset.}
\label{fig:HDVMineChallenges}
\end{figure}

We manually labelled the point clouds into three semantic classes: \texttt{wall}, \texttt{ground} and \texttt{other}. The classes chosen reflect the aim to understand the overall scene structure for surveying. Unlike in urban environments, \texttt{wall} and \texttt{ground} in a mining environment vary significantly in smoothness and orientation. The boundaries between \texttt{wall} and \texttt{ground} also defy simple geometric definitions, \eg, the surfaces are not cleanly at right angles. \cref{fig:WallSlant} illustrates these challenging features. Class \texttt{other} subsumes a variety of elements such as vegetation, rock piles, and man-made objects, where the latter encompass less than $1$\% of the points; see \cref{fig:OtherClass}. In total, 353 million points have been labelled. \cref{tab:datastats} shows the population size of the classes.

\begin{table}[ht]\centering
\begin{tabular}{|c|c|}
 \hline
 Class & Percentage in overall population\\
 \hline
 \texttt{wall}	& 52.4	\\
 \texttt{ground}	& 31.1	\\
 \texttt{other}	& 16.5	\\
 \hline
\end{tabular}
\caption{Overall proportion of each class in HDVMine dataset.}
\label{tab:datastats}
\end{table}

While the LiDAR scans in HDVMine can be combined into contiguous scenes, in our experiments in Sec.~\ref{sec:results}, each scan was treated as an individual input point cloud. Even within a single point cloud however, the local density variation is high (see \cref{fig:density_histogram}), which in turn leads to significant intra-class density variation (see \cref{fig:DensityVariation} for \texttt{wall} and \texttt{ground} examples).

\section{Experiments}\label{sec:results}

\begin{table*}
\begin{tabular}{|p{6cm}||p{1cm}||p{1cm}||p{1cm}||p{1cm}||p{1cm}||p{1cm}||p{1cm}|}
 \hline
 \multicolumn{8}{|c|}{Results on HDVMine across different densities} \\
 \hline
  & All & $I_5$ & $I_4$ & $I_3$ & $I_2$ & $I_1$ & $I_0$\\
 \hline
 Proportion Of Scene & 100\% & 0.14\% & 1.3\% & 4.4\% & 22.8\% & 14.0\% & 57.4\% \\ 
 \hline
 RandLA-Net Original & 47.2 & 36.6 & 56.5 & 47.3& 34.4& 39.6
 & 47.0\\
 \hline
 DGCNN 5 Metre & 41.8 & 15.0 & 22.7 & 17.6& 23.1& 50.3
 & 42.6\\
 \hline
 RandLA-Net + LN	& 67.8 & 52.1 & 68.2 & 66.8 & 64.4 & 72.0 & 64.1 \\
 \hline
 RandLA-Net + LN + LGS & 70.8 & 67.6 & 76.3 & 69.6 & 69.5 & 74.7 & 66.1 \\
 \hline
 RandLA-Net + LN (Downsampled) & 67.0 & 68.0 & \textbf{79.6} & \textbf{78.2} & \textbf{75.8} & 74.4 & 56.2 \\
 \hline
 HDVNet : DTC & 69.8 & 54.5 & 74.9 & 70.2 & 70.5 & 74.2 & 64.3\\
 \hline
 HDVNet : FCO & 70.7 & 63.2 & 77.1 & 71.6 & 71.1 & 75.0 & 64.9 \\
 \hline
 HDVNet : TCO & 73.5 & \textbf{72.5} & \underline{79.4} & 72.8 & 72.5 & \underline{76.6} & 69.9 \\
 \hline
 HDVNet : No FA & \underline{73.6} & 69.2 & 76.8 & 70.3 & 72.0 & 75.4 & \textbf{70.6} \\
 \hline
 HDVNet : No FA (small)  & 72.7 & 65.8 & 76.7 & 71.1 & 72.3 & 76.1 & 68.9 \\
 \hline
  HDVNet & \underline{73.6} & \underline{71.8} & 79.1 & 73.2 & \underline{74.6} & 76.2 & 69.3 \\
 \hline
 HDVNet : No ELFA & \textbf{74.5} & 68.7 & 78.1 & \underline{73.3} & 74.2 & \textbf{77.5} & \underline{70.4} \\
 \hline
\end{tabular}
\caption{Ablation results on our high-variation dataset HDVMine. Value is MIoU for all points belonging to density state $d$. Best for each density is bolded, second best underlined. ``All'' is the MIoU as calculated using all the points \textit{not} the weighted average of each density's MIoU.}
\label{tab:HDVStats}
\end{table*}

Experiments were run using three different datasets, HDVMine (high-resolution, large-scale terrestrial LiDAR), Semantic3D (high-resolution terrestrial LiDAR), and HelixNet (low-resolution automotive LiDAR). We ran ablation tests with multiple variations of our architecture:

\begin{itemize}
  \item HDVNet: The default network, using all methods as outlined in \cref{sec:methodology}
  \item DTC (Density aware Training Classifier): The training classifiers are modified to use DMLP and DC layers as the rest of HDVNet does.
  \item FCO (Fine Classifer Only): Immediately train using fine classifier, instead of using the training classifier from \cref{sec:InitialTraining} and locking the network weights prior to the classifier.
  \item TCO (Training Classifier Only): Inference is run using the training classifiers from \cref{sec:InitialTraining}. Each point $p_i$ uses either $g_1$, $g_2$, $g_3$ or $g_4$ according to which density state $I^{(d)}$ it belongs to.
  \item No FA (No Feature Allocation): All DC and DMLP layers take features of \textit{every} available density as input. Such DC layers have no practical difference to a FC layer, while each DMLP retains separate layernorm (LN) and activation (AVN) for each small MLPs which it is constructed from.
  \item No FA (small): As feature allocation reduces the number of weights used by almost half, this variant also uses less features per point throughout the network, for an equivalent number of weights.
  \item No ELFA: The Existential Local Feature Aggregator from \cref{sec:ELFA} is not applied
\end{itemize}

\subsection{Results on HDVMine}

As there are multiple key differences between the implemented architecture and RandLA-Net, additional ablation tests were run on HDVMine. All tests were run on a single 8GB Nvidia RTX 3070 for 50 Epochs (with each epoch being 1000 batches of batch size 4). To fit the graph on the smaller GPU all networks were trained with the same reduced number of features per point (maxing out at 256 features per point at the end of the decoder). For all tests points were passed in with $x,y,z,r,g,b$, as well as the density estimate $\rho_i$. 

While the network takes an already-downsampled point cloud $\cP^{(1)}$ as input, we upsample the labels and test on the original point cloud $\cP^{(0)}$. For analysis we identify the accuracy both on points with an inherent density $I^{(1)}$ and those with the extremely high density of $I^{(0)}$.

In addition, as our terrestrial LiDAR scans are too large to pass as input to a standard GPU, we used the same method as RandLA-Net to break it down. Points were randomly chosen from those not yet given a label and combined with a set number of their nearest neighbours, passed into the network as the input point cloud $\cP^{(1)}$. This process was repeated until every point had been processed at least once. Points processed in more than one of these ``spheres'' had their label chosen by weighting the different class distributions using the point's distance from the centre of each respective sphere, and then using the summed probability distribution.

Points are compared at different density groupings. $I^{(5)}$ is the coarsest, including all points where $\rho_i <= 0.12$ points per $m^3$. In comparison $I^{(0)}$ is the finest, in HDVMine this is all the points where $\rho_i > 30,558$ points per $m^3$. The specific $t_d$ thresholds for $d=0,1,2,3,4,5$ are (30558, 1739, 31, 1.9, 0.12, 0) respectively, based on the known distribution of the training data.

As shown in \cref{tab:HDVStats}, RandLA-Net's use of batchnorm makes it difficult for the network to stabilise when limited GPU memory requirements require a small batch size of 4. Simply swapping it for layernorm (LN) enabled RandLA-Net to train effectively. Replacing random subsampling with our Lidar-grid subsampling (LGS) improved results again. Even with the point's density $\rho_i$ directly passed in alongside rgb as a raw point value, it was unable to learn to combat the same level of density variation as our HDVNet. Finally, we ran RandLA-Net after downsampling the data heavily in pre-processing to obtain homogeneity in the dataset (if all points are sparse, there is no dense-to-sparse variation). This merely results in high accuracy on sparse points coupled with poor results on high-density ones. Unlike HDVNet these higher results on sparse objects come at too high a cost, reducing overall performance as fine features are completely abandoned.

Restricting the network from using the features in $\cF^{(a)}$ assigned to higher densities when predicting $\tilde{\bq}$ for a coarse point was shown by ``HDVNet: DTC'' to reduce performance. As each feature up to this final step is extracted using only information from a specific density and lower, and each prediction $\tilde{\bq}_i$ with corresponding loss $L_d$ is for points of a specific density, it better for the network to learn to ignore an unreliable feature than completely ignore them in the final class probability calculations.

Training with the final classifier from the get go with  ``HDVNet: FCO'' put the majority of HDVNet's architecture to waste. High density points make up the majority of the scene, so all else being equal their gradients will overwhelm those of low-density ones making it difficult for the network to learn robust coarse features. In contrast, the training classifier from \cref{sec:InitialTraining} enables the network to learn how to reliably extract coarse features.

\begin{table*}
\begin{tabular}{|p{6cm}||p{1cm}||p{1cm}||p{1cm}||p{1cm}||p{1cm}||p{1cm}|}
 \hline
 \multicolumn{7}{|c|}{Results on Semantic3D} \\
 \hline
  & All & $I_4$ & $I_3$ & $I_2$ & $I_1$ & $I_0$\\
 \hline
 Proportion Of Scene & 100\% & 0.003\% & 0.03\% & 0.9\% & 4.9\% & 94.2\% \\ 
 \hline
 RandLA-Net & \textbf{77.06} & 20.2 & \textbf{44.5} & \textbf{68.1} & \textbf{72.8} & \textbf{77.08}\\
 \hline
 HDVNet: Everything Implemented  & 67.9 & \textbf{31.5} & 36.2 & 62.4 & 67.4 & 67.5\\
 \hline
 HDVNet: No ELFA & 71.4 & 22.9 & 34.7 & 58.3 & 67.4 & 71.9\\
 \hline
\end{tabular}
\caption{Results of local testing on a dataset with low density variation, Semantic3D, broken down across densities. As the point cloud is so homogeneous, HDVNet's density-aware architecture becomes a hindrance. Allowing fine object features to affect extraction of sparse features is both reliable and beneficial when 99.1 percent of the points have the finest features seen by the network (belonging to either $I^{(1)}$ or the downsampled-in-preprocessing $I^{(0)}$).}
\label{tab:Sem3DResults}
\end{table*}

The fine tuning step described in \cref{sec:FineTuning} causes a minor improvement compared to ``HDVNet: TCO'' which does not use it. Applying each point's corresponding label provided by each of the four initial outputs remains sufficient if a faster training time is desired however.

One point of interest in the ablation results is density assigned feature vector subsections ($\bS^{(d)}$), a fundamental aspect of HDVNet. As expected, removing it in ``HDVNet: No FA'' resulted in lesser results on all but the (most common) highest-density category $I^{(0)}$. Without any forced allocation of features, the network prioritised the more frequent $I^{(0)}$ and $I^{(1)}$ points during training. 

It was confirmed with ``HDVNet: No FA (small)'' that it is the explicit assignment of features to density states $d$ improving the results on sparser objects, and not a result of being a simplified network with almost half the weights to learn. This smaller-version performed worse than both the full-size ``No FA'' and standard HDVNet, as expected.

The existential local neighbourhood feature extraction step (ELFA) can be considered optional, and to be included if the goal is a network which performs especially well on sparse objects in a high density scene. Unlike the other measures taken in HDVNet, the ablation shows that the benefit to sparse objects is outweighed by the cost to dense ones. Even for the high-variation dataset HDVMine, ``HDVNet: No ELFA'' performs the best overall.

Ultimately HDVNet (No ELFA) achieved a MIoU 6.7 points above that of a RandLA-Net with minimal modifications, outperforming across all densities as well as against further simple RandLA-Net modifications.

Tests were also run using DGCNN for further comparison to existing models. The standard hyperparameter used by DGCNN for indoor scenes is 1.5 metre cubic blocks, with DGCNN taking approximately 8000 points from each block. On the HDVMine dataset, the average block has 8000 points only at 5 metres, so we made this minor change to better accommodate the network. Even at 5 metres, this merely reflects the number of points in an "average" block, with many of the blocks created having less points, some substantially so. As shown in the \cref{tab:HDVStats} models such as DGCNN which split the scene into geometric sections (in this case, five metre cubes) perform poorly on high density variation data such as HDVMine, as they struggle to train with so many low-point blocks. In inference, DGCNN shows a further decrease in performance at lower densities, as those are the blocks which do not have sufficient points for the network to effectively extract features. Further modifications such as reducing the number of points expected from each block or increasing the block size further, would throw away the fine features within the many 5-metre blocks which do have 8000 or more points.

\subsection{Results on Semantic3D}

\begin{table*}
\begin{tabular}{|p{6cm}||p{1cm}||p{1cm}||p{1cm}||p{1cm}||p{1cm}||p{1cm}|}
 \hline
 \multicolumn{7}{|c|}{Results on Helixnet} \\
 \hline
  & All & $I_5$ & $I_4$ & $I_3$ & $I_2$ & $I_1$\\
 \hline
 Proportion Of Scene & 100\% & 0.34\% & 2.3\% & 12.2\% & 37.1\% & 48.0\%\\ 
 \hline
 RandLA-Net (LN + LGS) & 49.8 & 14.8 & 24.5 & 37.3 & 52.9 & 56.0 \\
 \hline
 HDVNet & 50.8 & \textbf{24.9} & \textbf{37.9} & 46.5 & 54.1 & 50.9 \\
 \hline
 HDVNet: No ELFA & 53.0 & 23.2 & 34.9 & 46.1 & 55.4 & 54.3 \\
 \hline
 HDVNet: No ELFA, Limited FA & \textbf{56.2} & 24.2 & 36.8 & \textbf{46.9} & \textbf{58.4} & \textbf{58.8} \\
 \hline
\end{tabular}
\caption{Results of local testing on automotive dataset HelixNet, broken down across densities. As the point cloud is already low resolution, there is no downsampling in preprocessing, resulting in no $I^{(0)}$}
\label{tab:HelixNetResults}
\end{table*}

HDVNet was also applied to the task of Semantic3D \citep{Semantic3D}. The original metadata $\bM_i$ is not publicly available so angles were estimated using $x,y,z$, and from these angles rows and columns roughly approximated. Three of the fifteen scans typically used as part of the training set were instead put aside to use for testing. This was done as the Semantic3D test dataset does not have a public ground-truth point annotation, so detailed analysis across densities required sectioning off some of the publicly labelled training data. 

As shown in \cref{fig:density_histogram} smaller scale terrestrial LiDAR such as Semantic3D is significantly more homogeneous than HDVMine. The majority of points belong to the density state $I^{(0)}$, which for Semantic3D is a threshold of $\rho_i > 141,471$ points per $m^3$. \cref{tab:Sem3DResults} confirms that the improved performance seen on the HDVMine dataset does not carry over to datasets with a more homogeneous density, although it continues to perform adequately. In contrast to existing networks HDVNet is designed with the inherent assumption of density variation in the data, instead of homogeneity.

It should be noted that ``HDVNet: Everything Implemented'' performing better on ``All'' densities than at any individual one is \textit{not} a calculation error but a natural result of how the MIoU is calculated. As a general trend, individual classes get the highest IoU for the density they most commonly occur, as this density state is also how they commonly appeared in the training data. In Semantic3D this is $I^{(0)}$ for all classes except``High Vegetation'', which has 44\% of its testing points at $I^{(1)}$, despite that density only including 4.9\% of the testing dataset's points. The MIoU at $I^{(0)}$ averages each across every class, and so is affected by (relatively) poorer performance of ``High Vegetation''. Similarly the MIoU at $I^{(1)}$ is negatively affected by the IoU of classes which are most populous at $I^{(0)}$. When calculated for ``all'' densities, each class IoU is affected primarily by the density where it has the majority of points (each of those points being either a true or false positive in the IoU calculation). This is what results in the ``All'' point MIoU of 67.9\% being higher than for any of its density subsets $I^{(d)}$. The tables with the IoU of every class, at every density, for every network architecture, are not included in this paper for brevity. 

\subsection{Results on HelixNet}
Analysis was also performed using the automotive LiDAR dataset HelixNet \citep{HelixNet}. Automotive LiDAR datasets are typically much lower resolution, however also have a higher variance in density than public terrestrial datasets such as Semantic3D. Once again we show improved performance compared to the similarly point-based network RandLA-Net, with performance especially improved on lower resolutions. Similarly ELFA once more improves performance on coarse points, but is detrimental to the overall performance. 

HDVNet is built with the assumption that the point cloud still has useful features after downsampling steps. We found that due to the resolution being low to begin with, this assumption no longer holds. Assigning features to the densities $I^{(5)}$ and $I^{(4)}$ was counterproductive, with a point cloud downsampled more than three times becoming too sparse to still have useful features to extract from the raw point data. Restricting the density assignment of features to the first three density states resulted in a small increase in performance.

While the assumption of downsampled density states still having features worth extracting is an important weakness of our method to note, it is ultimately intended for high-resolution scenes such as our HDVMine. For low resolution LiDAR scans, state of the art voxel networks have demonstrated great success compared to direct point cloud processing. For Helixnet, as well as other automotive datasets, grid-based networks significantly outperform our method, RandLA-Net, and other methods which directly process raw point clouds. Whether converting to a cylindrical representation, voxels, pillars, \etc, a low-resolution point cloud does not have as much information and detail to potentially be lost in the conversion, reducing the need for direct point processing. 

\subsection{Qualitative Results}

In addition to the tables \cref{tab:HDVStats,tab:Sem3DResults,tab:HelixNetResults}, we have produced qualitative results for all architectures on all datasets. We visualise both the class predictions, as well as the point accuracy.

\hspace*{2cm}
\newline \hspace*{2cm}
\newline \hspace*{2cm}
\newline \hspace*{2cm}
\newline \hspace*{2cm}
\newline \hspace*{2cm}
\newline \hspace*{2cm}
\newline \hspace*{2cm}
\newline \hspace*{2cm}
\newline \hspace*{2cm}
\newline \hspace*{2cm}
\newline \hspace*{2cm}
\newline \hspace*{2cm}

\begin{figure*}
     \centering
     \begin{subfigure}[b]{0.33\textwidth}
         \centering
         \includegraphics[width=\textwidth]{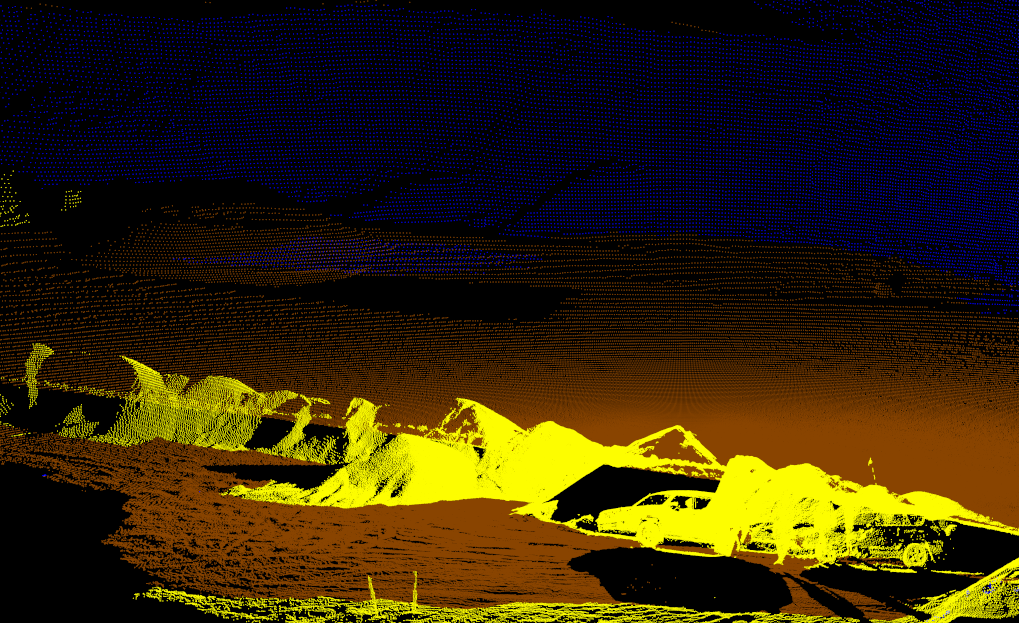}
         \caption{Ground Truth}
         \label{fig:y equals x}
     \end{subfigure}
     \begin{subfigure}[b]{0.33\textwidth}
         \centering
         \includegraphics[width=\textwidth]{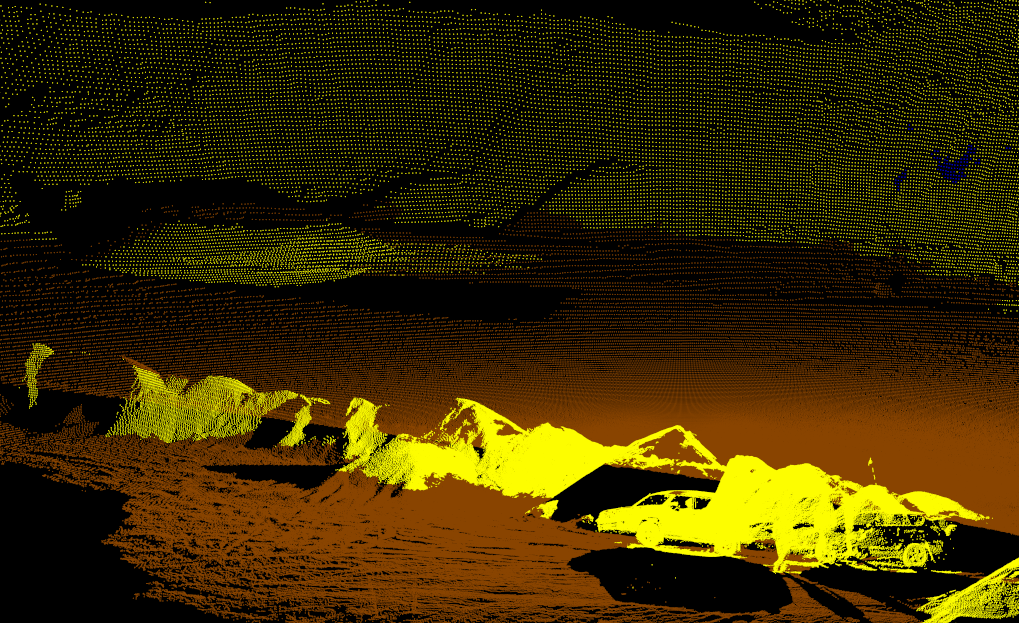}
         \caption{RandLA-Net}
         \label{fig:five over x}
     \end{subfigure}
     \begin{subfigure}[b]{0.33\textwidth}
         \centering
         \includegraphics[width=\textwidth]{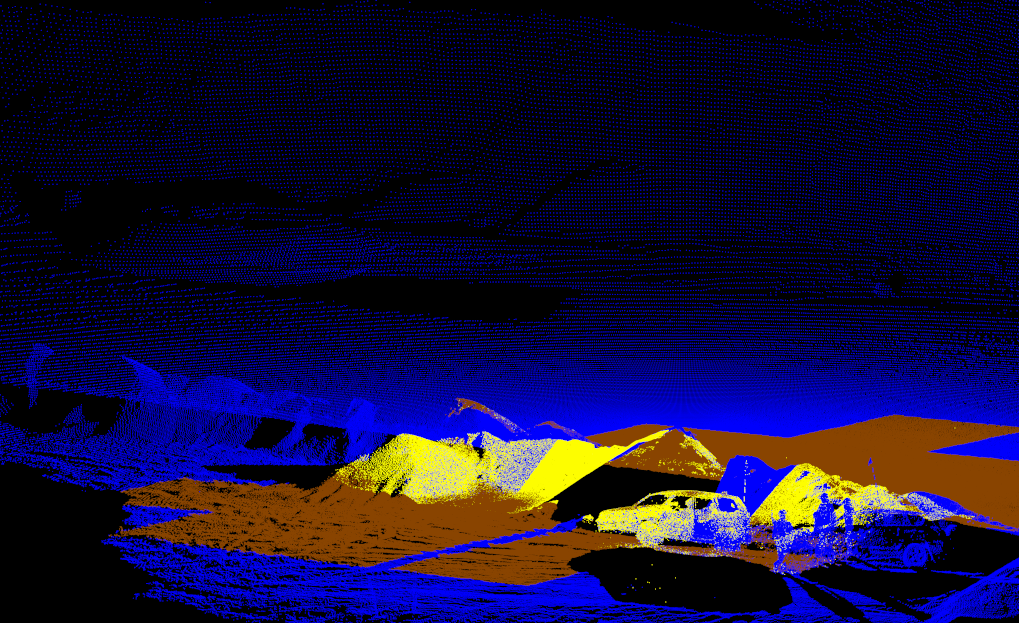}
         \caption{DGCNN}
         \label{fig:three sin x}
     \end{subfigure}
     \begin{subfigure}[b]{0.33\textwidth}
         \centering
         \includegraphics[width=\textwidth]{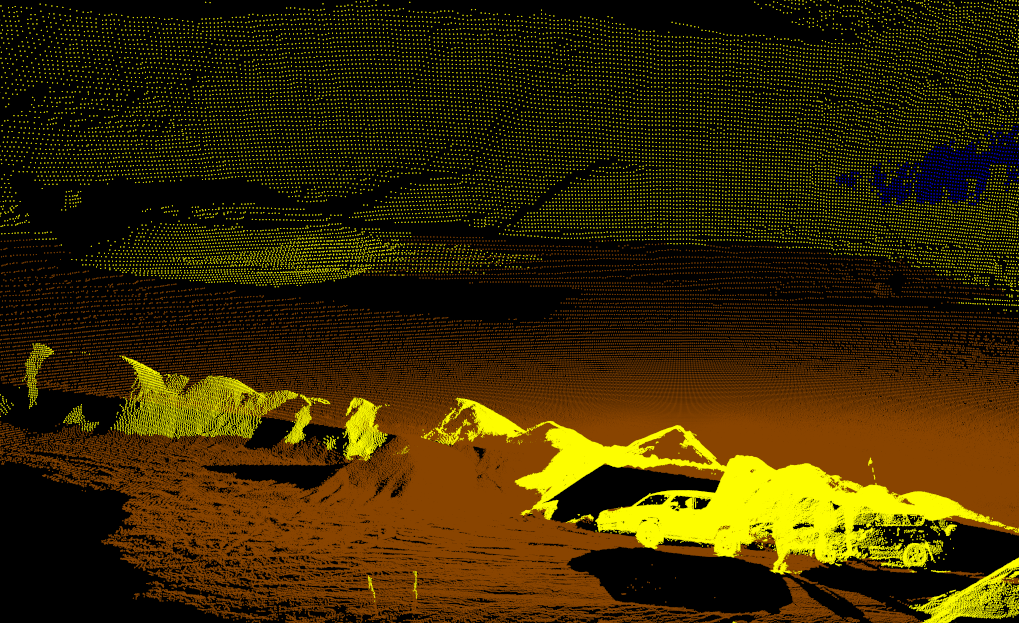}
         \caption{RandLA-Net + LN}
         \label{fig:five over x}
     \end{subfigure}
     \begin{subfigure}[b]{0.33\textwidth}
         \centering
         \includegraphics[width=\textwidth]{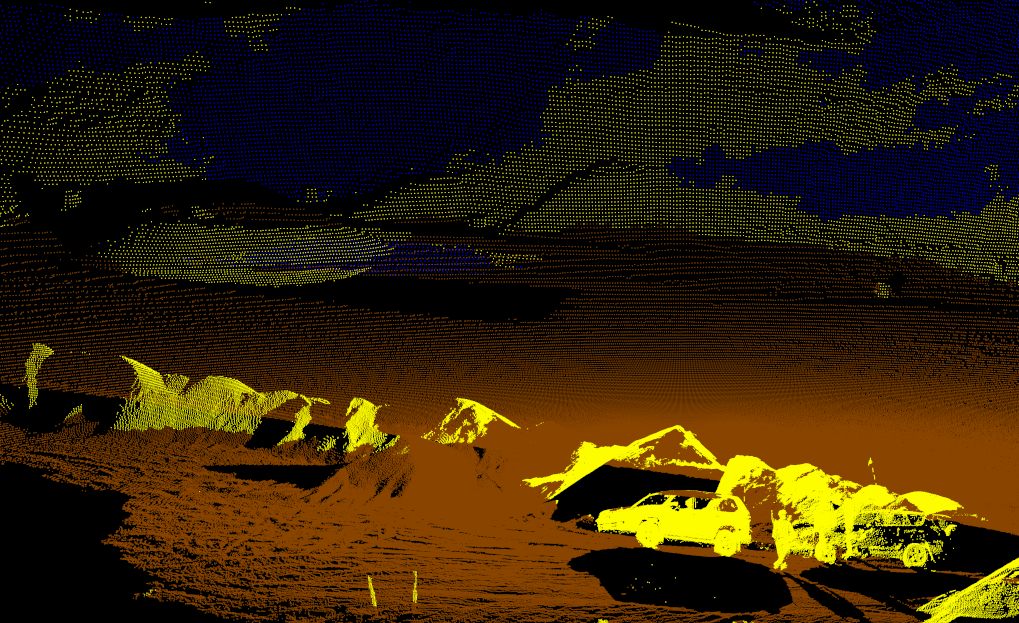}
         \caption{RandLA-Net + LN + LGS}
         \label{fig:five over x}
     \end{subfigure}
     \begin{subfigure}[b]{0.33\textwidth}
         \centering
         \includegraphics[width=\textwidth]{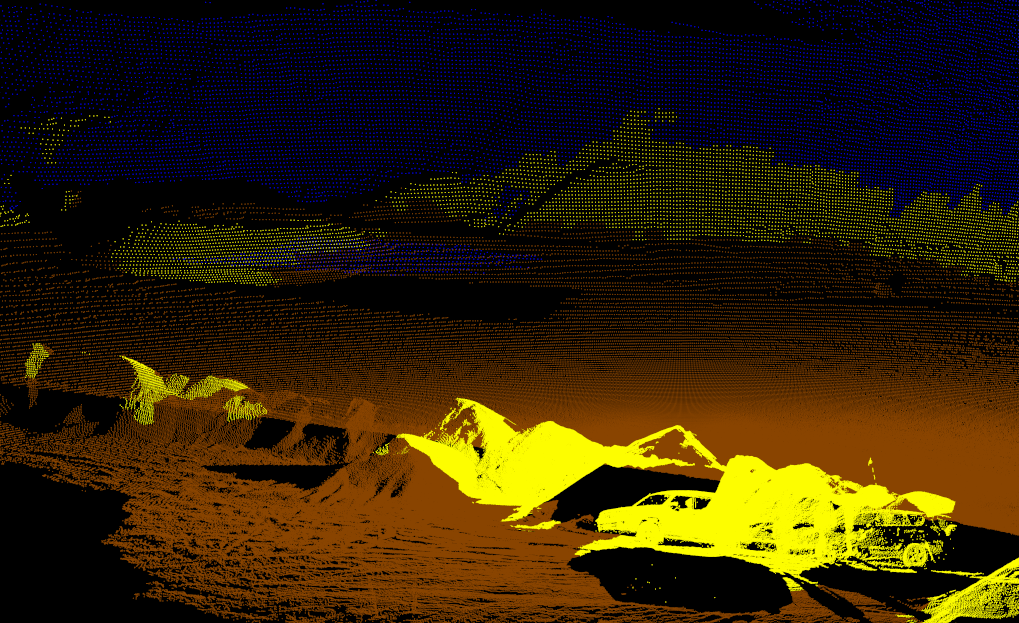}
         \caption{RandLA-Net + LN (Downsampled)}
         \label{fig:five over x}
     \end{subfigure}
     \begin{subfigure}[b]{0.33\textwidth}
         \centering
         \includegraphics[width=\textwidth]{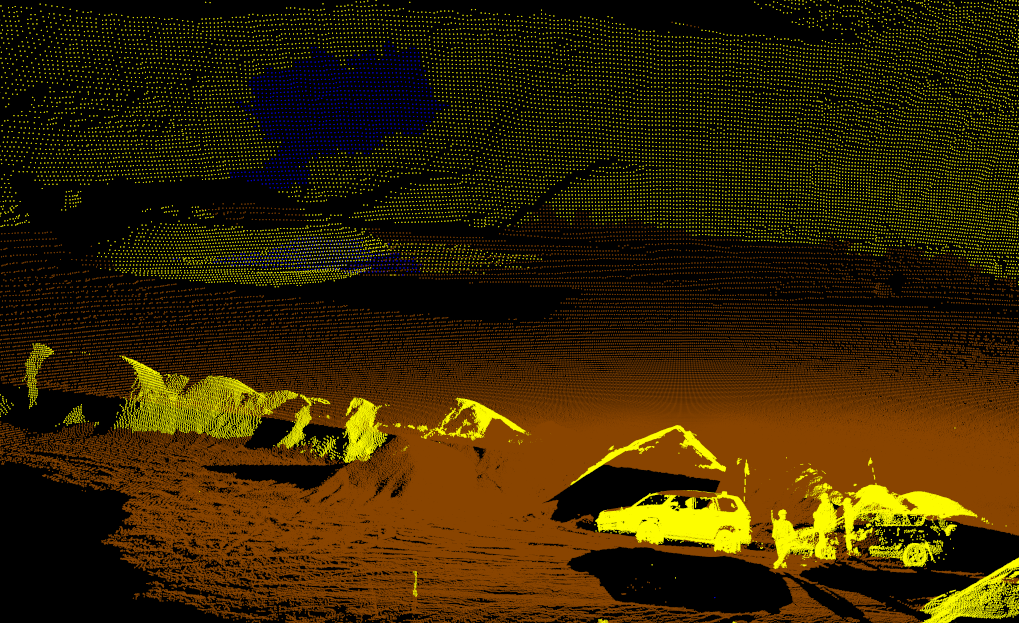}
         \caption{HDVNet: DTC}
         \label{fig:five over x}
     \end{subfigure}
     \begin{subfigure}[b]{0.33\textwidth}
         \centering
         \includegraphics[width=\textwidth]{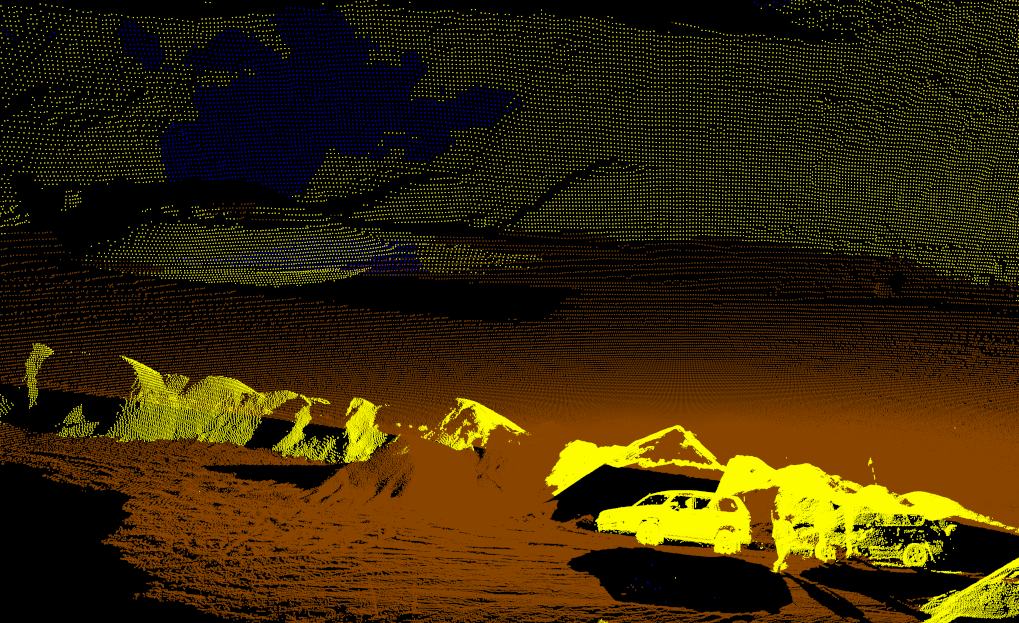}
         \caption{HDVNet: FCO}
         \label{fig:five over x}
     \end{subfigure}
     \begin{subfigure}[b]{0.33\textwidth}
         \centering
         \includegraphics[width=\textwidth]{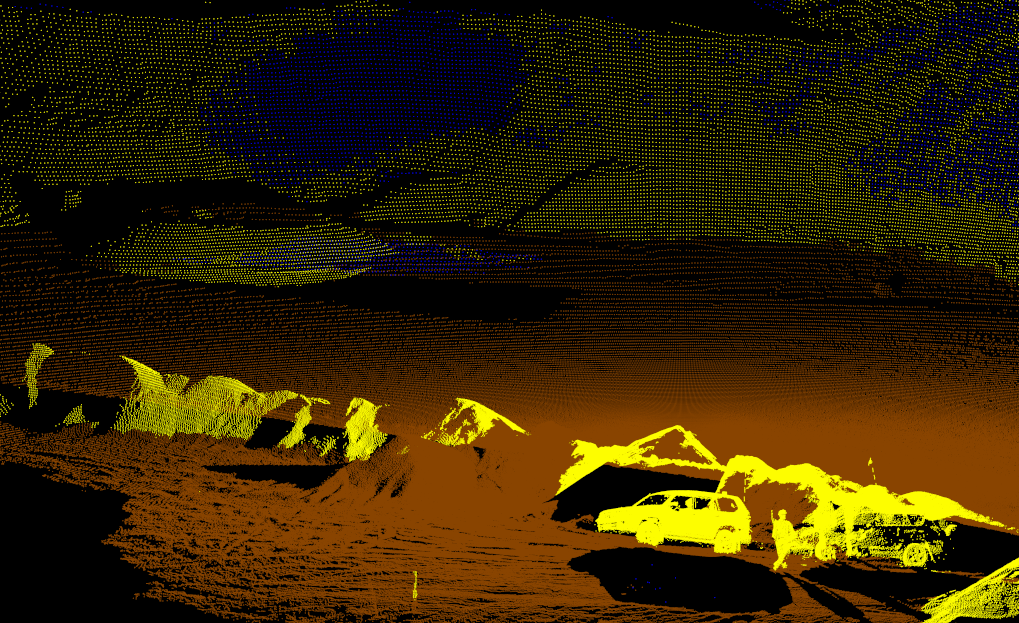}
         \caption{HDVNet: TCO}
         \label{fig:five over x}
     \end{subfigure}
     \begin{subfigure}[b]{0.33\textwidth}
         \centering
         \includegraphics[width=\textwidth]{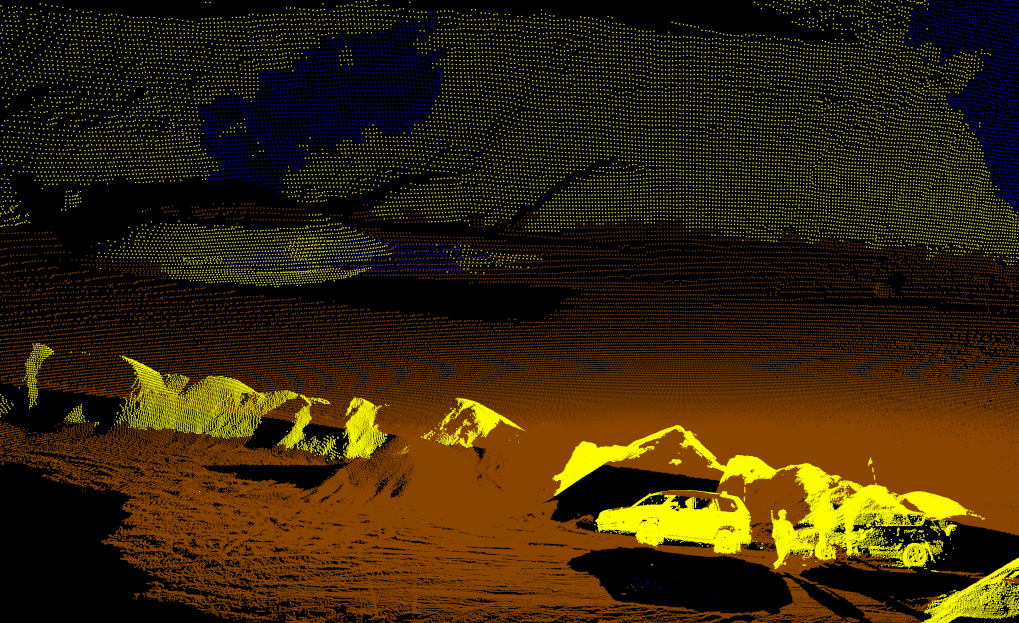}
         \caption{HDVNet: No FA}
         \label{fig:five over x}
     \end{subfigure}
     \begin{subfigure}[b]{0.33\textwidth}
         \centering
         \includegraphics[width=\textwidth]{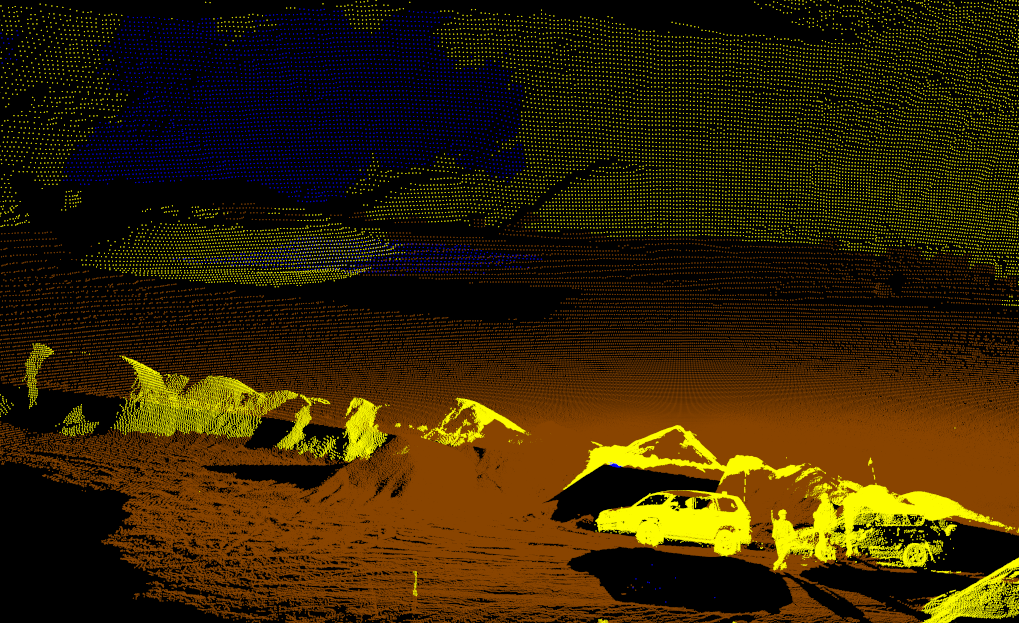}
         \caption{HDVNet: No FA (small)}
         \label{fig:five over x}
     \end{subfigure}
     \begin{subfigure}[b]{0.33\textwidth}
         \centering
         \includegraphics[width=\textwidth]{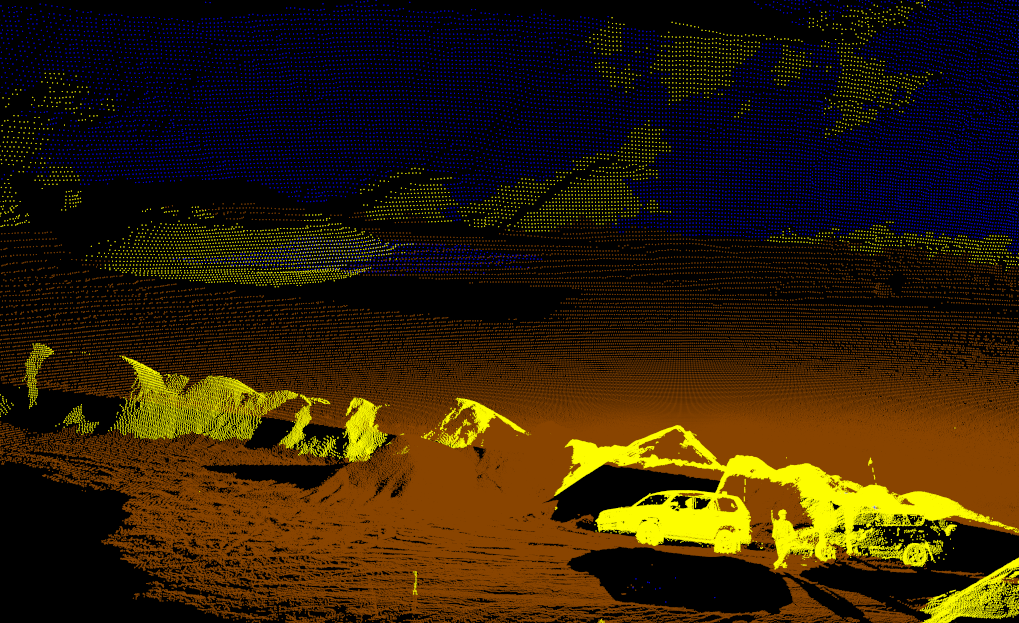}
         \caption{HDVNet}
         \label{fig:five over x}
     \end{subfigure}
     \begin{subfigure}[b]{0.33\textwidth}
         \centering
         \includegraphics[width=\textwidth]{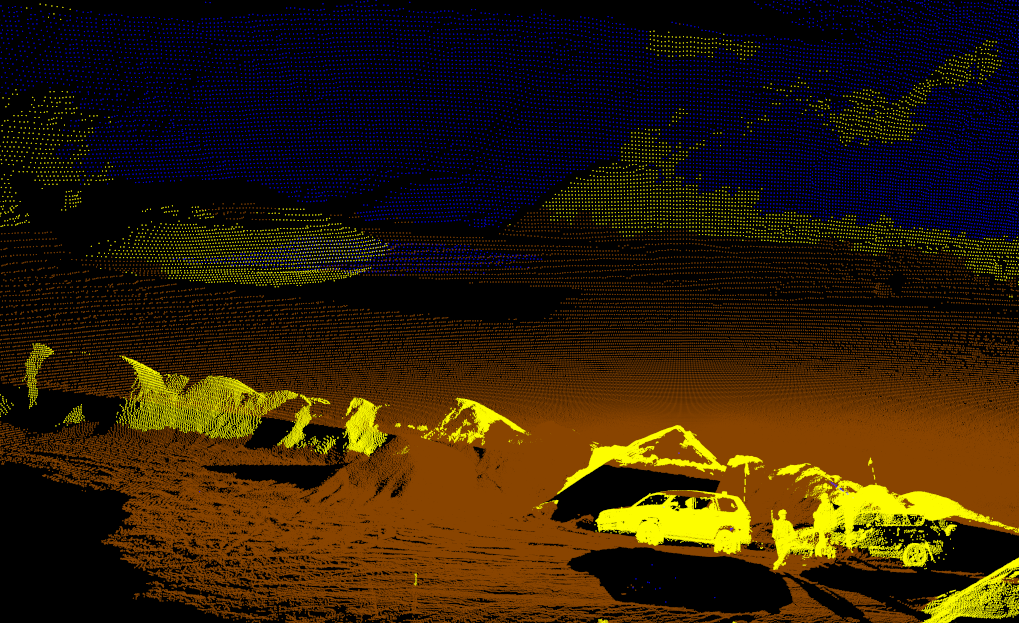}
         \caption{HDVNet: No ELFA}
         \label{fig:five over x}
     \end{subfigure}
        \caption{HDVMine qualitative results. Classes are : \fcolorbox{black}{Blue}{\rule{0pt}{6pt}\rule{6pt}{0pt}} Wall \fcolorbox{black}{Brown}{\rule{0pt}{6pt}\rule{6pt}{0pt}} Ground \fcolorbox{black}{Yellow}{\rule{0pt}{6pt}\rule{6pt}{0pt}} Other}
        \label{fig:HDVMineClassNear}
\end{figure*}

\begin{figure*}
     \centering
     \begin{subfigure}[b]{0.33\textwidth}
         \centering
         \includegraphics[width=\textwidth]{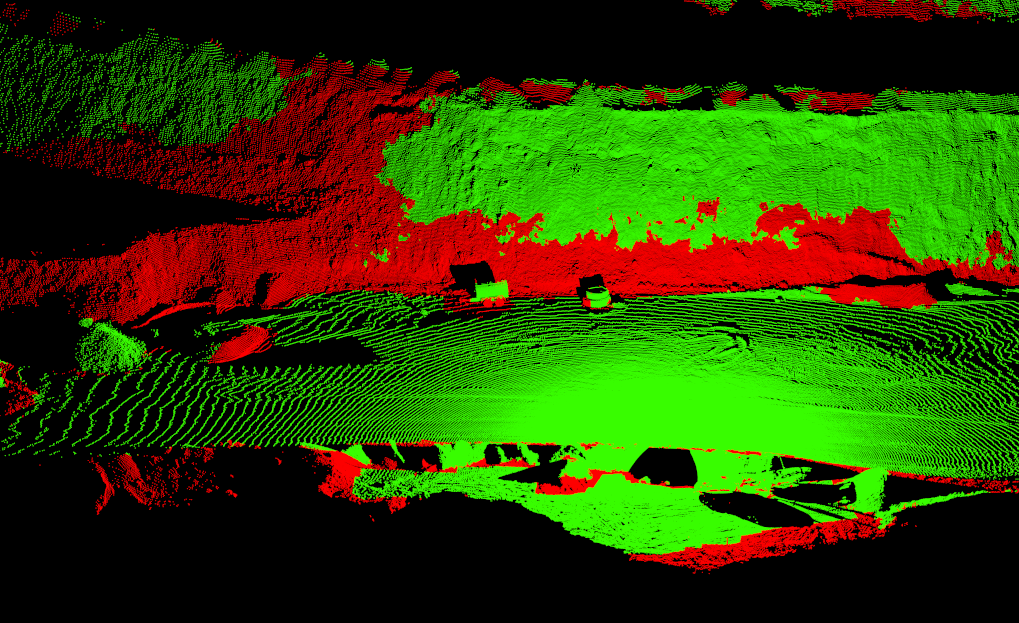}
         \caption{RandLA-Net}
         \label{fig:five over x}
     \end{subfigure}
     \begin{subfigure}[b]{0.33\textwidth}
         \centering
         \includegraphics[width=\textwidth]{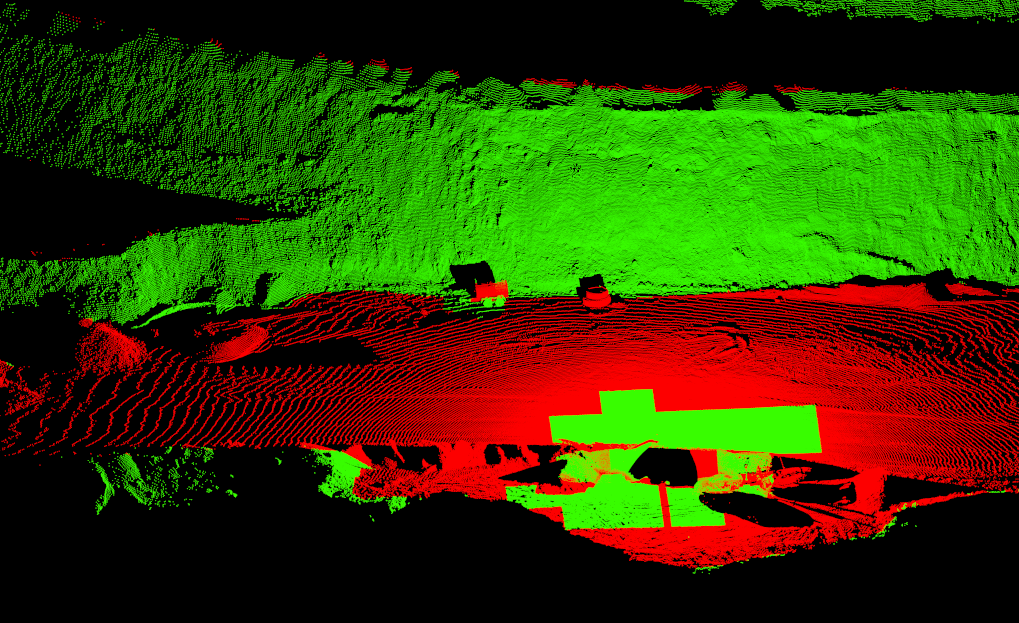}
         \caption{DGCNN}
         \label{fig:three sin x}
     \end{subfigure}
     \begin{subfigure}[b]{0.33\textwidth}
         \centering
         \includegraphics[width=\textwidth]{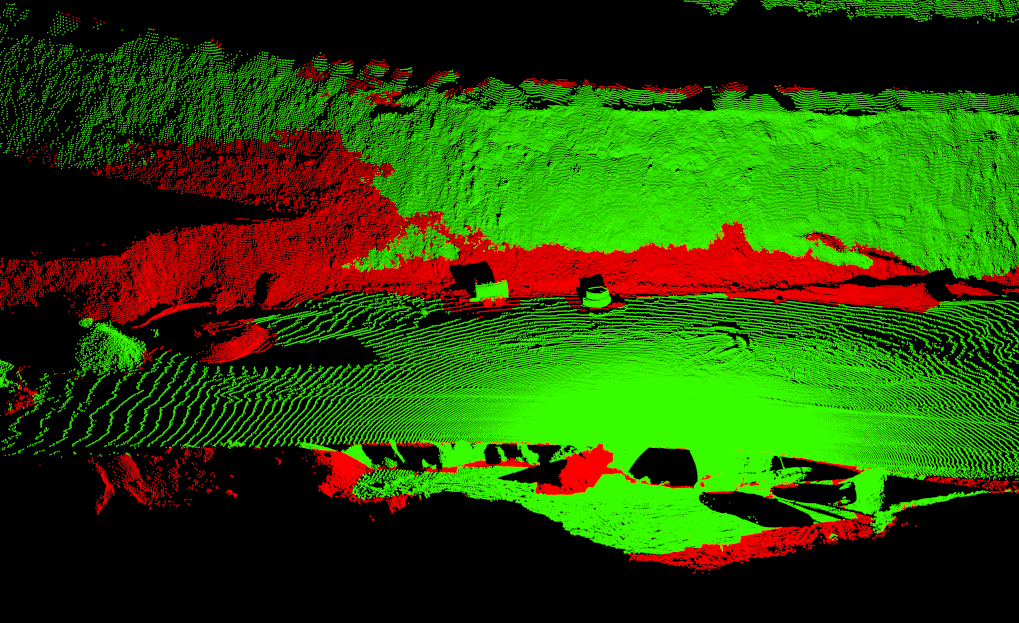}
         \caption{RandLA-Net + LN}
         \label{fig:five over x}
     \end{subfigure}
     \begin{subfigure}[b]{0.33\textwidth}
         \centering
         \includegraphics[width=\textwidth]{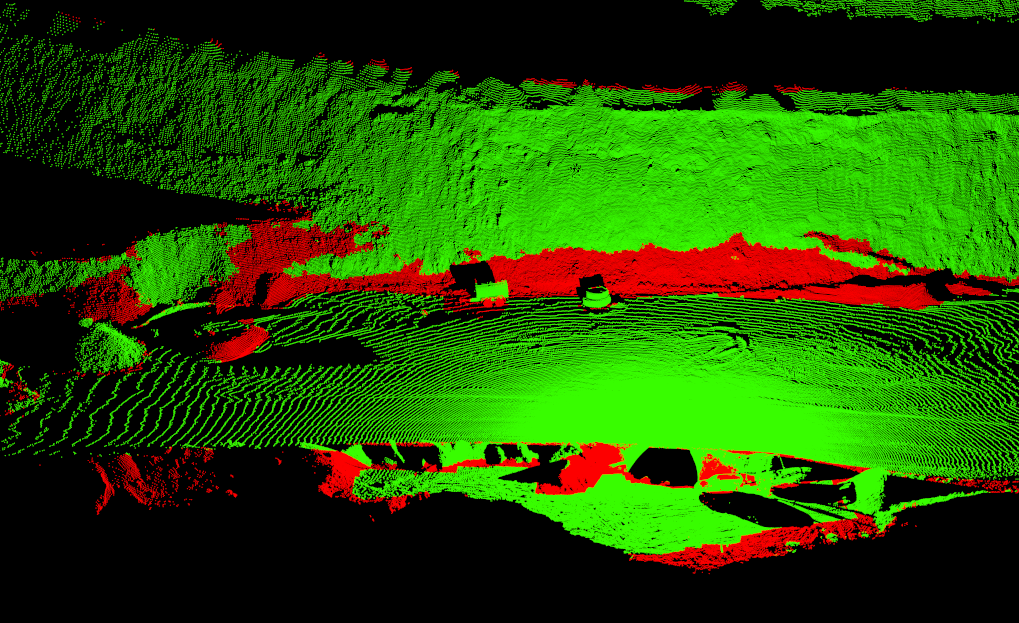}
         \caption{RandLA-Net + LN + LGS}
         \label{fig:five over x}
     \end{subfigure}
     \begin{subfigure}[b]{0.33\textwidth}
         \centering
         \includegraphics[width=\textwidth]{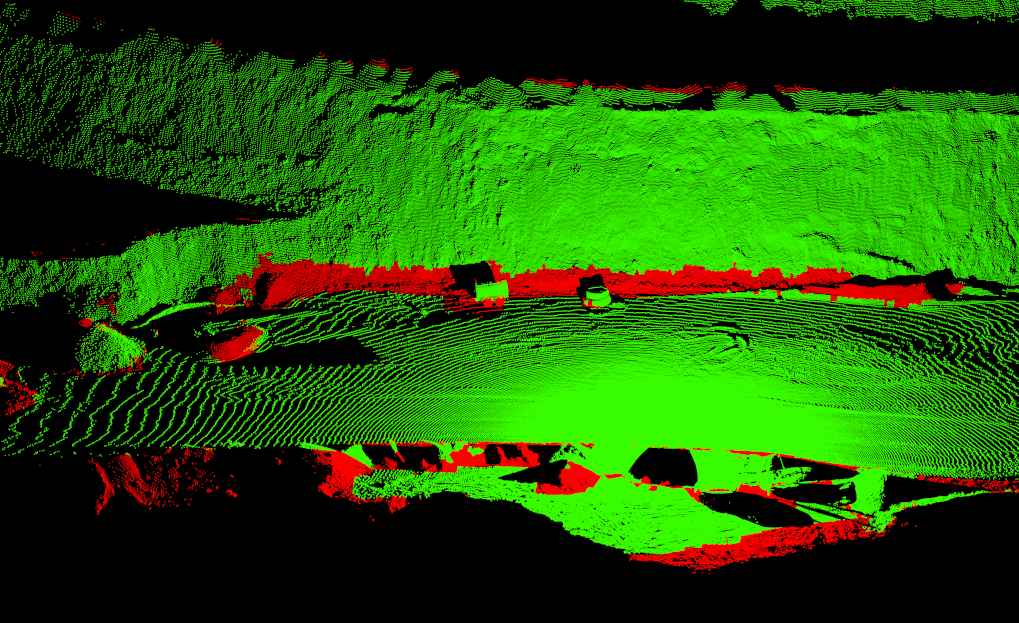}
         \caption{RandLA-Net + LN (Downsampled)}
         \label{fig:five over x}
     \end{subfigure}
     \begin{subfigure}[b]{0.33\textwidth}
         \centering
         \includegraphics[width=\textwidth]{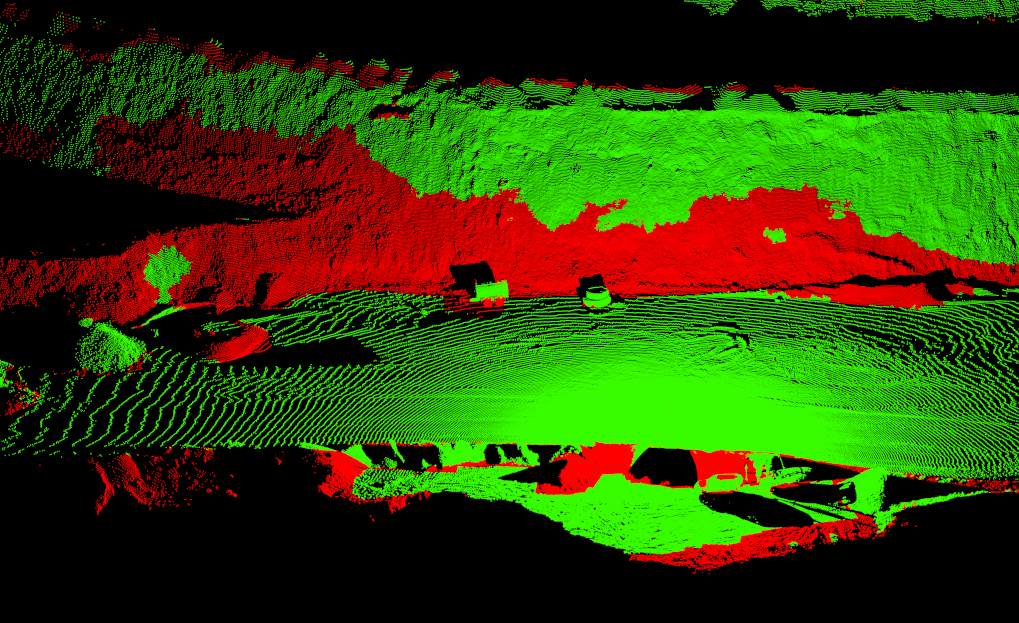}
         \caption{HDVNet: DTC}
         \label{fig:five over x}
     \end{subfigure}
     \begin{subfigure}[b]{0.33\textwidth}
         \centering
         \includegraphics[width=\textwidth]{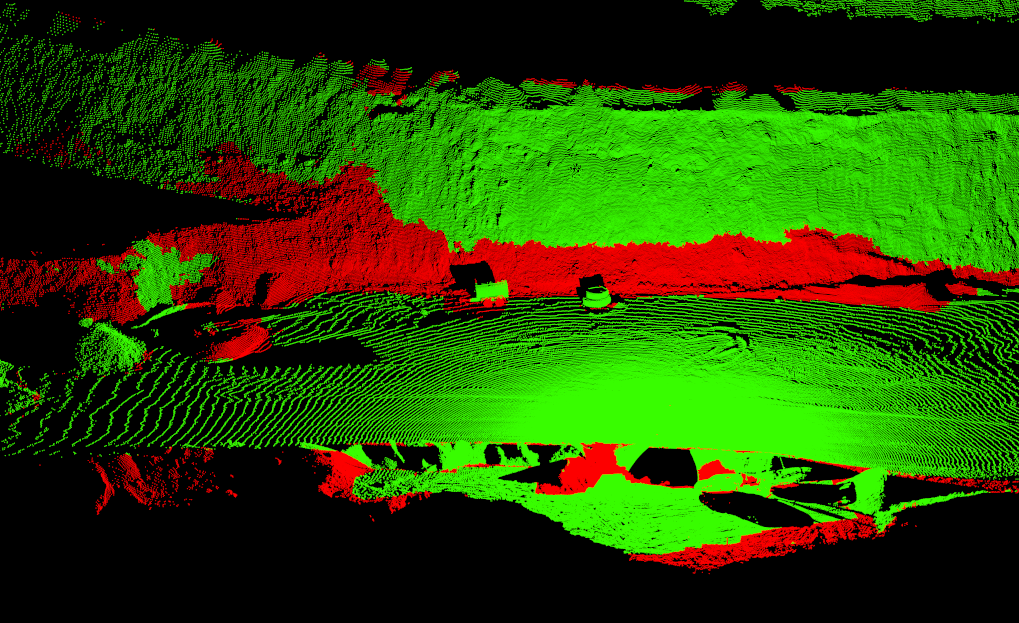}
         \caption{HDVNet: FCO}
         \label{fig:five over x}
     \end{subfigure}
     \begin{subfigure}[b]{0.33\textwidth}
         \centering
         \includegraphics[width=\textwidth]{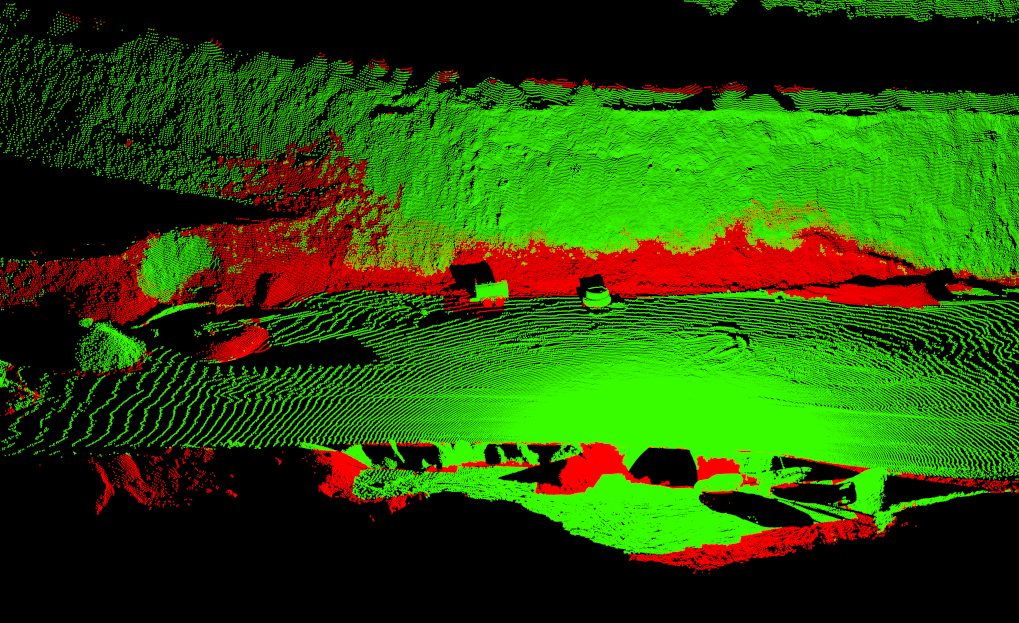}
         \caption{HDVNet: TCO}
         \label{fig:five over x}
     \end{subfigure}
     \begin{subfigure}[b]{0.33\textwidth}
         \centering
         \includegraphics[width=\textwidth]{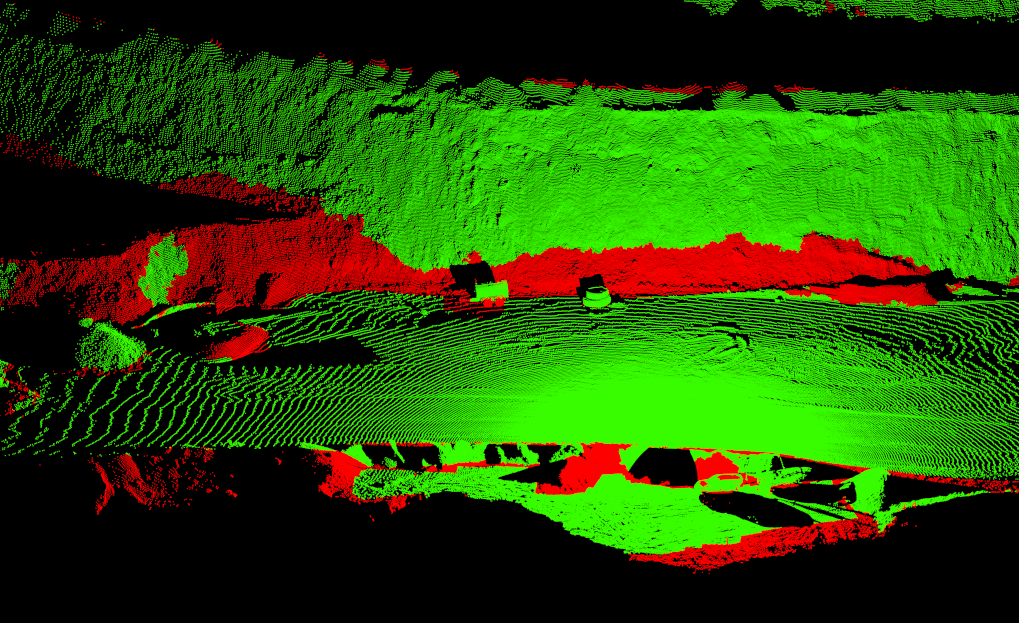}
         \caption{HDVNet: No FA}
         \label{fig:five over x}
     \end{subfigure}
     \begin{subfigure}[b]{0.33\textwidth}
         \centering
         \includegraphics[width=\textwidth]{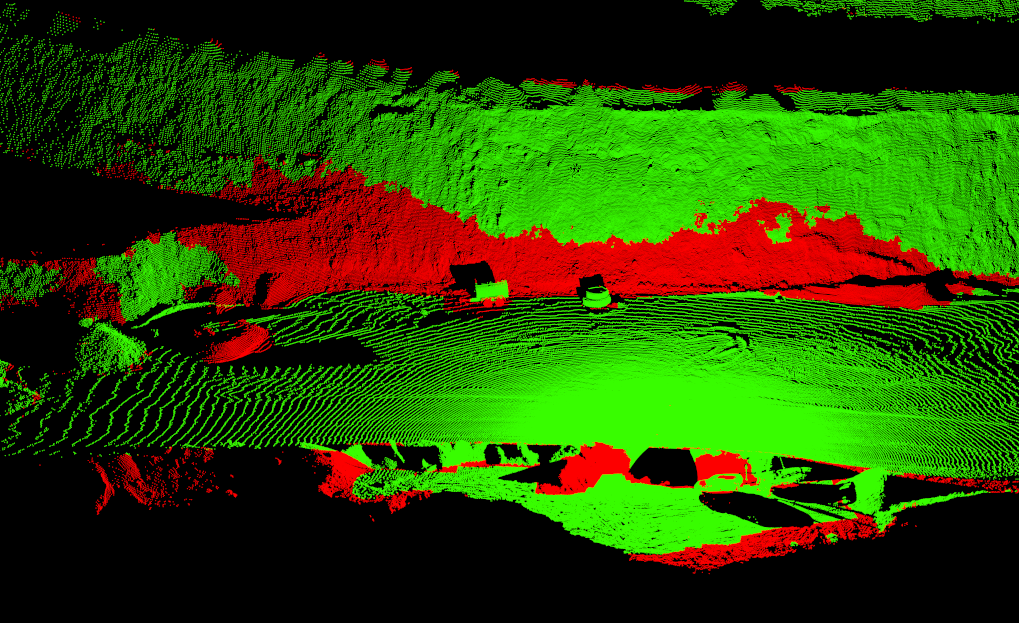}
         \caption{HDVNet: No FA (small)}
         \label{fig:five over x}
     \end{subfigure}
     \begin{subfigure}[b]{0.33\textwidth}
         \centering
         \includegraphics[width=\textwidth]{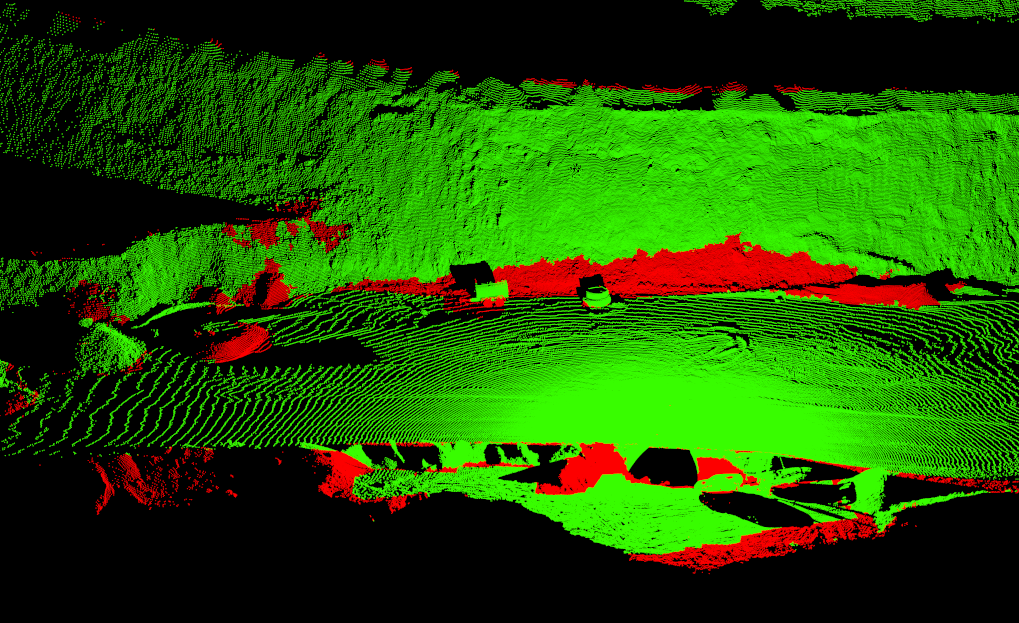}
         \caption{HDVNet}
         \label{fig:five over x}
     \end{subfigure}
     \begin{subfigure}[b]{0.33\textwidth}
         \centering
         \includegraphics[width=\textwidth]{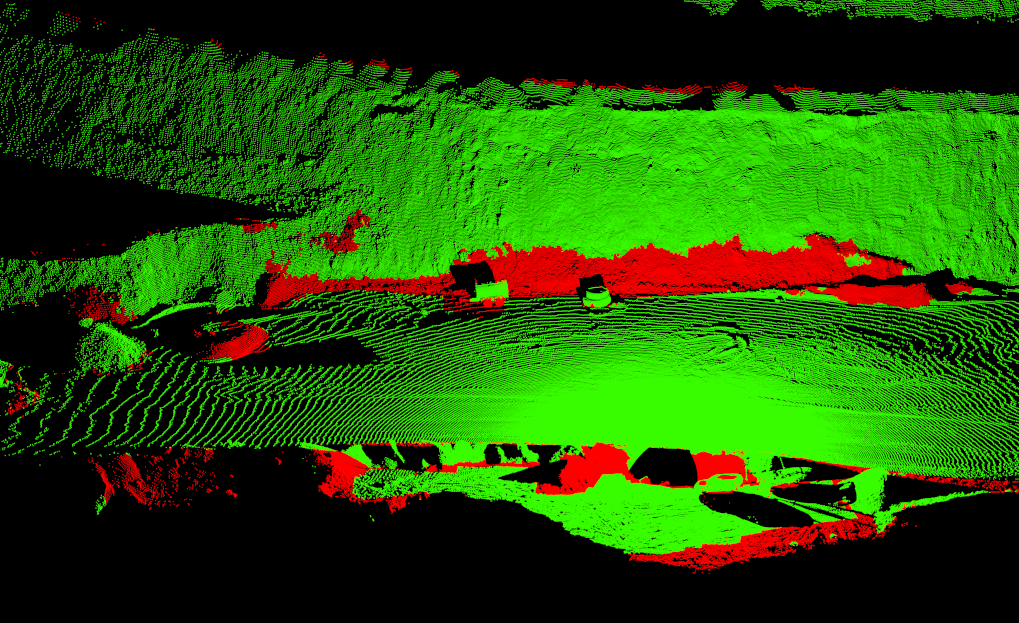}
         \caption{HDVNet: No ELFA}
         \label{fig:five over x}
     \end{subfigure}
        \caption{HDVMine qualitative results. Incorrect points are red, correct are green}
        \label{fig:HDVMineAccNear}
\end{figure*}

\begin{figure*}
     \centering
     \begin{subfigure}[b]{0.33\textwidth}
         \centering
         \includegraphics[width=\textwidth]{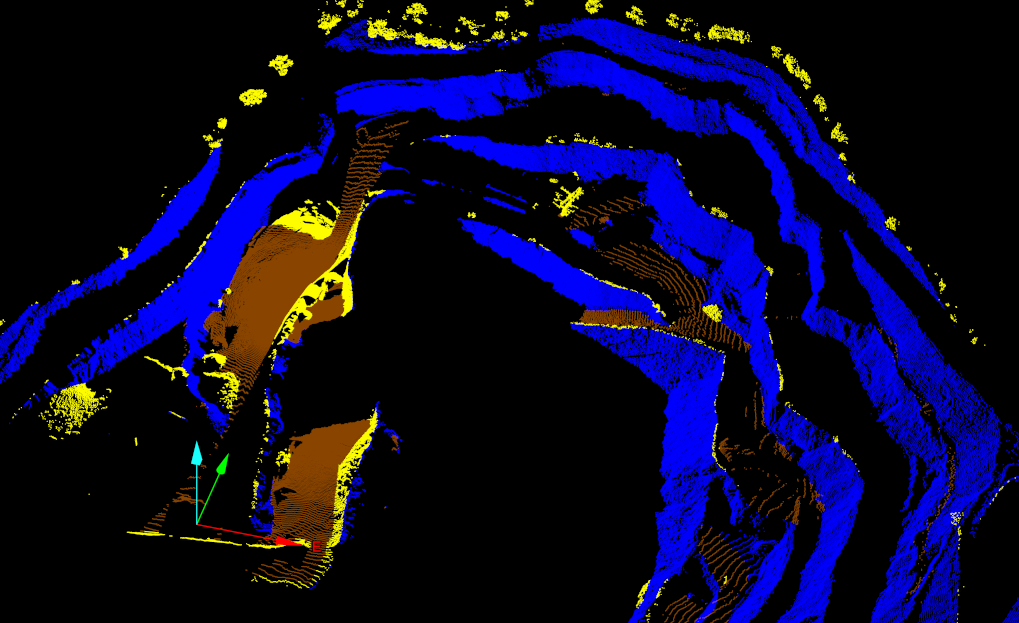}
         \caption{Ground Truth}
         \label{fig:y equals x}
     \end{subfigure}
     \begin{subfigure}[b]{0.33\textwidth}
         \centering
         \includegraphics[width=\textwidth]{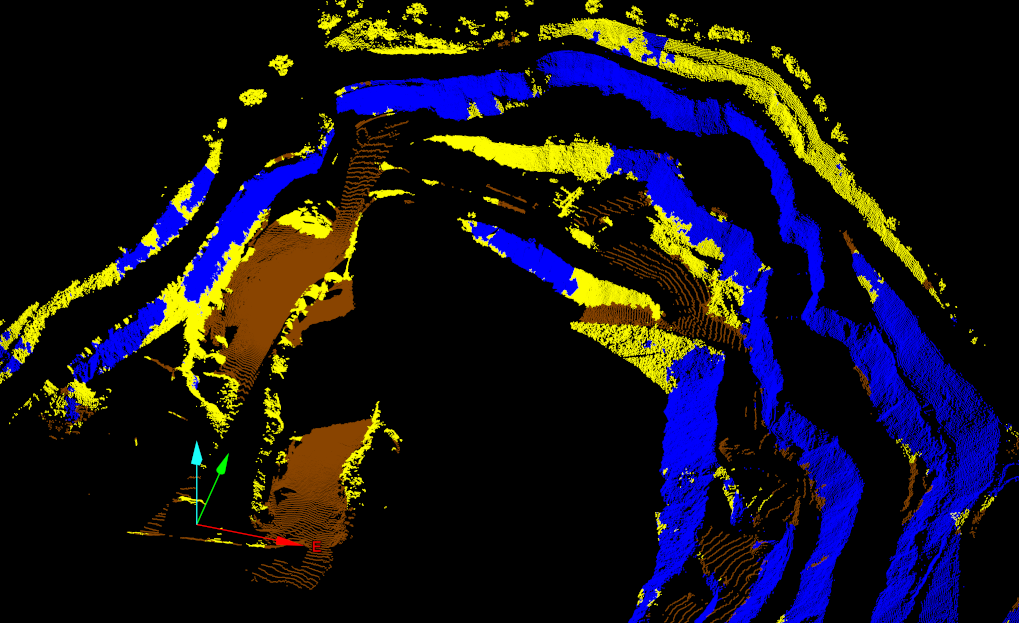}
         \caption{RandLA-Net}
         \label{fig:five over x}
     \end{subfigure}
     \begin{subfigure}[b]{0.33\textwidth}
         \centering
         \includegraphics[width=\textwidth]{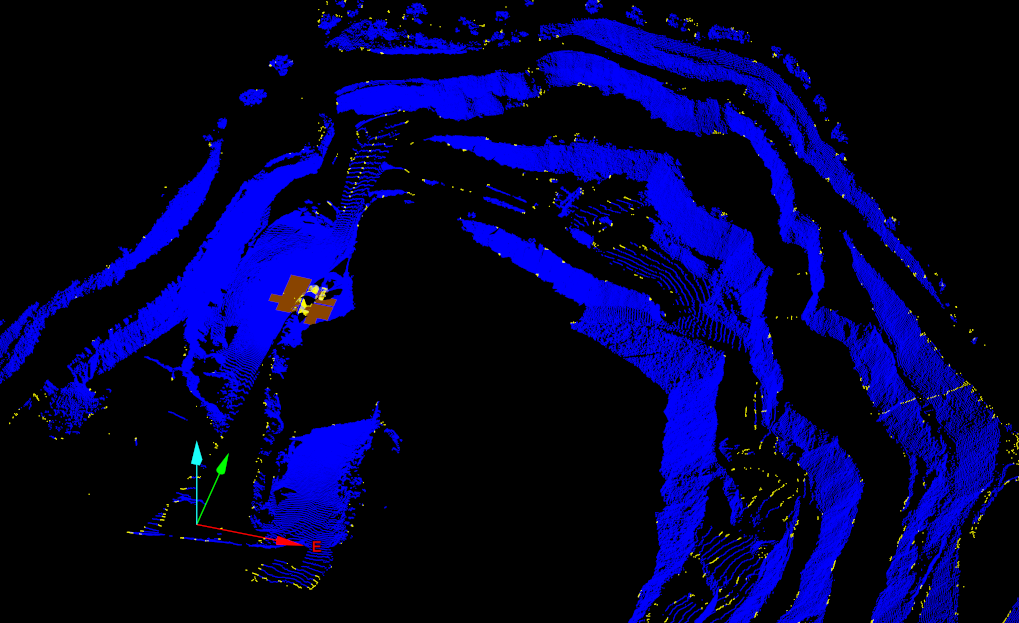}
         \caption{DGCNN}
         \label{fig:three sin x}
     \end{subfigure}
     \begin{subfigure}[b]{0.33\textwidth}
         \centering
         \includegraphics[width=\textwidth]{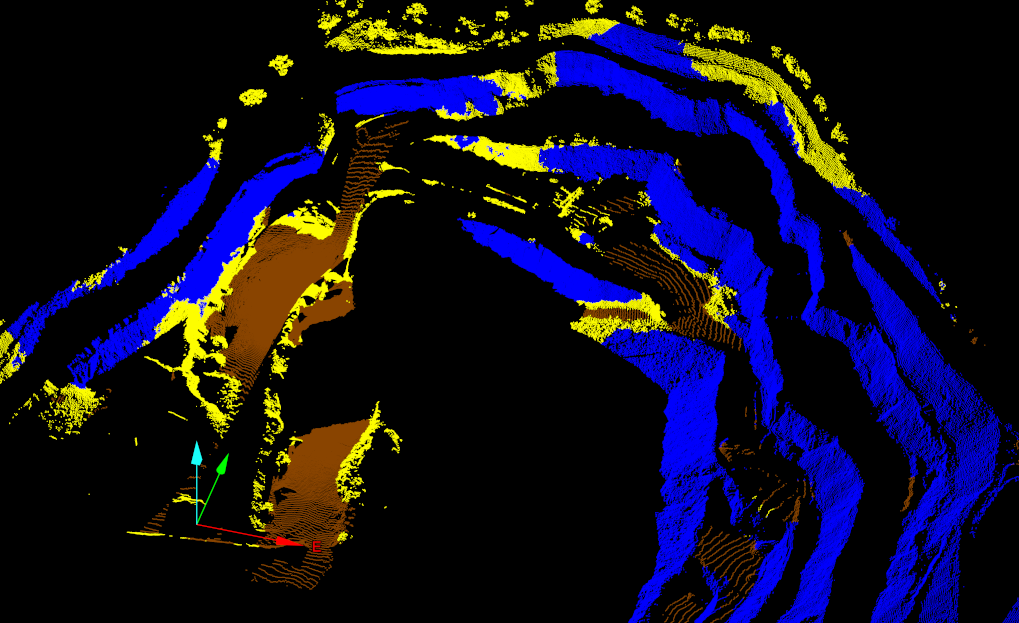}
         \caption{RandLA-Net + LN}
         \label{fig:five over x}
     \end{subfigure}
     \begin{subfigure}[b]{0.33\textwidth}
         \centering
         \includegraphics[width=\textwidth]{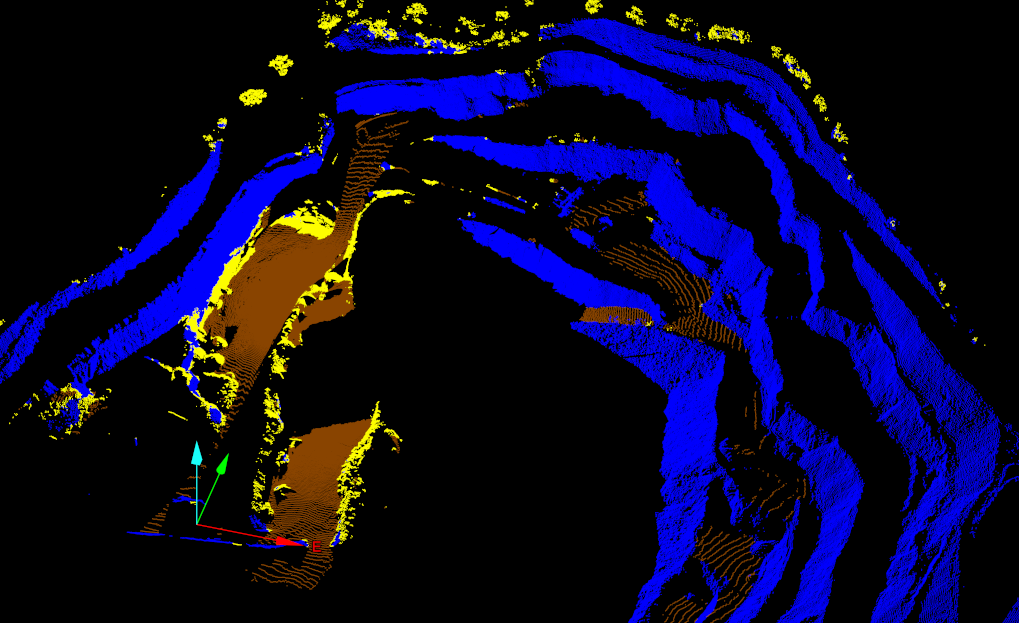}
         \caption{RandLA-Net + LN + LGS}
         \label{fig:five over x}
     \end{subfigure}
     \begin{subfigure}[b]{0.33\textwidth}
         \centering
         \includegraphics[width=\textwidth]{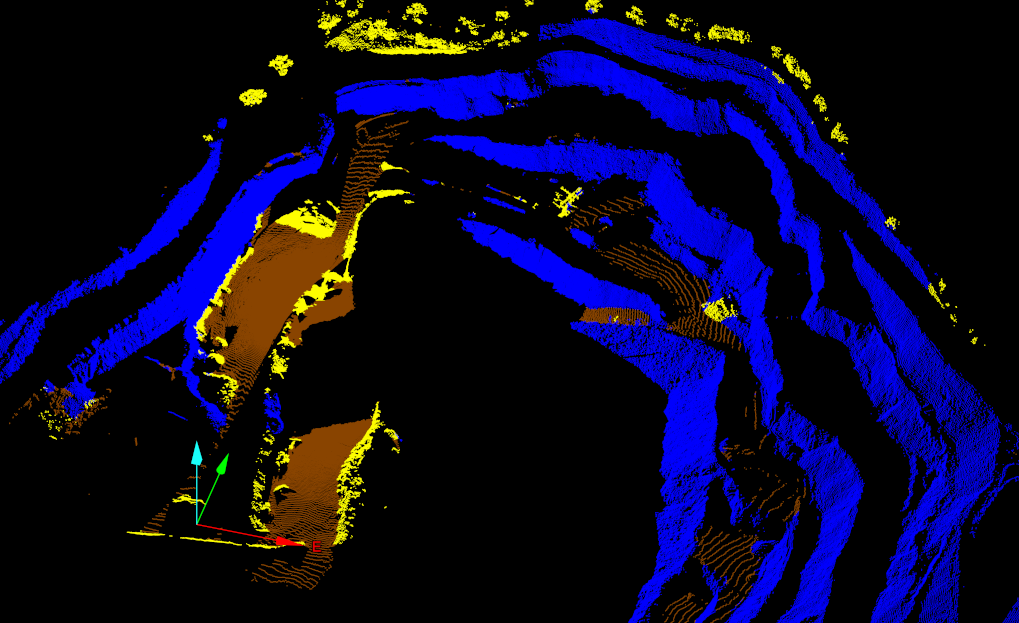}
         \caption{RandLA-Net + LN (Downsampled)}
         \label{fig:five over x}
     \end{subfigure}
     \begin{subfigure}[b]{0.33\textwidth}
         \centering
         \includegraphics[width=\textwidth]{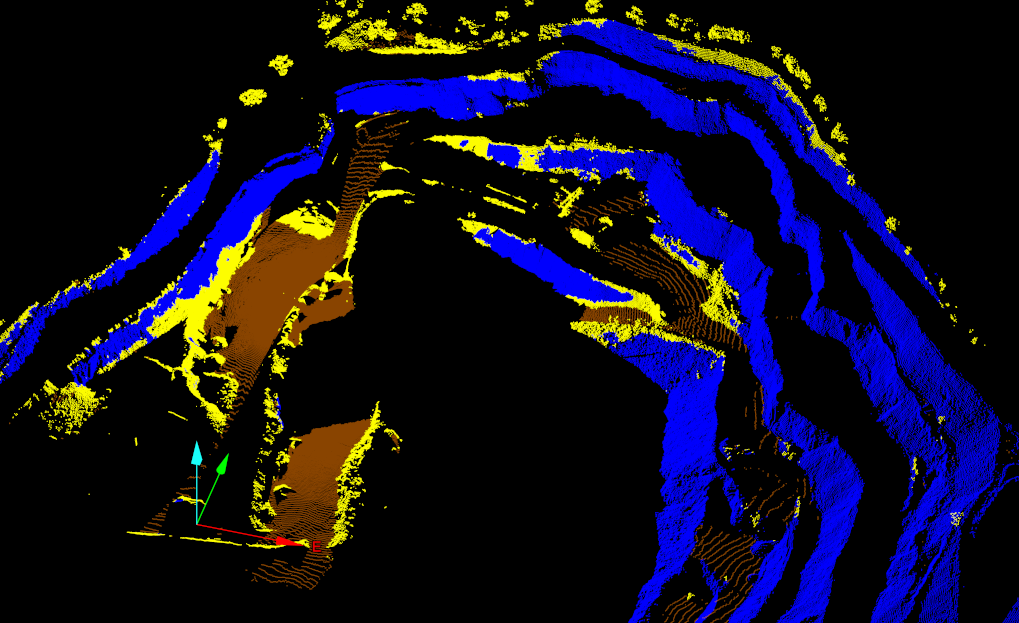}
         \caption{HDVNet: DTC}
         \label{fig:five over x}
     \end{subfigure}
     \begin{subfigure}[b]{0.33\textwidth}
         \centering
         \includegraphics[width=\textwidth]{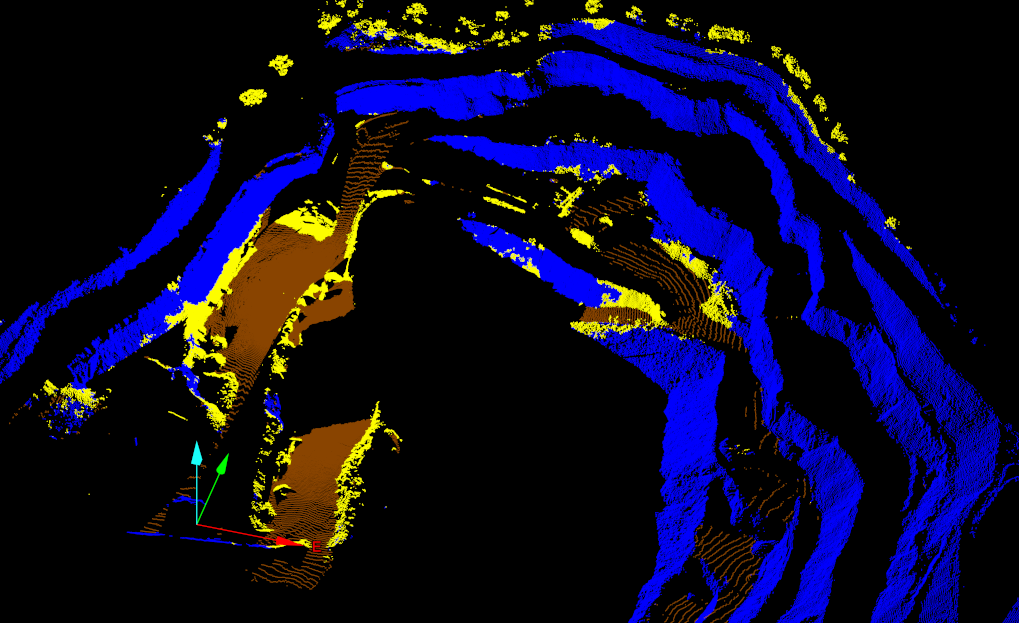}
         \caption{HDVNet: FCO}
         \label{fig:five over x}
     \end{subfigure}
     \begin{subfigure}[b]{0.33\textwidth}
         \centering
         \includegraphics[width=\textwidth]{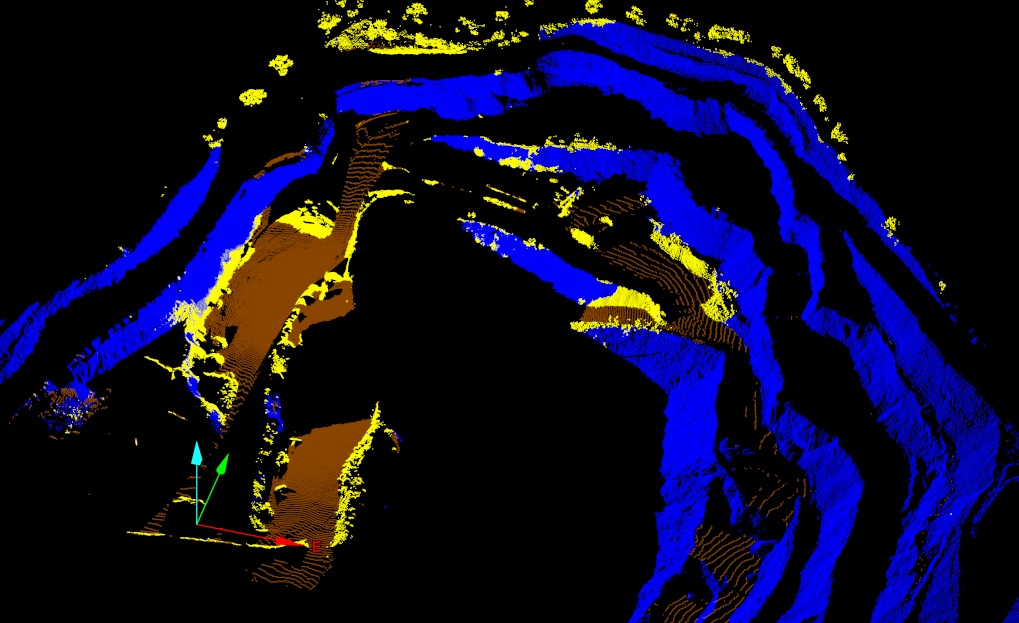}
         \caption{HDVNet: TCO}
         \label{fig:five over x}
     \end{subfigure}
     \begin{subfigure}[b]{0.33\textwidth}
         \centering
         \includegraphics[width=\textwidth]{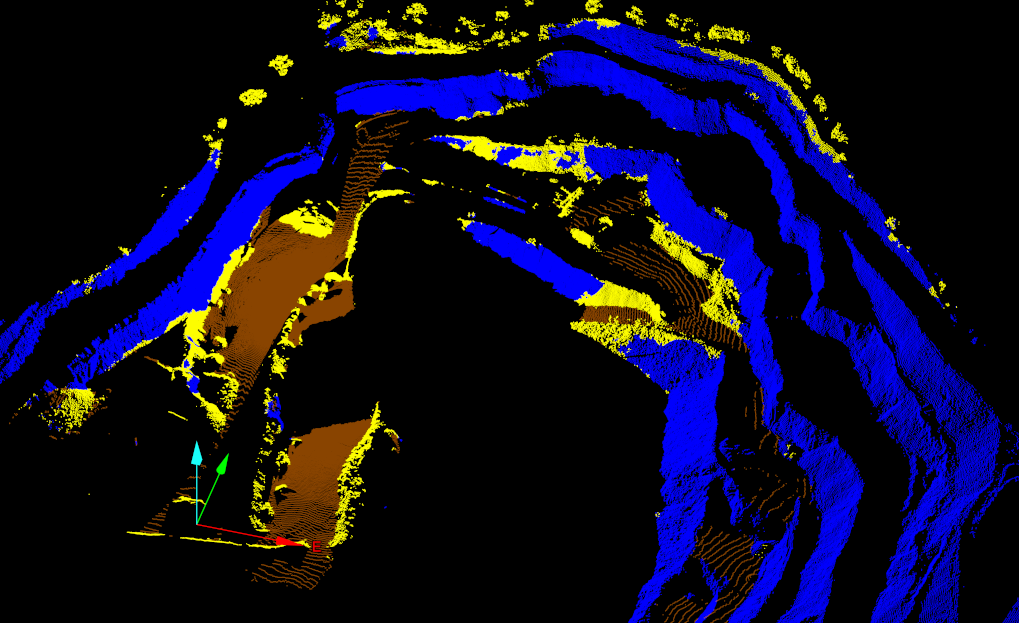}
         \caption{HDVNet: No FA}
         \label{fig:five over x}
     \end{subfigure}
     \begin{subfigure}[b]{0.33\textwidth}
         \centering
         \includegraphics[width=\textwidth]{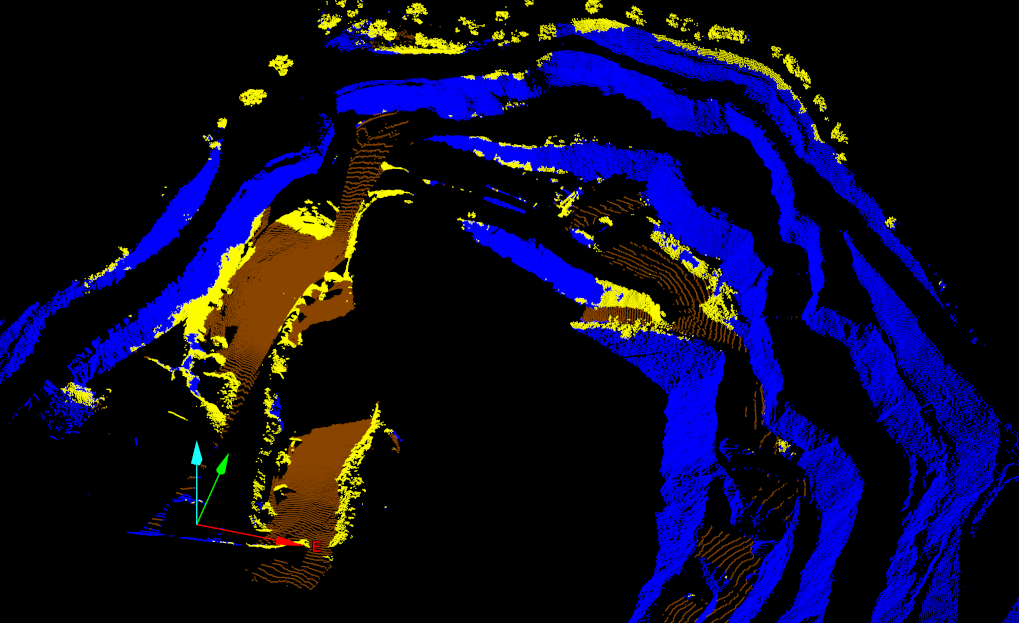}
         \caption{HDVNet: No FA (small)}
         \label{fig:five over x}
     \end{subfigure}
     \begin{subfigure}[b]{0.33\textwidth}
         \centering
         \includegraphics[width=\textwidth]{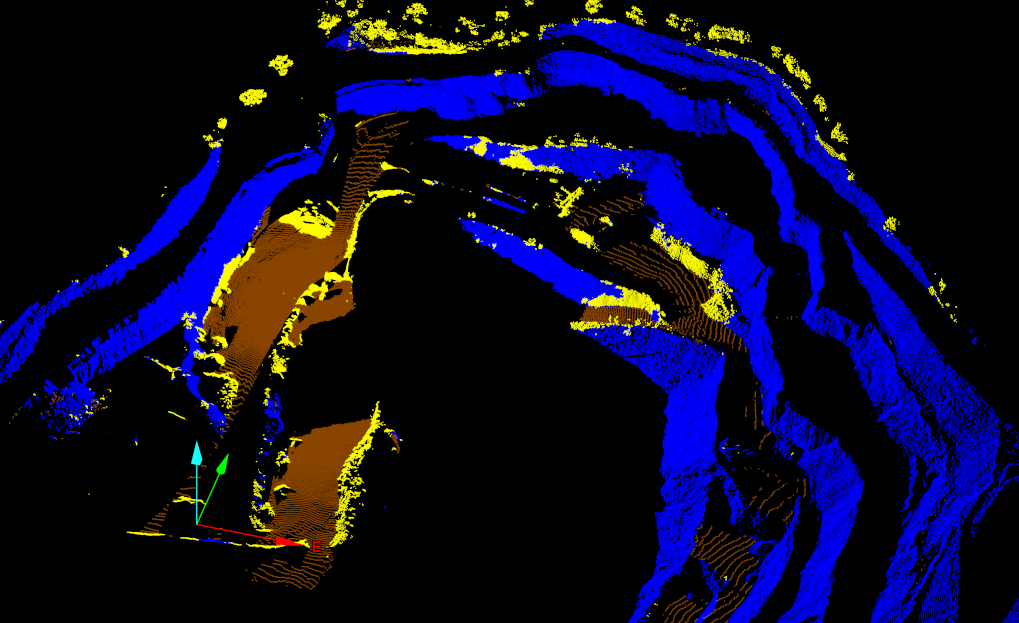}
         \caption{HDVNet}
         \label{fig:five over x}
     \end{subfigure}
     \begin{subfigure}[b]{0.33\textwidth}
         \centering
         \includegraphics[width=\textwidth]{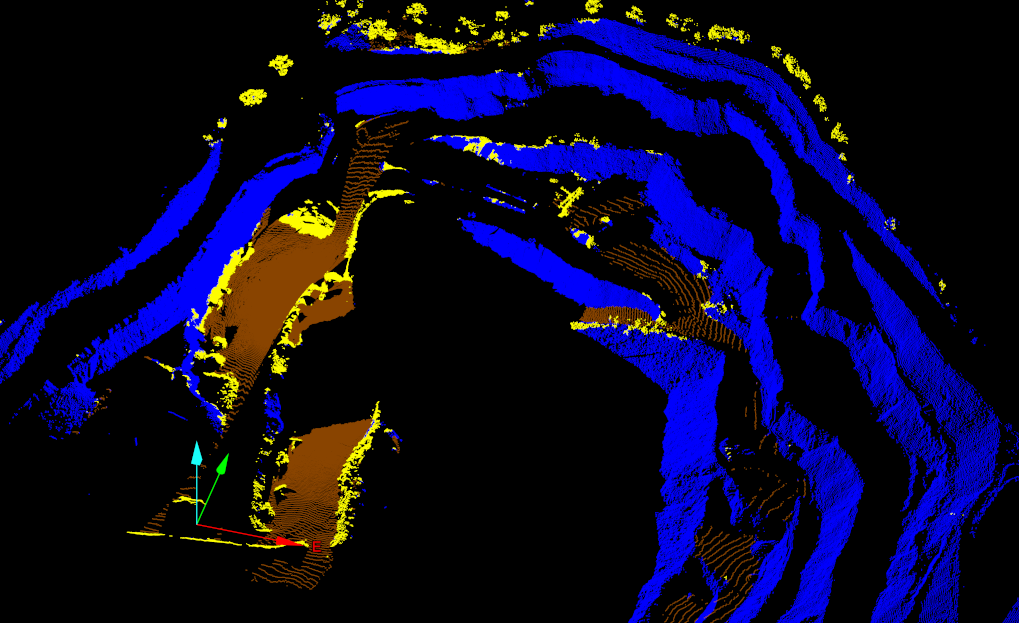}
         \caption{HDVNet: No ELFA}
         \label{fig:five over x}
     \end{subfigure}
        \caption{HDVMine qualitative results. Classes are : \fcolorbox{black}{Blue}{\rule{0pt}{6pt}\rule{6pt}{0pt}} Wall \fcolorbox{black}{Brown}{\rule{0pt}{6pt}\rule{6pt}{0pt}} Ground \fcolorbox{black}{Yellow}{\rule{0pt}{6pt}\rule{6pt}{0pt}} Other}
        \label{fig:HDVMineClassNear}
\end{figure*}

\begin{figure*}
     \centering
     \begin{subfigure}[b]{0.33\textwidth}
         \centering
         \includegraphics[width=\textwidth]{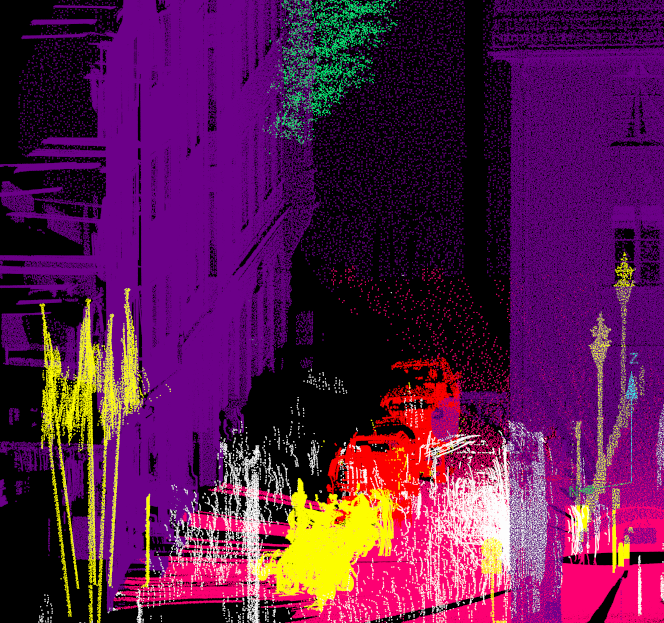}
         \caption{Ground Truth}
         \label{fig:y equals x}
     \end{subfigure}
     \begin{subfigure}[b]{0.33\textwidth}
         \centering
         \includegraphics[width=\textwidth]{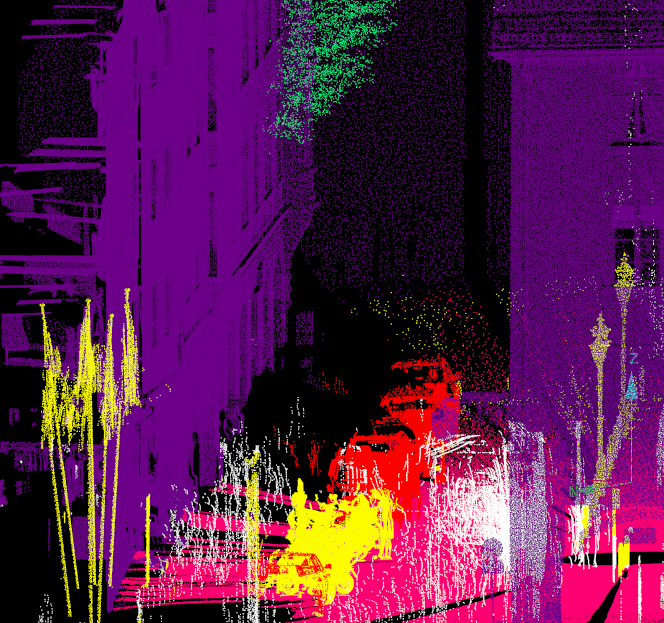}
         \caption{RandLA-Net}
         \label{fig:five over x}
     \end{subfigure}
     \begin{subfigure}[b]{0.33\textwidth}
         \centering
         \includegraphics[width=\textwidth]{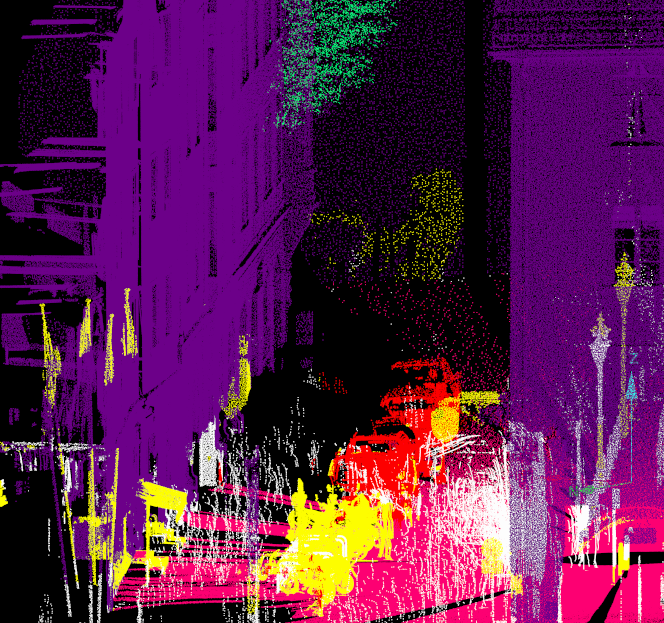}
         \caption{HDVNet}
         \label{fig:three sin x}
     \end{subfigure}
     \begin{subfigure}[b]{0.33\textwidth}
         \centering
         \includegraphics[width=\textwidth]{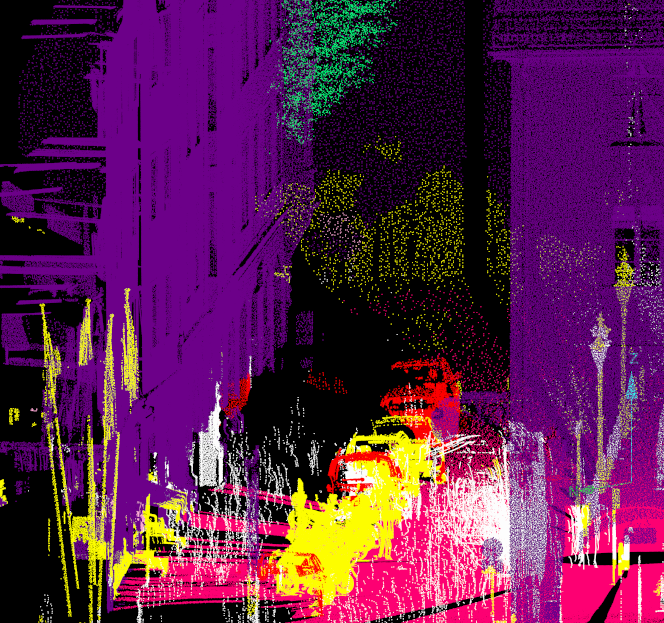}
         \caption{HDVNet: No ELFA}
         \label{fig:five over x}
     \end{subfigure}
        \caption{Semantic3D qualitative results. Classes are : \hspace{0.3cm}  \fcolorbox{black}{RubineRed}{\rule{0pt}{4pt}\rule{4pt}{0pt}} Man-made Terrain \fcolorbox{black}{SkyBlue}{\rule{0pt}{4pt}\rule{4pt}{0pt}} Natural Terrain \fcolorbox{black}{LimeGreen}{\rule{0pt}{4pt}\rule{4pt}{0pt}} High Vegetation \fcolorbox{black}{CarnationPink}{\rule{0pt}{4pt}\rule{4pt}{0pt}} Low Vegetation \newline \hspace*{6.66cm} \fcolorbox{black}{violet}{\rule{0pt}{4pt}\rule{4pt}{0pt}} Buildings \fcolorbox{black}{Yellow}{\rule{0pt}{4pt}\rule{4pt}{0pt}} Hardscape \fcolorbox{black}{White}{\rule{0pt}{4pt}\rule{4pt}{0pt}} Scanning Artefacts \fcolorbox{black}{Red}{\rule{0pt}{4pt}\rule{4pt}{0pt}} Cars}
        \label{fig:Sem3DClassNear}
\end{figure*}

\begin{figure*}
     \centering
     \begin{subfigure}[b]{0.33\textwidth}
         \centering
         \includegraphics[width=\textwidth]{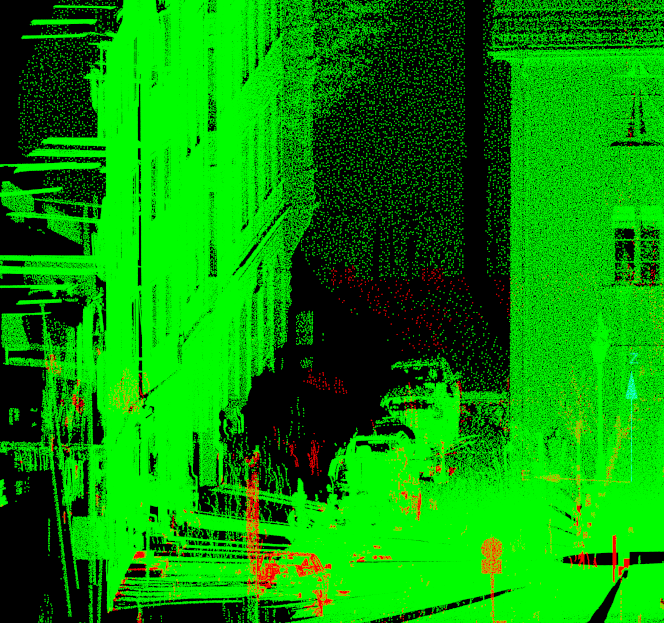}
         \caption{RandLA-Net}
         \label{fig:five over x}
     \end{subfigure}
     \begin{subfigure}[b]{0.33\textwidth}
         \centering
         \includegraphics[width=\textwidth]{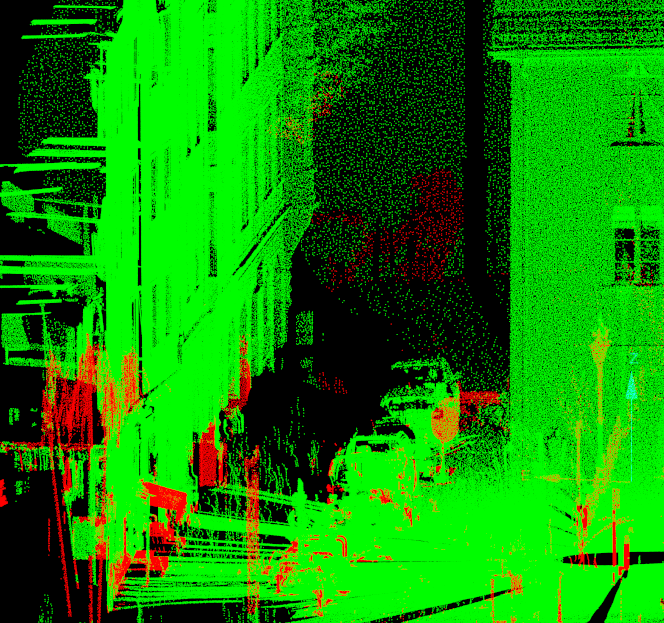}
         \caption{HDVNet}
         \label{fig:three sin x}
     \end{subfigure}
     \begin{subfigure}[b]{0.33\textwidth}
         \centering
         \includegraphics[width=\textwidth]{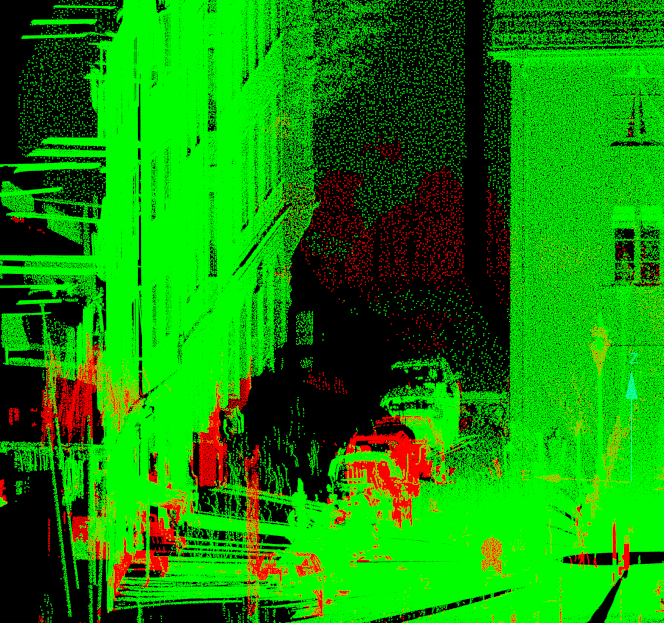}
         \caption{HDVNet: No ELFA}
         \label{fig:five over x}
     \end{subfigure}
         \caption{Semantic3D qualitative results. Incorrect points are red, correct are green}
        \label{fig:Sem3DAccNear}
\end{figure*}

\begin{figure}[!t]
     \centering
     \begin{subfigure}[b]{0.45\columnwidth}
         \centering
         \includegraphics[width=1\columnwidth]{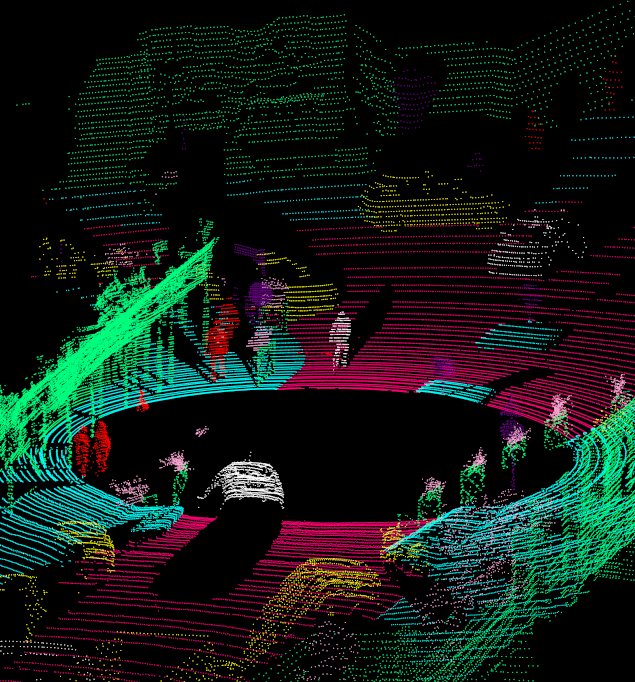}
         \caption{Ground Truth}
         \label{fig:y equals x}
     \end{subfigure}
     \begin{subfigure}[b]{0.45\columnwidth}
         \centering
         \includegraphics[width=1\columnwidth]{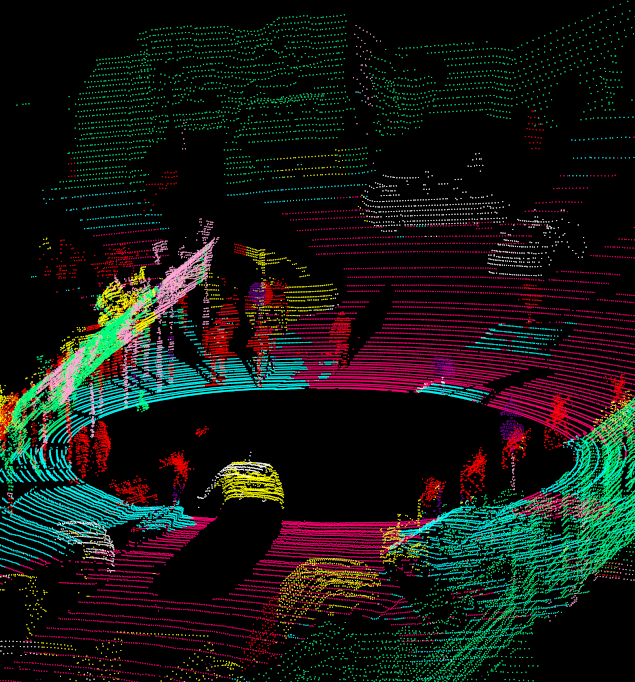}
         \caption{RandLA-Net}
         \label{fig:five over x}
     \end{subfigure}
     \begin{subfigure}[b]{0.45\columnwidth}
         \centering
         \includegraphics[width=1\columnwidth]{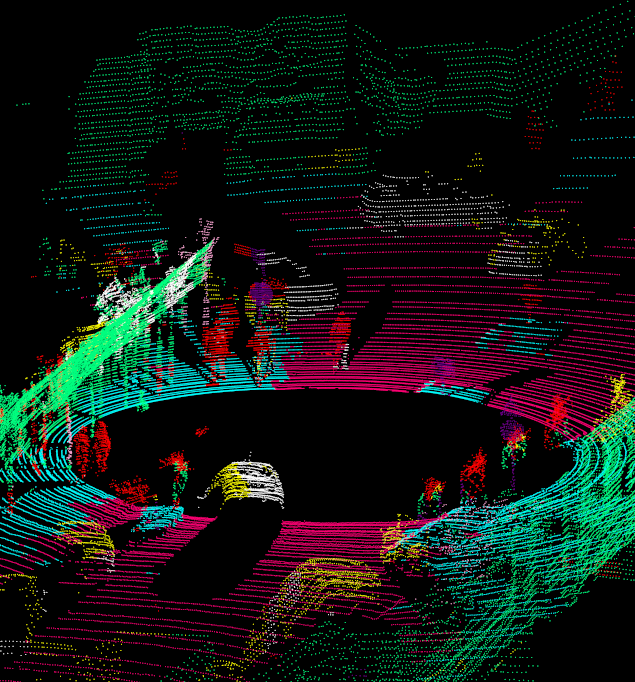}
         \caption{HDVNet}
         \label{fig:three sin x}
     \end{subfigure}
     \begin{subfigure}[b]{0.45\columnwidth}
         \centering
         \includegraphics[width=1\columnwidth]{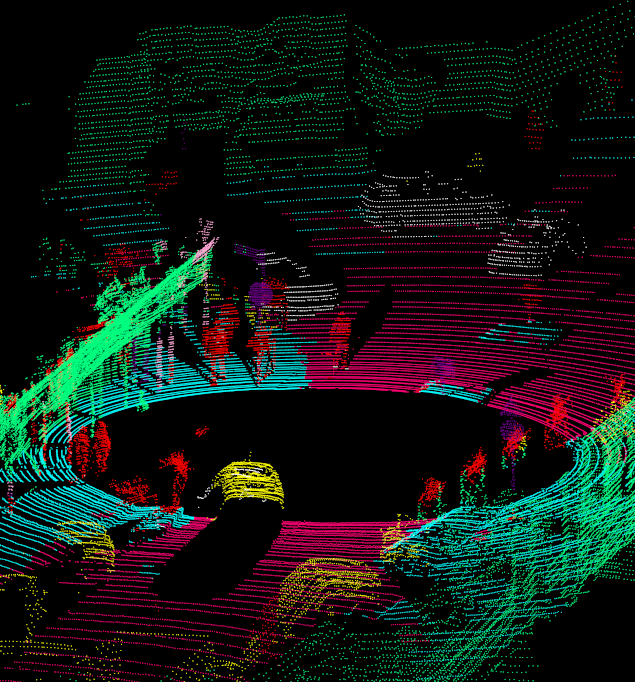}
         \caption{HDVNet: No ELFA}
         \label{fig:five over x}
     \end{subfigure}
     \begin{subfigure}[b]{0.45\columnwidth}
         \centering
         \includegraphics[width=1\columnwidth]{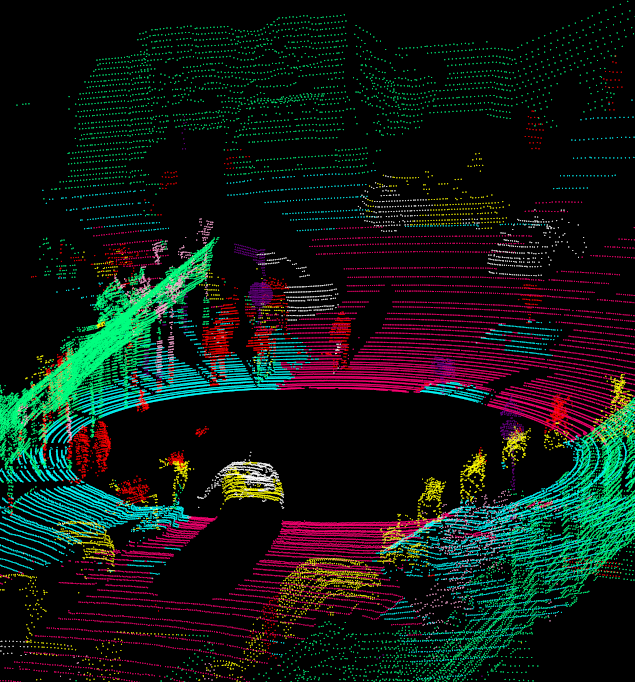}
         \caption{HDVNet: Limited FA}
         \label{fig:five over x}
     \end{subfigure}
        \caption{HelixNet qualitative results. Classes are : \newline \newline \fcolorbox{black}{RubineRed}{\rule{0pt}{4pt}\rule{4pt}{0pt}} Road \fcolorbox{black}{SkyBlue}{\rule{0pt}{4pt}\rule{4pt}{0pt}} Other Surface \fcolorbox{black}{LimeGreen}{\rule{0pt}{4pt}\rule{4pt}{0pt}} Building \fcolorbox{black}{CarnationPink}{\rule{0pt}{4pt}\rule{4pt}{0pt}} Vegetation \newline \fcolorbox{black}{violet}{\rule{0pt}{4pt}\rule{4pt}{0pt}} Traffic Sign \fcolorbox{black}{Yellow}{\rule{0pt}{4pt}\rule{4pt}{0pt}} Static Vehicle \fcolorbox{black}{White}{\rule{0pt}{4pt}\rule{4pt}{0pt}} Moving Vehicle \newline  \fcolorbox{black}{Red}{\rule{0pt}{4pt}\rule{4pt}{0pt}} Pedestrian \fcolorbox{black}{Brown}{\rule{0pt}{4pt}\rule{4pt}{0pt}} Artefact}
        \label{fig:three graphs}
\end{figure}

\begin{figure}[!t]\centering
     \centering
     \begin{subfigure}[b]{0.45\columnwidth}
         \centering
         \includegraphics[width=1\columnwidth]{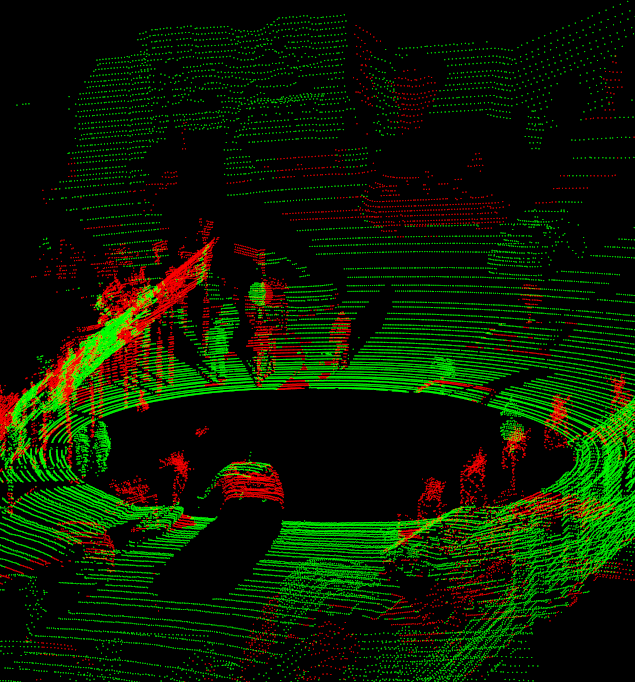}
         \caption{RandLA-Net}
         \label{fig:five over x}
     \end{subfigure}
     \begin{subfigure}[b]{0.45\columnwidth}
         \centering
         \includegraphics[width=1\columnwidth]{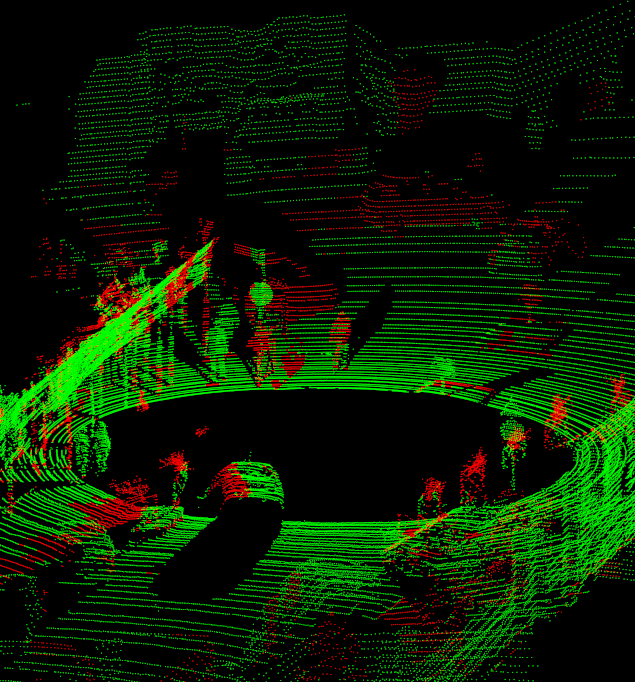}
         \caption{HDVNet}
         \label{fig:three sin x}
     \end{subfigure}
     \begin{subfigure}[b]{0.45\columnwidth}
         \centering
         \includegraphics[width=1\columnwidth]{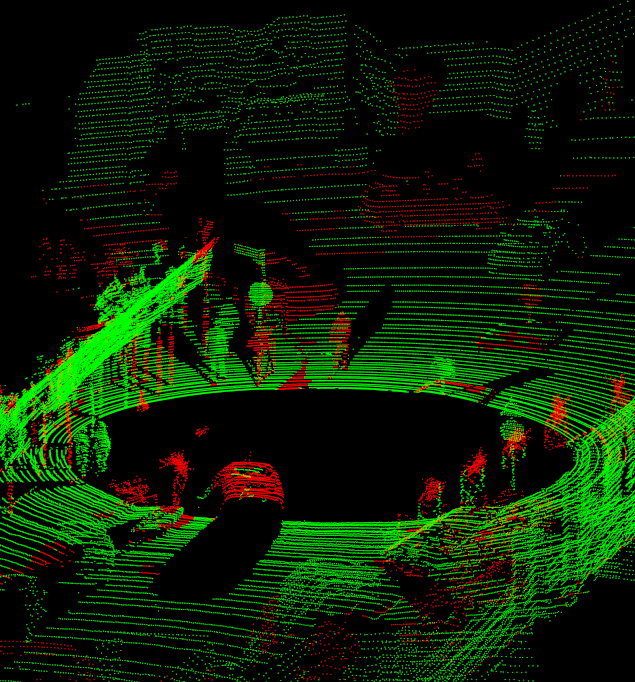}
         \caption{HDVNet: No ELFA}
         \label{fig:five over x}
     \end{subfigure}
     \begin{subfigure}[b]{0.45\columnwidth}
         \centering
         \includegraphics[width=1\columnwidth]{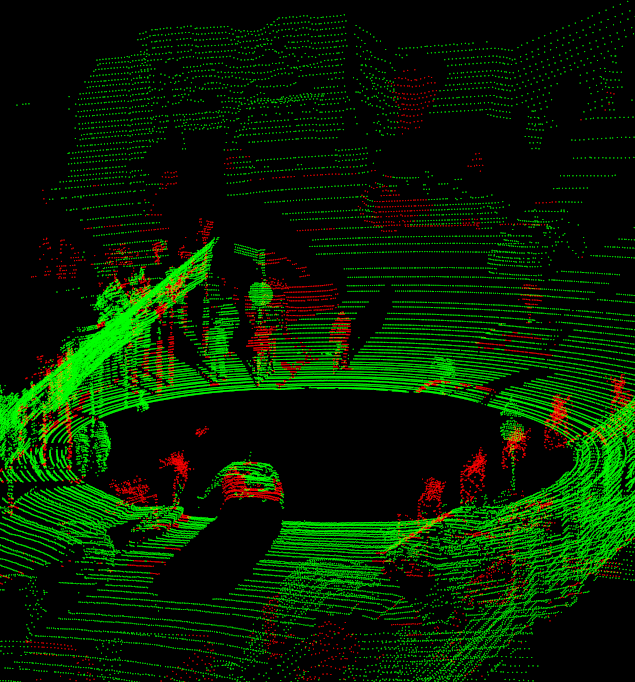}
         \caption{HDVNet: Limited FA}
         \label{fig:five over x}
     \end{subfigure}
         \caption{HelixNet qualitative results. Incorrect points are red, correct are green}
        \label{fig:Sem3DAccNear}
\end{figure}


\newpage

\section{Conclusions}
In this paper we introduced the novel network architecture HDVMine for direct point cloud segmentation. We demonstrated improved performance consistent across all densities on data with high density variation, such as that from large-scale land-surveying or mining. The measures ingrained into the architecture were each tested separately in an ablation study to confirm their individual contributions to the final results. 

We confirmed that this performance benefit does not translate to more homogeneous terrestrial LiDAR data such as Semantic3D, and while performance in inhomogeneous low-resolution LiDAR scenes improves, grid-based methods remain the state of the art option for low-resolution LiDAR. Further research is required to determine if the ``Existential'' local neighbourhood feature extraction step could be beneficial on data with more variance than HDVMine, or if its improved performance on sparse objects in the scene is always outweighed by the detriment to the higher density objects which make up the majority of a scan.

\section{Acknowledgements}
This research was carried out with support from the company Maptek, from which data was used to create the dataset HDVMine, and software was used to both label and visualise point clouds.

Funding: Ryan Faulkner was supported by an Australian Government Research Training Program (RTP) Scholarship as well as a supplementary University of Adelaide Industry PhD (UAiPhD) Scholarship funded by Maptek; Tat-Jun Chin is SmartSat CRC Professorial Chair of Sentient Satellites.

\clearpage



 \bibliographystyle{elsarticle-num} 
 \bibliography{cas-refs}





\end{document}